\documentclass[preprint,12pt]{elsarticle}

\usepackage{amssymb}
\usepackage{times}
\usepackage{subfigure} 

\usepackage{algorithm}
\usepackage{algorithmic}
\usepackage{amsmath}
\usepackage{adjustbox}
\usepackage{xcolor}
\usepackage{wrapfig}

\begin{document}

\begin{frontmatter}

\title{Tree Memory Networks for Modelling Long-term Temporal Dependencies}


\author[label1]{Tharindu~Fernando\corref{cor1}}
\ead{t.warnakulasuirya@qut.edu.au}

 \author[label1]{Simon~Denman}
\ead{s.denman@qut.edu.au}

\author[label2]{Aaron~McFadyen}
\ead{aaron.mcfadyen@qut.edu.au }

\author[label1]{Sridha~Sridharan}
\ead{s.sridharan@qut.edu.au}

 \author[label1]{ Clinton~Fookes}
\ead{c.fookes@qut.edu.au}

 \cortext[cor1]{Corresponding author at: Image and Video Research Laboratory, SAIVT, Queensland University of Technology, Australia.}
 \address[label1]{Image and Video Research Laboratory, SAIVT, Queensland University of Technology, Australia.}
 \address[label2]{Robotics and Autonomous Systems, Queensland University of Technology, Australia.}

\begin{abstract}
In the domain of sequence modelling, Recurrent Neural Networks (RNN) have been capable of achieving impressive results in a variety of application areas including visual question answering, part-of-speech tagging and machine translation. However this success in modelling short term dependencies has not successfully transitioned to application areas such as trajectory prediction, which require capturing both short term and long term relationships.  In this paper, we propose a Tree Memory Network (TMN) for jointly modelling both long term relationships between multiple sequences and short term relationships within a sequence, in sequence-to-sequence mapping problems. The proposed network architecture is composed of an input module, controller and a memory module. In contrast to related literature which models the memory as a sequence of historical states, we model the memory as a recursive tree structure. This structure more effectively captures temporal dependencies across both short and long term time periods through its hierarchical structure.  We demonstrate the effectiveness and flexibility of the proposed TMN in two practical problems: aircraft trajectory modelling and pedestrian trajectory modelling in a surveillance setting. In both cases the proposed approach outperforms the current state-of-the-art. Furthermore, we perform an in depth analysis on the evolution of the memory module content over time and provide visual evidence on how the proposed TMN is able to map both short and long term relationships efficiently via a hierarchical structure. 
\end{abstract}


\end{frontmatter}

\section{Introduction}
\label{Introduction}

Sequence-to-sequence modelling is a vital element in machine learning and knowledge representation, with multiple application areas including machine translation \citep{shafieibavani2016efficient}, trajectory prediction \citep{judah2014active}, and part-of-speech tagging \citep{lafferty2001conditional}. This problem can be represented as predicting an output sequence, $Y=[\bf{y_1}, \ldots, \bf{y_T} ]$, given an input sequence $X=[\bf{x_1}, \ldots, \bf{x_T}]$. 
 Predicting a future element, $\bf{y_t}$, of the sequence at time instance $t$, utilising the current input to the model, $\bf{x_t}$, and the content of the memory from the previous time step, $\mathrm{M_{t-1}}$, can be represented as,
\begin{equation}
\bf{y_t} = f(\bf{x_t}, \mathrm{M_{t-1}}) . 
\end{equation}

 Modelling long term relationships in between sequences can be considered one of the most challenging problems within the machine learning community \citep{dietterich2002machine,bao2017deep}. Although many memory architectures proposed for sequence-to-sequence modelling are capable of mapping short term relationships, they are less successful when handling long term dependencies \citep{xu2016cached}. \par
Long term relationships within the data are extremely useful and can significantly influence the accuracy of the predictions when considering the repetitive nature of many processes. For instance consider an air traffic modelling problem. Even though short term dependencies such as current weather and neighbouring traffic are the most influential factors, the repetitive nature of the aircraft trajectories and the flight schedules suggests that one can easily deduce a coherence among trajectories over a period of days and/or seasons. For example, at a point in time a certain runway may be in use, so similar trajectories should be observed over the short term, but a sudden change of runway (due to weather) means ``new'' trajectories will be seen. Having a long term memory means that these are not actually ``new'' trajectories, but can instead be recalled from an earlier, similar weather event or operational configuration. \par
\begin{figure}[t]
\begin{center}
\subfigure[High pedestrian flow in $\color{red} \square $]{\includegraphics[width = .3 \textwidth]{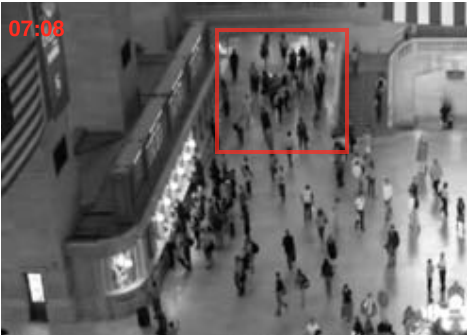}}
\subfigure[Low pedestrian flow in $\color{red} \square $]{\includegraphics[width = .3 \textwidth]{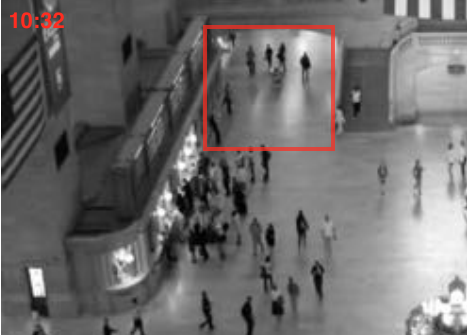}}
\subfigure[Similar flow to (a) in $\color{red} \square $]{\includegraphics[width =.3 \textwidth]{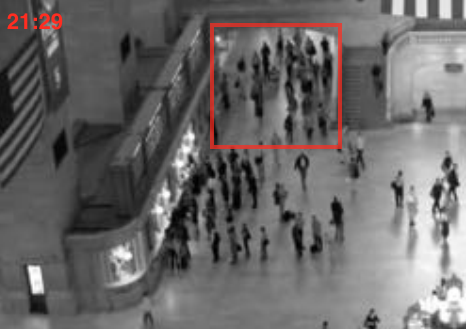}}
\end{center}
\vspace{-5pt}
\caption{Pedestrian flow at different times in the Grand central dataset \cite{Yi_CVPR_2015}. A high pedestrian flow in the highlighted area is observed at the 07:08 time stamp with the arrival of a train before the flow decreases at the 10:32 time stamp. A similar flow to subfigure (a) is observed at the 21:29 time stamp confirming our hypothesis that future pedestrian flow can be anticipated with the aid of historical data.}
\label{fig:fig_similar_contexts}
\end{figure}

A similar logic can be applied when modelling pedestrian behaviour in a surveillance scenario, such as in the example shown in Fig. \ref{fig:fig_similar_contexts}. While the current location of neighbouring pedestrians is most influential, one cannot discard the influence of historical behaviour under similar contexts and events. Pedestrians may be wandering in a free area, and as new trains arrive a new ``flow'' of pedestrian movement appears in response to the congestion. But as we posses historical data from similar contexts, one should be able to accurately anticipate such pedestrian motion.  
\par
In this paper we are interested in efficiently aggregating such long term dependencies among input data. In the sample scenario from the Grand Central dataset \citep{Yi_CVPR_2015} presented in Fig. \ref{fig:fig_similar_contexts}, a high pedestrian flow in the highlighted area is observed at the 07:08 time stamp with the arrival of a train before the flow decreases at the 10:32 time stamp. A similar flow to Fig. \ref{fig:fig_similar_contexts} (a) is observed at the 21:29 time stamp in Fig. \ref{fig:fig_similar_contexts} (c), confirming our hypothesis that future pedestrian flow can be anticipated with the aid of historical data.

\par 
As such, we propose an augmented memory architecture which can be generalised to any sequence to sequence modelling problem. The contributions of this work can be summarised as follows:
\begin{enumerate}
 \item A new recursive memory network architecture capable of modelling long term temporal dependencies, using an efficient tree structure.
 \item Application of the proposed memory architecture to two practical problems: aircraft trajectory modelling and pedestrian trajectory modelling in a surveillance setting, where in both cases we are able to achieve state-of-the-art results.
 \item An in depth analysis on the evolution of the memory module content, where we study the changes in hidden state representations over time and discuss interpretable patterns.
 \end{enumerate}

The two applications we demonstrate the proposed approach on, aircraft trajectory prediction and pedestrian trajectory prediction, are both related sequence-to-sequence modelling tasks, however they have distinct characteristics that illustrate the adaptability of the proposed approach. Aircraft trajectories are primarily a function of the flight schedule, which is typically fixed over a week, but still varies according to changes such as weather and off schedule arrivals/departures. Pedestrian trajectories however are less dependent on a schedule and are more influenced by the behaviour of other nearby pedestrians. \par
We would like to emphasise the fact that even though we are demonstrating our approach on two different application scenarios from the trajectory prediction domain, the varied nature of these problems demonstrates how the proposed model can be directly applied to any sequence-to-sequence prediction problem where modelling long term relationships is necessary. Possible application areas include diver behaviour modelling for autonomous driving \citep{chen2015deepdriving,liu2017visualization,Fernando_2017_ICCV}, text and video synthesis \citep{reed2016generative}, and context aware machine translation \citep{banchs2014principled}. 
 
\section{Related works}
Related work within the scope of this paper can be categorised into memory architectures (Section \ref{sec:memory}), aircraft trajectory prediction approaches (Section \ref{sec:aircraft}) and pedestrian trajectory prediction approaches (Section \ref{sec:pedestrian}). 
\subsection{Memory architectures}
\label{sec:memory}

\begin{figure*}[!h]
    \centering
     \includegraphics[width=.7\textwidth]{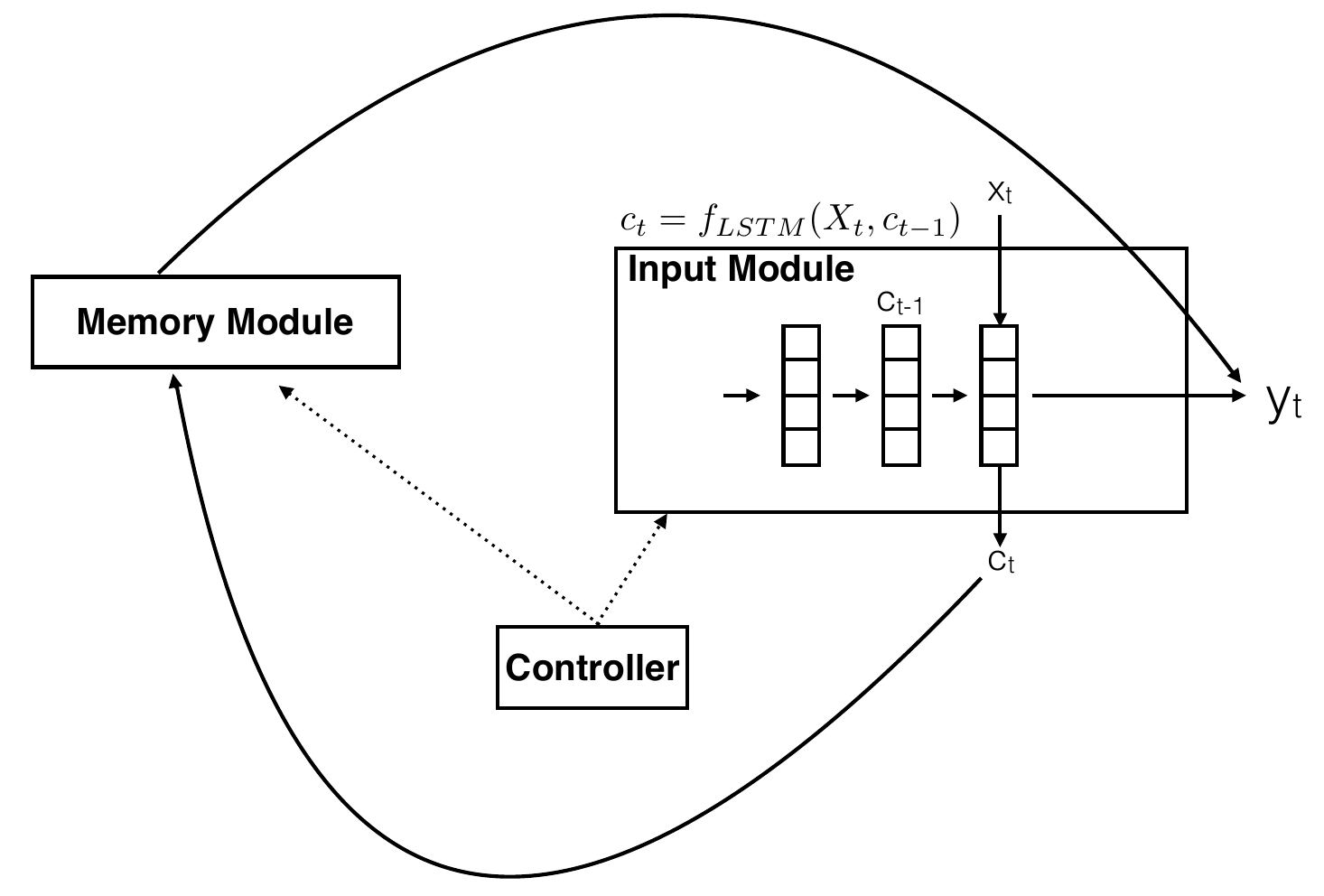}
\caption{Neural Network architecture with an external memory. The memory is used to store important historical facts which can be utilised for future predictions. The controller is responsible for issuing read and write commands in order to take out and write back to the memory. An input module is used to encode and generate a vector representation from the input }
	\label{fig:fig1a}
\end{figure*}

Deep learning models such as Recurrent Neural Networks (RNN) have been applied extensively for many sequence-to-sequence modelling problems and have been capable of producing state-of-the-art results. A number of approaches \citep{weston2014memory,askMeAnything,malinowski2014multi,joulin2015inferring,our_iccv,Fernando_2017_ICCV,our_wacv2} have also utilised what are termed ``memory modules'', to aid prediction. The memory stores important facts from historical inputs and then generates the future predictions based on the stored knowledge. A sample architecture with an input module, controller and an external memory is shown in Fig. \ref{fig:fig1a}. Firstly the input module generates a vector representation, $\mathbf{c_t}$, for the input, $\mathbf{x_t}$, at time instance $t$. The controller then triggers a memory read operation. The memory module, with an attention process, searches the history and outputs relevant facts. The final output is generated by merging $c_t$ with the memory output. Finally, the controller triggers a memory update operation where the memory, $M_{t-1}$, is updated with $\mathbf{c_t}$.  \par

The authors in \cite{weston2014memory} have utilised a memory module to improve performance for natural language processing tasks. Their proposed memory architecture is not fully extendible given the use of an offline feature engineering process using a bag-of-words approach.  In similar works, such as  \cite{joulin2015inferring} and \cite{kaiser2015neural} for image caption generation, and  \cite{malinowski2014multi} and \cite{chen2014learning} for visual question answering, the authors have extensively applied the notion of external memory. The memory architecture, ``episodic memory'', proposed in  \cite{askMeAnything} has been shown to be capable of outperforming the other external memory architectures noted above \cite{weston2014memory, malinowski2014multi, joulin2015inferring, kaiser2015neural, chen2014learning} in terms of accuracy. 

\begin{figure}[!h]
\center
\subfigure[ Episodic memory model proposed in \cite{askMeAnything}  with LSTM memory cells]{\includegraphics[width = .65\linewidth]{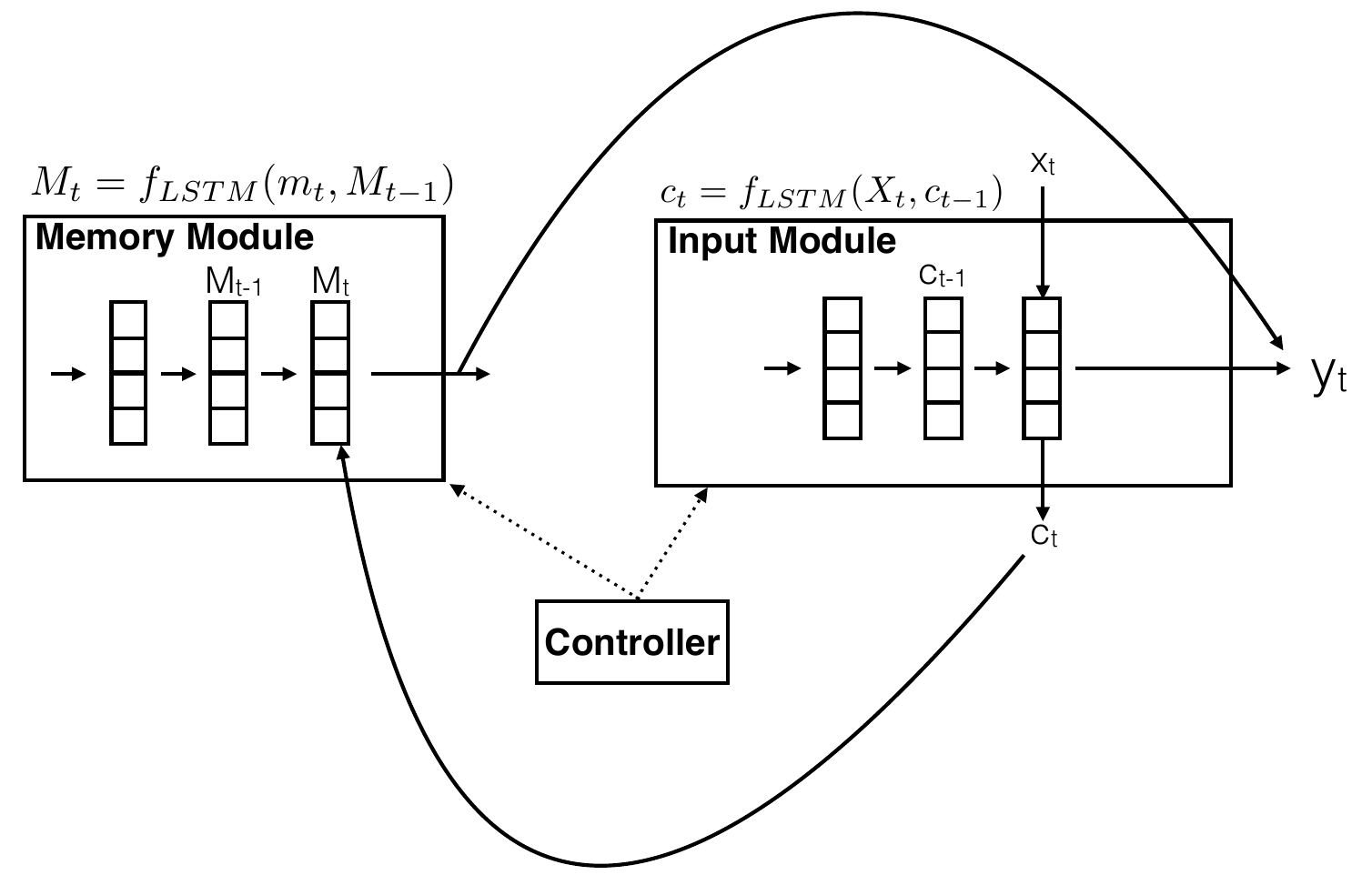}}
\center
\subfigure[Proposed TMN model with S-LSTM memory cells. ]{\includegraphics[width = .65\linewidth]{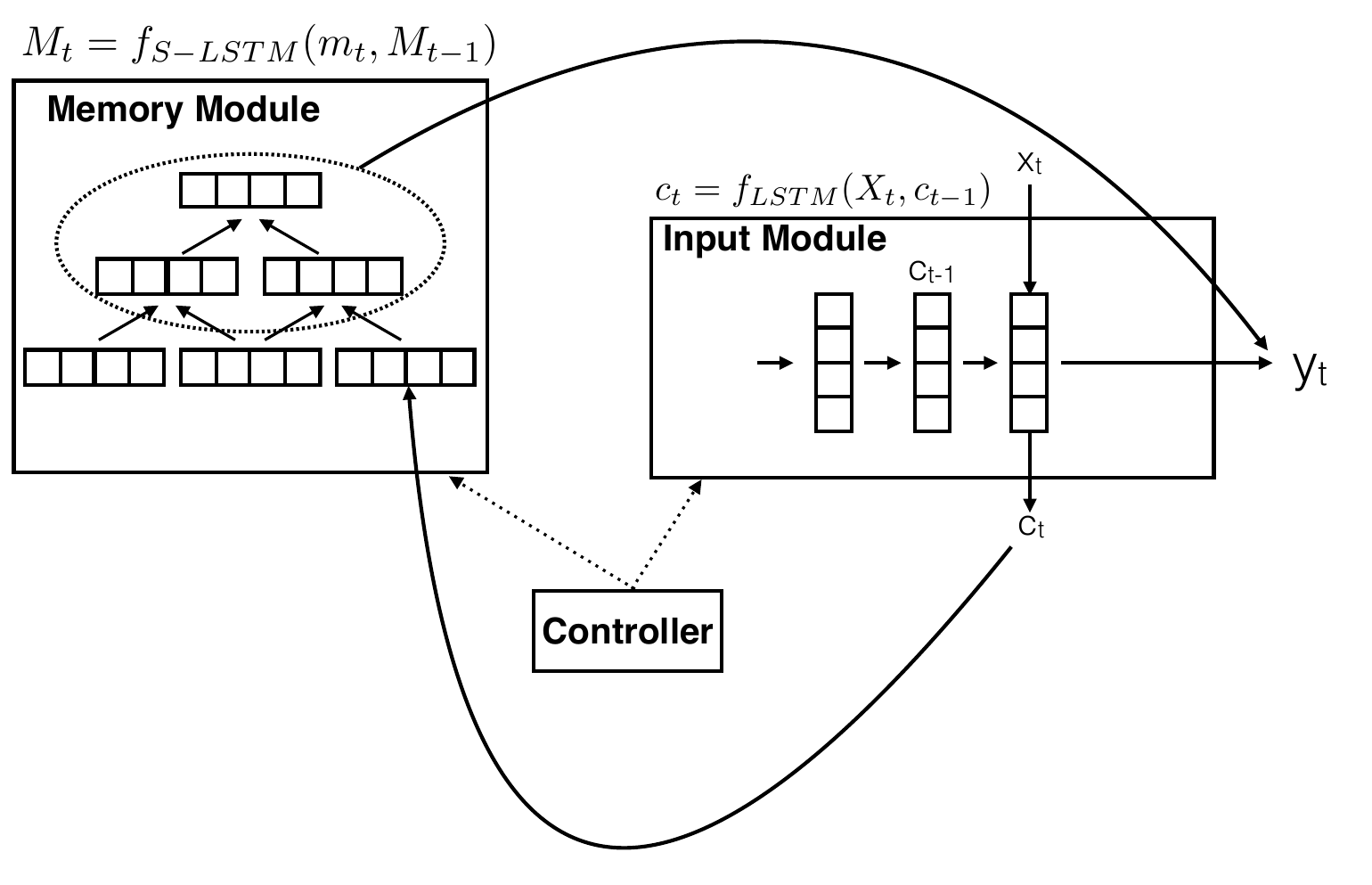}}
\caption{Comparison of the memory model proposed in \cite{askMeAnything} (a) with the proposed memory model (b). In both approaches at time instance $t$, the input module generates representation $\mathbf{c_t}$ for the input  $\mathbf{x_t}$. Then the controller triggers a memory read operation. The memory module, with the attention process, outputs relevant facts. The final output is given by merging $\mathbf{c_t}$ with the memory output. Finally the memory update operation updates the memory, $\mathrm{M_{t-1}}$, with $\mathbf{c_t}$. }
\label{fig:fig1}
\end{figure}

Fig. \ref{fig:fig1} (a) depicts the episodic memory model proposed in \cite{askMeAnything}.  The authors model the ``episodic memory'' as a hierarchical recurrent sequence model utilising the sequential nature of the memory.  The authors propose a generalised neural sequential module with recurrent LSTM memory cells for sequence encoding, memory mechanism and response generation. The above work is further extended in \cite{neuralProgrammer} through incorporating a shared memory architecture. Even with the exemplary results for short term dependency modelling problems, none of the above stated architectures are capable of handling sequences with long term relationships. \par

In approaches such as \citep{askMeAnything,neuralProgrammer,neuralSemanticEncoder} the memory is a composed of a single layer of memory units. The memory update mechanism in \citep{askMeAnything,neuralProgrammer} can be written as,
\begin{equation}
 \mathrm{M_{t}}= f_{LSTM}(\mathbf{m_t},\mathrm{M_{t-1}}),
\end{equation}
where $\mathbf{m_t}$  is a score value that quantifies the relevance of the content of the memory module ($\mathrm{M_{t-1}}$) at time $t-1$ to the current context, $\mathbf{c_t}$ (see \ref{fig:fig1} (a)); where as in \cite{neuralSemanticEncoder} the authors completely update the content of the memory locations based on $\mathbf{m_t}$. \par

The main drawback of using approaches such as \citep{askMeAnything,neuralProgrammer,neuralSemanticEncoder} with long term dependency modelling and big data sets is that, in most cases, the attention mechanism will generate small scores for all the examples as they are all dissimilar to the input. Therefore with those approaches we are required to maintain an extremely large memory sequence as similar inputs only occur after long time intervals. Furthermore, in recurrent models such as in LSTMs, when the sequence becomes too long the output becomes biased towards recent observations \citep{enhancingCombining}, rather than considering the entire set of prior observations equally. \par 

The work of Liang et. al \cite{liang2016semantic,liang2017interpretable} on semantic object parsing has also provided an in-depth analysis on the limitations of sequentially structured LSTMs and how it affects the information flow when modelling data with complex, multilevel correlations. In \cite{liang2016semantic} they propose a graph structured LSTM network where they model the contextual dependencies within an image at the super pixel level at the lowest level of the graph. This idea of hierarchical modelling is extended in \cite{liang2017interpretable} in order to have a dynamically evolving multi-level graph structure instead of a static hierarchy as in \cite{liang2016semantic}. They achieve state-of-the-art results for semantic segmentation via this hierarchical representation learning approach. However, in contrast to their work, which focuses on learning semantic correspondences within a particular example, we are interested in learning long range temporal dependencies in between examples.  \par
Hypothetically, if the memory module has enough non-linearity and if we have a sufficiently large database, any output should be capable of being produced from the contents of the memory. Furthermore the model should be capable of learning and modelling both short and long term contextual effects from the memory contents. This can be seen as a dictionary learning process where the memory is the dictionary being learnt. The memory module should learn to produce an accurate prediction for different combinations of inputs and historical trajectories. \par

Recently the recursive LSTM (S-LSTM) \citep{LSTMOverTree} was proposed where the authors extend the sequential LSTM to tree structures, in which a memory cell can reflect the historical memories of multiple child cells or multiple descendant cells in a recursive process. The authors in \cite{enhancingCombining} have performed an in depth analysis of the strengths and weaknesses of sequential and recursive structures for a neural language modelling task; and found syntactical relationships, such as structure and logic in the input data are best captured via a recursive LSTM (i.e. S-LSTM). In contrast when encoding the semantics of sentences, sequential LSTM models provide state-of-the-art results. \par
The proposed model is illustrated in Fig \ref{fig:fig1} (b). Motivated by the positive characteristics that the S-LSTM architecture exhibits such as feature compression power and the preservation of semantic relationships among data, our approach eschews a sequence representation in favour of a tree-based approach. A detailed analysis of the architecture in comparison to state-of-the-art methods is presented in Section \ref{Relation_to_current_state_of_the_art}.

\subsection{Aircraft trajectory prediction approaches}
\label{sec:aircraft}
Approaches such as \citep{prandini2000probabilistic, paielli1997conflict} utilise probability models for aircraft dynamics to generate predictions of future aircraft motion. They rely solely on the assumptions made regarding the dynamics of the aircraft. Importantly, they ignore all historic information, constituting a major drawback. \par
In \cite{choi2006learning, de2013machine, winder2004hazard} researchers treated aircraft trajectory prediction as a machine learning problem, in which they train the model using historical trajectory data together with weather observations. Most recently the authors in \cite{aircraftTrajectoryPrediction} proposed an approach that considered trajectories as a set of 4 dimensional data cubes,  together with weather parameters. Initially they performed time series clustering on data for segmentation and then learnt a HMM for each cluster. However due to the uncertainty with weather observations, these trajectory prediction approaches become inefficient. \par

Several efforts have been made to improve the trajectory prediction by better wind estimation \citep{mondoloni2003improving, cole2000wind, rekkas1991three, delahaye1992wind, hollister1989using}, yet these approaches have failed to achieve significant improvement in the task of trajectory prediction. Furthermore we would like to emphasise the fact that all of the above stated approaches consider aircrafts individually, without considering the air traffic within the neighbourhood, completely discarding important factors such as the volume and the proximity of nearby air traffic. Even though weather is a vital factor for future predictions it is implicit in the behaviour of neighbouring traffic. Therefore, inferring a notion of weather through neighbouring traffic is computationally inexpensive compared to tedious interpolations that ground based weather observations require \citep{aircraftTrajectoryPrediction}. 

\subsection{Pedestrian trajectory prediction approaches}
\label{sec:pedestrian}
When considering the literature for human behaviour prediction the social force model \citep{Helbing_SF, KoppulaS13, PellegriniEG10, conf/cvpr/YamaguchiBOB11, Jingxin_2012}  and its variants can be considered to be well-established. Such approaches generate attractive and repulsive forces between pedestrians, and thus define the optimal path under different contexts with respect to the neighbourhood. 

Despite the prevalence of social force models, a number of probabilistic approaches have also been proposed. Zhou et al. \cite{ZhouTW15} proposed a a mixture model approach, but this technique ignored the interactions among pedestrians. Wang et al. \cite{WangMNG08} proposed a ``topic model''  which was extended to incorporate spatio-temporal dependencies in \cite{HospedalesGX09} and \cite{EmonetVO11}. All of the above stated approaches use hand-engineered features as the input to the prediction module, which can be considered their main drawback as they fail to account for the semantics of the scene. Depending on the domain knowledge of the feature engineer,  hand-crafted features may only capture abstract semantics of the environment.
\par
Alahi et al. \cite{social_LSTM} removed the need for hand-crafted features via an unsupervised feature learning approach. The authors encode the trajectory of each pedestrian in the scene at that particular time using LSTMs. The hidden states of the neighbouring pedestrians at the immediately preceding time step are used in generating their position at the current time step. As pointed out in \cite{our_WACV}, this approach is only able to generate reactive behaviours such as collision avoidance, and fails to generate smooth trajectories for long term trajectory planning. Fernando et al. \cite{our_WACV} have extended the idea of \cite{social_LSTM} to incorporate the entire trajectory of the pedestrian of interest as well as the neighbouring pedestrians. To the best of our knowledge none of the literature addressing human behaviour prediction has considered the long-term relationships among human behavioural patterns. Motivated by this limitation,  we intend to explore the utility of temporal data for trajectory prediction via a tree memory network. 

\section{Tree Memory Network (TMN) Model}
In this work, we are motivated by the exemplary results that were achieved from a tree structure for the task of discriminative dictionary learning from trajectories in \cite{fernando2015discovering}. In contrast to mapping all the historic data with a shallow layer of recurrent memory cells, we hierarchically map the memory with a bottom up tree structure where all historic states are represented in the bottom layer of the tree, and as we progress up the hierarchy we concatenate the most significant features in order to generate the output at a particular time step. \par
Furthermore, rather than stacking individual recurrent layers such as in \citep{HierarchicalLatentVariable,hierarchicalAutoencoder,gammulle2017two}, we utilise a Tree-LSTM structure as it focusses only on the historical information from its two neighbours. Therefore it can be seen as propagating significant features from two temporally adjacent neighbours to the upper layer. 

\subsection{Input Module}
Let $X^{i}=[\mathbf{x_{1}^{i}}, \mathbf{x_{2}^{i}}, \ldots, \mathbf{x_{T}^{i}}]$ be the $i^{th}$ input sequence where $T$ represents the number of time steps. The input module computes a vector representation $\mathbf{c_{t}}$ for the input sequence via a LSTM layer,
\begin{equation}
\mathbf{c_{t}}=f_{LSTM}(\mathbf{x_{t}},\mathbf{c_{t-1}}),   
\end{equation}
where $\mathbf{c_{t}} \in \mathbb{R}^{k}$, and $k$ is the embedding dimension of the LSTM.

\subsection{ Memory Module}
Consider $N \in \mathbb{R}^{p \times k }$ as the sequence of historical LSTM embeddings, with length $p$ and embedding dimension $k$, that we would like to model as the memory.  It can be seen as a queue structure with length $p$ and each data element within the queue has a dimension of $k$. 
We adapt the S-LSTM model to represent the memory module of the proposed framework.  It extends the general sequential structure in LSTMs to a bottom up tree structure composed of a compressed representation of children nodes to a parent node. This provides us with a principled way of considering long-distance interactions between memory inputs, and avoids the current drawbacks of LSTM models when handling lengthy sequences \cite{enhancingCombining}. 
For simplicity, we represent the recursive-LSTM structure as a binary tree, where each parent node has two child nodes; however the extension of this model to any tree structure is straightforward. 

\subsubsection{Memory Read} 
When computing an output at time instance $t$ we extract out the tree configuration at time instance $t-1$. 
 Let $M_{t-1} \in \mathbb{R}^{k \times (2^l -1)}$ be the memory matrix resultant from concatenating nodes from the tree from the top to $l=[1, \ldots]$ depth. This allows us to capture different levels of abstraction that exist in our memory network. Let $f^{score}$ be an attention scoring function which can be implemented as a multi-layer perceptron \citep{neuralTreeIndexers}, 
\begin{equation}
\mathbf{m_t}=f^{score} (\mathrm{M_{t-1}},\mathbf{c_t}),
\label{eq:fscore}
\end{equation}
\begin{equation}
\alpha=softmax(\mathbf{m_t}).
\label{eq:alp}
\end{equation}
Eq. \ref{eq:fscore} and Eq. \ref{eq:alp} provide an attention mechanism that finds the most relevant memory items given the current input,
\begin{equation}
 \mathbf{z_t}=\mathrm{M_{t-1}}\alpha^T.
 \label{eq:z}
\end{equation}
Then the final output can be represented as, 
\begin{equation}
 y_t= ReLU(W_{out}\mathbf{z_t} + (1-W_{out})\mathbf{c_t}),
\end{equation}
where $W_{out}$ are the output weights.

 \subsubsection{Memory update}
\begin{figure}[t]
\center
\includegraphics[width = .7\linewidth]{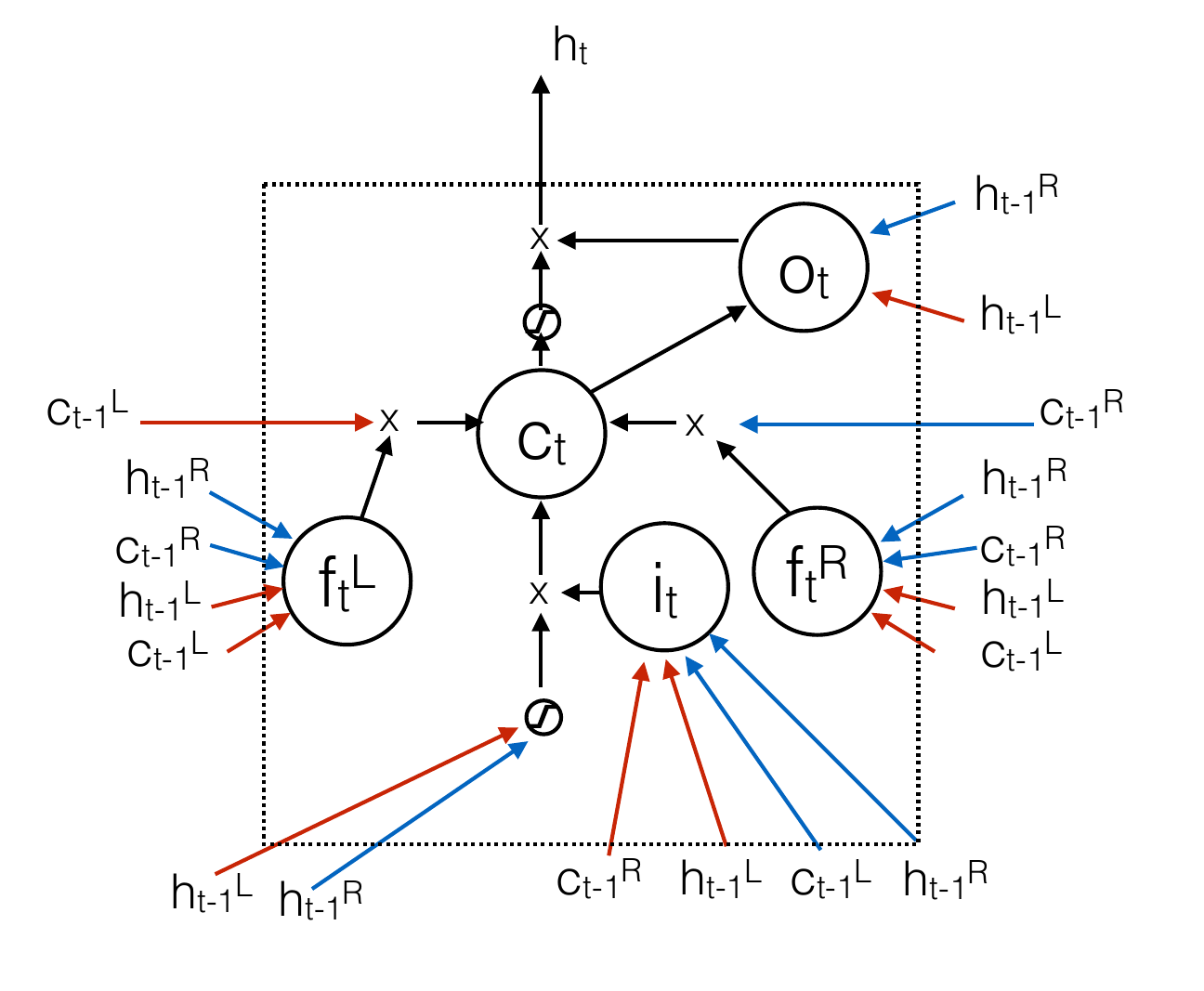}
\caption{Tree memory cell architecture. $f_t^L, f_t^R, o_t, i_t$ represents the left forget gate, right forget gate, output gate and input gate respectively. $\times$ represents multiplication}
\label{fig:memory_cell}
\end{figure}

In the proposed memory architecture each memory cell contains one input gate, $i_t$, one output gate, $o_t$, and two forget gates, $f_t^L$ and $f_t^R$. At time instance $t$ each node in the memory network is updated in the following manner,
 
 \begin{equation}
i_t=\sigma(W_{hi}^Lh_{t-1}^L + W_{hi}^Rh_{t-1}^R +W_{ci}^Lc_{t-1}^L +W_{ci}^Rc_{t-1}^R ) ,
\label{eq:i_t}
\end{equation}
\begin{equation}
f_t^L=\sigma(W_{hf_l}^Lh_{t-1}^L + W_{hf_l}^Rh_{t-1}^R +W_{cf_l}^Lc_{t-1}^L +W_{cf_l}^Rc_{t-1}^R ) ,
\end{equation}
\begin{equation}
f_t^R=\sigma(W_{hf_r}^Lh_{t-1}^L + W_{hf_r}^Rh_{t-1}^R +W_{cf_r}^Lc_{t-1}^L +W_{cf_r}^Rc_{t-1}^R ) ,
\end{equation}
\begin{equation}
\beta=W_{hc}^Lh_{t-1}^L+ W_{hc}^Rh_{t-1}^R ,
\end{equation}
\begin{equation}
c_t^P=f_t^L \; \times \; c_{t-1}^L + f_t^R \; \times \;  c_{t-1}^R + \;  i_t  \; \times \; tanh(\beta) ,
\end{equation}
\begin{equation}
o_t=\sigma(W_{ho}^Lh_{t-1}^L + W_{ho}^Rh_{t-1}^R +W_{co}^Pc_{t}^P) ,
\end{equation}
\begin{equation}
h_t^P=o_t \;  \times \; tanh(c_t^P) ,
\label{eq:h_t}
\end{equation}

where $h_{t-1}^L $,  $h_{t-1}^R $, $c_{t-1}^L$ and $c_{t-1}^R$ are the hidden vector representations and cell states of the left and right children respectively. The relevant weight vectors, $W$, are represented with appropriate super and subscripts where the superscript represents the relevant child node, and the subscript represents the relevant gate and the vector the weight is attached to. The process is illustrated in Fig \ref{fig:memory_cell}.

\subsection{Relation to current state of the art}
\label{Relation_to_current_state_of_the_art}

The major difference between the proposed approach and current state of the art methods such as  \cite{askMeAnything, neuralProgrammer,neuralSemanticEncoder} is the representation of the memory. In those approaches the memory is a composed of a single layer of memory units. The memory update mechanism in \cite{askMeAnything} and \cite{neuralProgrammer} can be written as,
\begin{equation}
 \mathrm{M_{t}}= LSTM(\mathbf{m_t},\mathrm{M_{t-1}}),
\end{equation}
where $m_t$ can be obtained using Eq. \ref{eq:fscore}. In \cite{neuralSemanticEncoder} the authors completely update the contents of the memory locations with respect to the output of Eq. \ref{eq:z}. This process can be written as,
\begin{equation}
 \mathrm{M_{t}}= \mathrm{M_{t-1}(}1-(\mathbf{z_t} \otimes \mathbf{e_k})^T) + (h_t \otimes \mathbf{e_p})(\mathbf{z_t} \otimes \mathbf{e_k})^T ,
\end{equation}
where $1$ is a matrix of ones and $e_k$ and $e_p$ are vectors of ones. $\otimes$ denotes the outer product which duplicates its left vector $p$ or $k$ times to form a matrix.
In contrast we model our memory using a binary tree structure where it is updated in a bottom up fashion utilising Eq. \ref{eq:i_t} to \ref{eq:h_t}. \par
The main drawback of using approaches such as \citep{askMeAnything,neuralProgrammer,neuralSemanticEncoder} with long term dependency modelling and big data sets is that, in most cases, the attention mechanism will generate small score values for all the examples as they are dissimilar to the input. Therefore with those approaches we are required to maintain an extremely large memory sequence as similar inputs only occur after long time intervals. However in recurrent models such as in LSTMs, when the sequence becomes too long the output becomes biased towards recent historic states \citep{enhancingCombining}, rather than considering the entire set of historic states equally. In contrast, we represent memory with a tree structure where we learn the logical coherence among neighbouring memory cells with different levels of abstraction. \par

\section{Experimental results}
We present the experimental results on two trajectory datasets: an aircraft trajectory database from the south east Queensland (SEQ) region of Australia; and a widely utilised pedestrian trajectory database. These datasets are specifically chosen to demonstrate the capability of the proposed model to handle varying dimensionalities and temporal relationships and present its real world applicability. 

\subsection{Experiment 1: Terminal Area Air Traffic Prediction}
\label{sec:ex_1}
We obtain air traffic data from the south east Queensland (SEQ) region in Australia from 30-11-2014 to 30-11-2015. We use the real position reports recorded by the Australian Advanced Air Traffic System (TAAATS) used for air traffic management in Australia \citep{mcfadyen2016terminal}. As a pre-processing step, the data is transformed into trajectories utilising the aircraft identification tags and the reported timing. Trajectories with less than 3 position reports are removed as they are too short for the trajectory modelling task. Finally, each trajectory is re-sampled to a length of 50 data points. When resampling, we interpolate the data points such that they have equal distance in the time domain. This is done to ensure sufficient data for training by up-sampling small trajectories. This gives us 260,735 trajectories. The aircraft trajectories are represented as 3 dimensional data streams where each point $\bf{x_t^{i}}$ of the input sequence $X^{i}=[\bf{x_{1}^{i}}, \bf{x_{2}^{i}}, \ldots, \bf{x_{T}^{i}}]$ can be represented as,
\begin{equation}
\bf{x_t^{i}}=\left (
                \begin{array}{ll}
                  {x}^{i}_{t}\\
                  {y}^{i}_{t}\\
                  {z}^{i}_{t}
                \end{array}
              \right ),
\end{equation}
where $T$ is the number of time steps in the $i^{th}$ input sequence.
In this experiment we observed the first 25 frames of the aircraft trajectory, and predicted the next 25 frames. For training we selected the first 182,515 (i.e 70\%) trajectories chronologically. The remaining 78,220 trajectories were used for testing. 
\par
Based on the recommendations provided in \citep{aircraftTrajectoryPrediction, gong2004methodology, paglione2007implementation}, we measure the following three error metrics for the aircraft trajectory prediction experiment. The trajectory prediction errors are calculated for each observed radar track point in each input trajectory segment. Let $n$ be the number of trajectories in the testing set, and we seek to predict the trajectory for the time period $t= T^{obs}+1$ to $T^{pred}$, having observed the trajectory of the same aircraft from $t= 1$ to $T^{obs}$.  Let the predicted course from North for the trajectory $i$ at $t^{th}$ time instance be denoted by $\hat{\theta^{i}_{t}}$. Then,
\begin{equation}
\Delta{x}^{i}_{t}= \hat{x^{i}_{t}} - x^{i}_{t} ,
\end{equation}
and,
\begin{equation}
\Delta{y}^{i}_{t}= \hat{y^{i}_{t}} - y^{i}_{t} .
\end{equation}
Now we can define,
\begin{enumerate}
\item \textit{Average along track error (AE): } 
\begin{equation}
AE= \cfrac{\sum\limits_{i=1}^{n}\sum\limits_{t=T^{obs}+1}^{T^{pred}}(\Delta{x}^{i}_{t} sin(\hat{\theta^{i}_{t}}) + \Delta {y}^{i}_{t} cos(\hat{\theta^{i}_{t}}))}{n(T^{pred}-(T^{obs}+1))}
\end{equation}

\item \textit{Average cross track error (CE): } 
\begin{equation}
CE= \cfrac{\sum\limits_{i=1}^{n}\sum\limits_{t=T^{obs}+1}^{T^{pred}}(\Delta{x}^{i}_{t} cos(\hat{\theta^{i}_{t}}) - \Delta {y}^{i}_{t} sin(\hat{\theta^{i}_{t}}))}{n(T^{pred}-(T^{obs}+1))}
\end{equation}

\item \textit{Average altitude error (ALE): } 
\begin{equation}
ALE= \cfrac{\sum\limits_{i=1}^{n}\sqrt{\sum\limits_{t=T^{obs}+1}^{T^{pred}}(\hat{z^{i}_{t}} - z^{i}_{t} )^2}}{n(T^{pred}-(T^{obs}+1))}
\end{equation}
\end{enumerate}

\begin{figure}[t]
\begin{center}
\subfigure[Average altitude error vs length of memory module, $p$]{\includegraphics[width = .3 \textwidth]{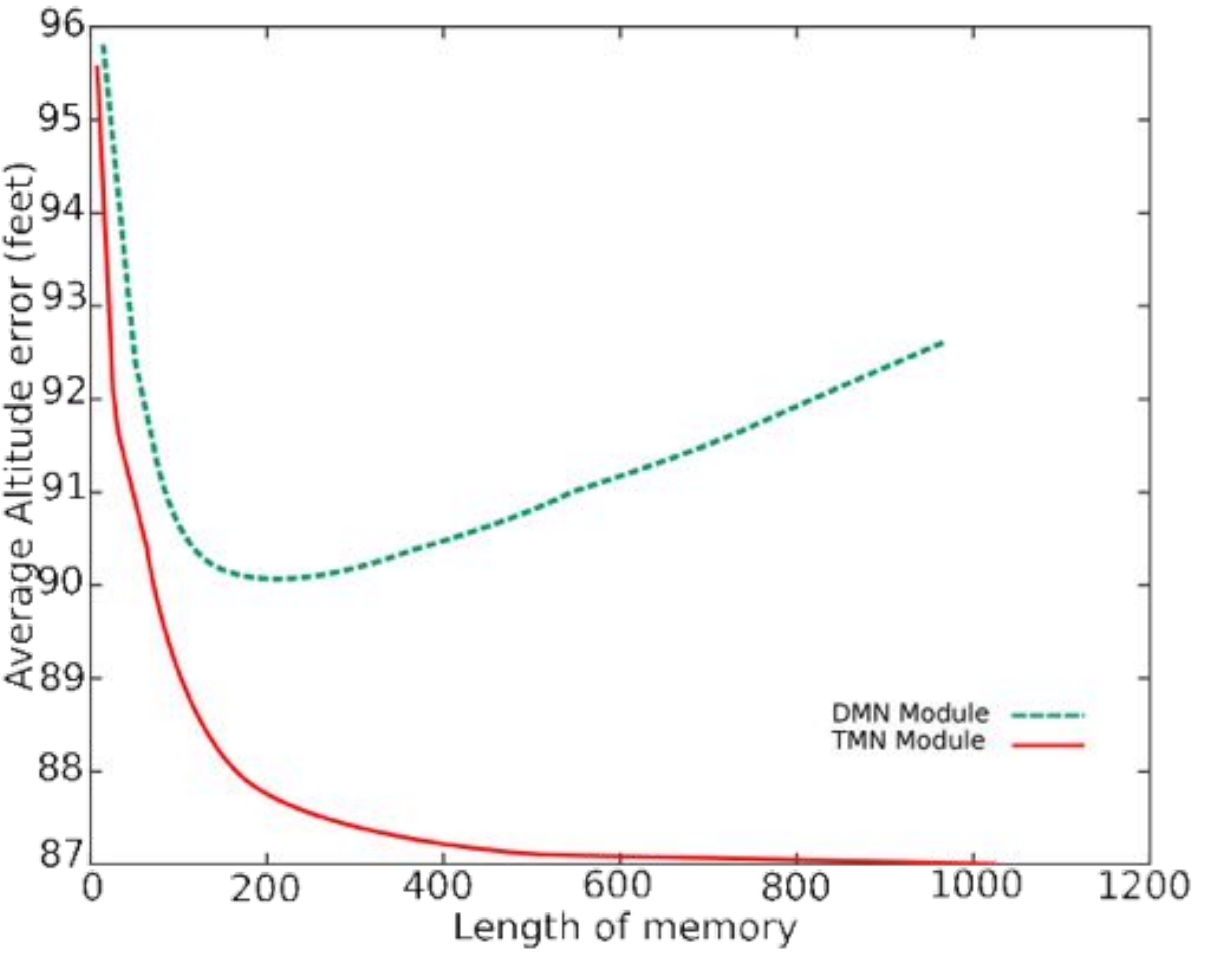}}
\subfigure[Average altitude error vs embedding dimension, $k$]{\includegraphics[width = .3 \textwidth]{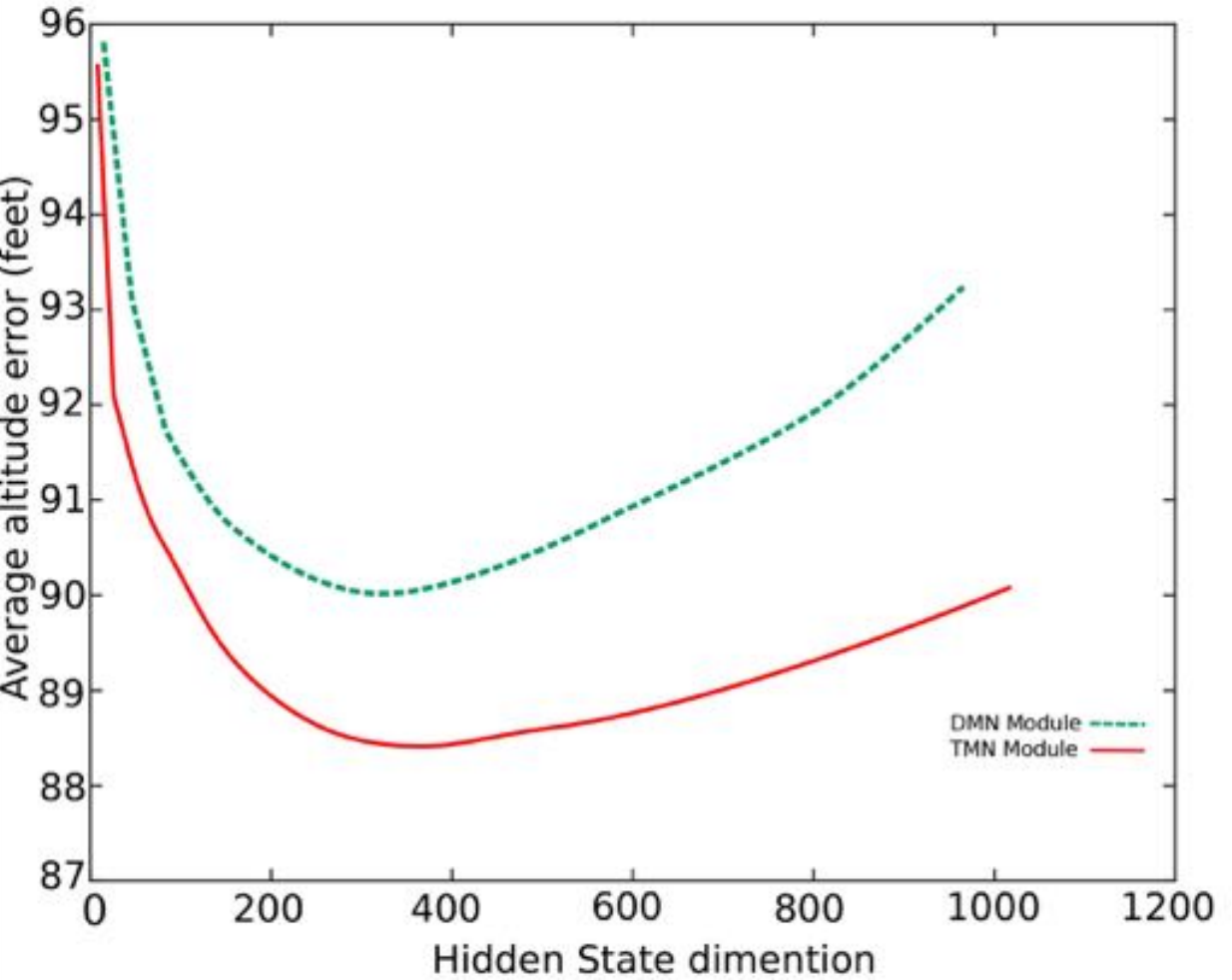}}
\subfigure[Average altitude error vs number of levels, from the top of the memory tree, in the memory read, $l$]{\includegraphics[width = .285 \textwidth]{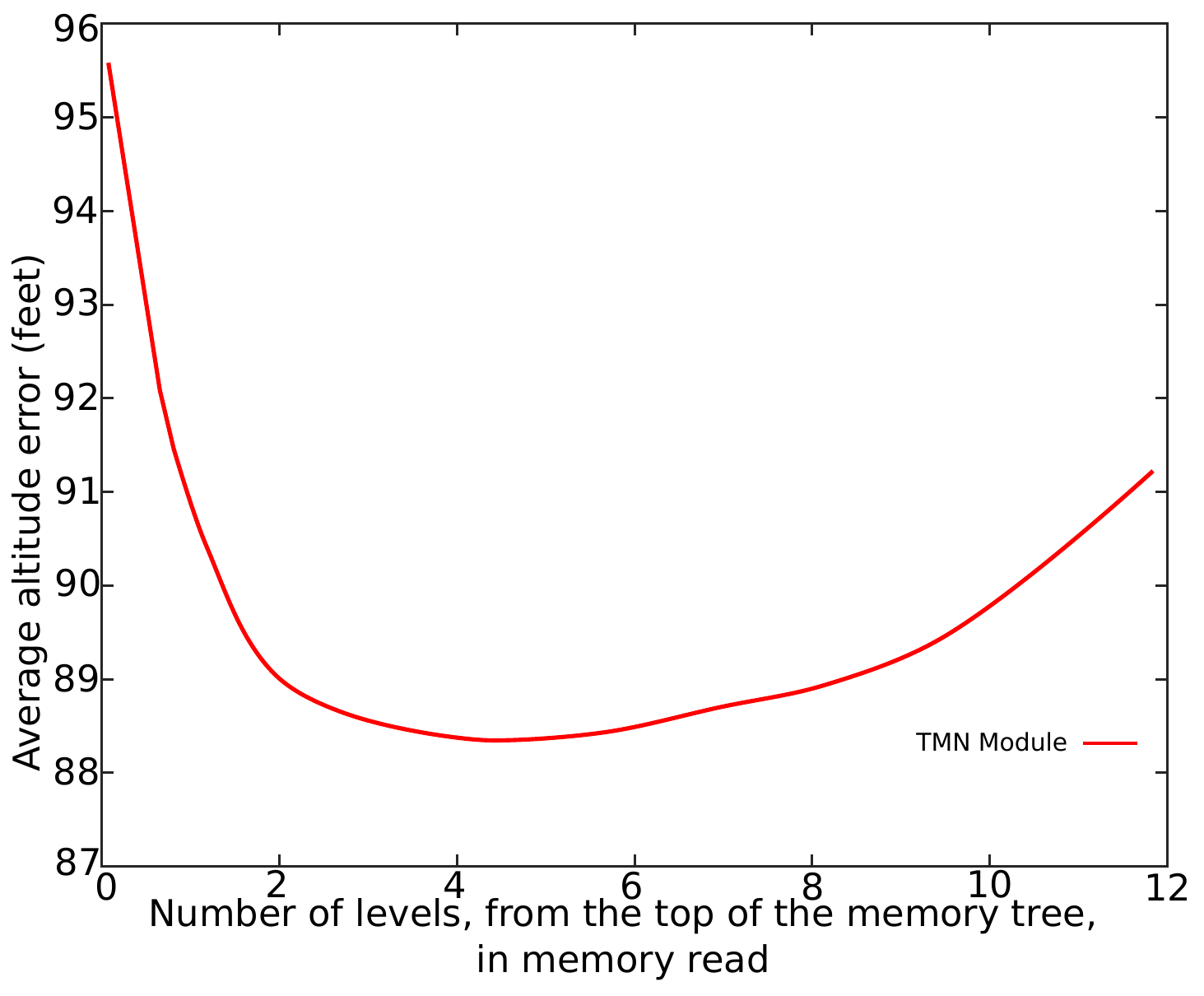}}
\end{center}
\caption{Parameter evaluation for length of memory module, $p$, embedding dimension, $k$, and the number of levels, from the top of the memory tree, in the memory read, $l$ }
\label{fig:fig_error}
\end{figure}

As baseline models we implement the HMM approach (\textbf{HMM}) proposed in \cite{aircraftTrajectoryPrediction} and the Dynamic Memory Networks approach (\textbf{DMN}) given in \cite{askMeAnything}. For the \textbf{HMM} the required weather data is obtained from the Australian Bureau of Meteorology \citep{aus_bom}. We observed wind speed and direction and temperature at one minute frequency. The data interpolation and parameter quantisation is performed in the same way as in \cite{aircraftTrajectoryPrediction}. \par

\subsubsection{Prediction under normal conditions}
Hyper parameters, the length of the memory module, $p$, and the embedding dimension, $k$, of the proposed memory module (\textbf{TMN}) and \textbf{DMN} are evaluated experimentally. Fig. \ref{fig:fig_error} (a) shows the variation in average altitude error against $p$ for the \textbf{TMN} and for \textbf{DMN} modules in solid red and dashed green lines respectively. For the proposed \textbf{TMN}, as the error converges around $p = 500$, we set the value of $p$ as 512. For \textbf{DMN} the plot shows that error decreases at first and it starts increasing again when the length of the memory exceeds 200 hidden units. This verifies our assertion that naive memory models with sequential LSTM architectures fail to model long term dependencies. As $p=180$ gives the lowest altitude error, for the \textbf{DMN} model we $p$ to 180. We evaluate the optimal embedding dimension, $k$, in a similar manner. Fig. \ref{fig:fig_error} (b) shows the variation of average altitude error against $k$ for the \textbf{TMN} and for \textbf{DMN} modules. For both modules $k=300$ produces the smallest altitude error, and as such we set the embedding dimension to 300 units. 
\par Finally, we evaluate the number of levels from the top of the memory tree in the memory read, $l$, of the proposed memory module (\textbf{TMN}). The evaluation results, shown in Fig. \ref{fig:fig_error} (c), suggest that $l=4$ produces optimal results. It is observed that the error is reduced until $l=4$ before increasing again when the number of levels in the memory read operation exceeds 6 levels. This is due to the density of the extracted memory activation. Using the tree structure of the memory we capture the information in a hierarchical manner, where only vital information from the bottom layers is passed to the top layer. Therefore when the extracted matrix becomes too dense, the decoding function fails to extract pertinent information, and the performance degrades. 
\par We train the TMN model using stochastic gradient descent (SGD) with momentum. Evaluation results are presented in Table \ref{tab:tab_1}.

\begin{table}[!h]
  \centering

  \begin{tabular}{|c|c|c|c|}
    \hline
   Metric  & HMM & DMN & TMN \\
    \hline
   AE & 1.103  & 1.039 & \textbf{1.020} \\
  
    \hline\hline
    CE & 1.042 & 1.056 & \textbf{1.011} \\

       \hline\hline
    ALE& 147.801   & 92.039 & \textbf{87.001} \\
   
    \hline
  \end{tabular}

  \caption{Quantitative results for aircraft trajectory prediction. In all the methods the forecast trajectories are of length 25 frames. The first row reports the along track error (AE), the second row shows cross track error (CE) and the final row shows the altitude error (ALE).}
  \label{tab:tab_1}
\end{table}

The results in Table \ref{tab:tab_1} illustrate the ability of the proposed model to infer different modes of air traffic behaviour. We note that without explicitly modelling the weather or neighbourhood, the proposed architecture is able to learn the salient aspects and long term dependencies which are necessary for modelling aircraft trajectories. The accuracy improvement from the \textbf{DMN} to the \textbf{TMN} model, demonstrates that flat memory architectures such as \cite{askMeAnything} fail to capture long term dependencies; where as our proposed hierarchical memory architecture is able to successfully learn those long term relationships. In Section \ref{Analysis_of_memory_activations} we perform an in depth analysis on the hidden state activations of the memory module which illustrates how the proposed multi-layer architecture generates future trajectories while encoding the necessary information from the history. \par

\begin{figure}[!htpb]
\begin{center}
\subfigure[Latitude vs Longitude]{\includegraphics[width = .38\textwidth]{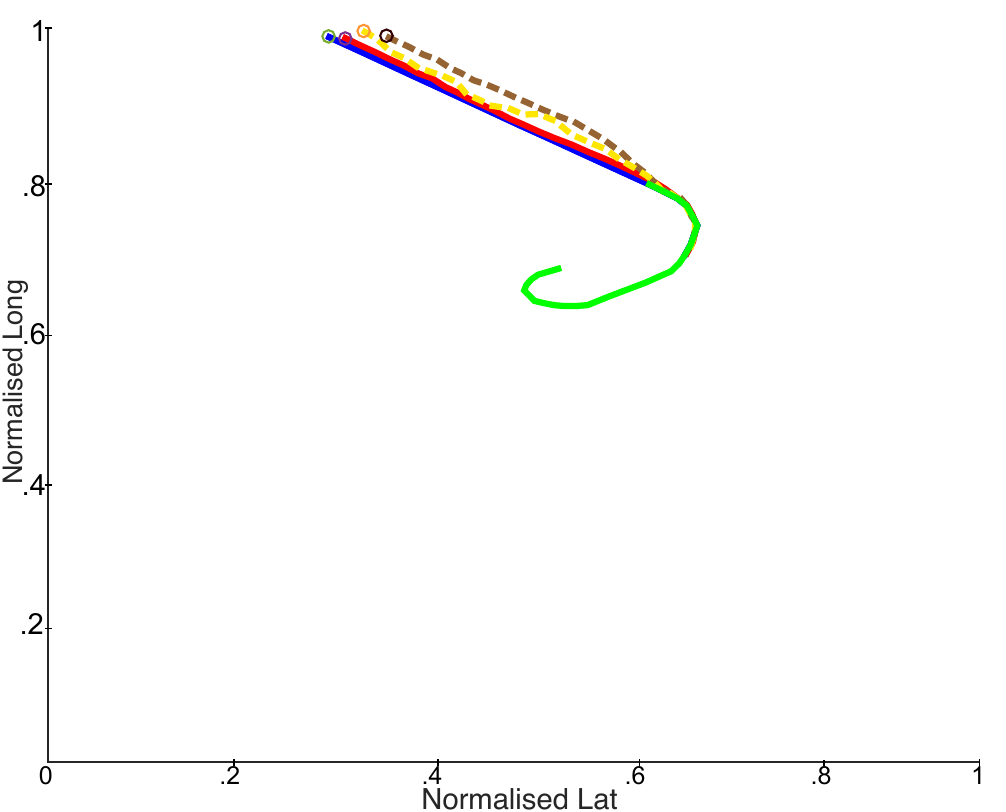}}
\subfigure[Latitude vs Altitude]{\includegraphics[width = .38\textwidth]{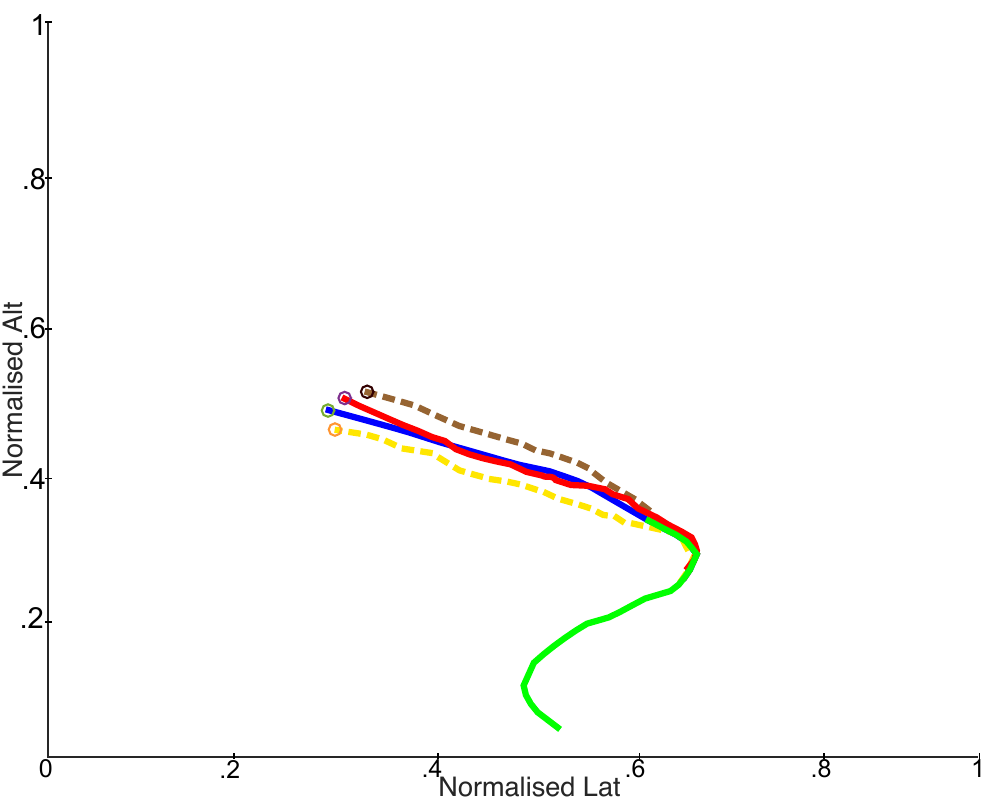}}

\subfigure[Latitude vs Longitude ]{\includegraphics[width = .38\textwidth]{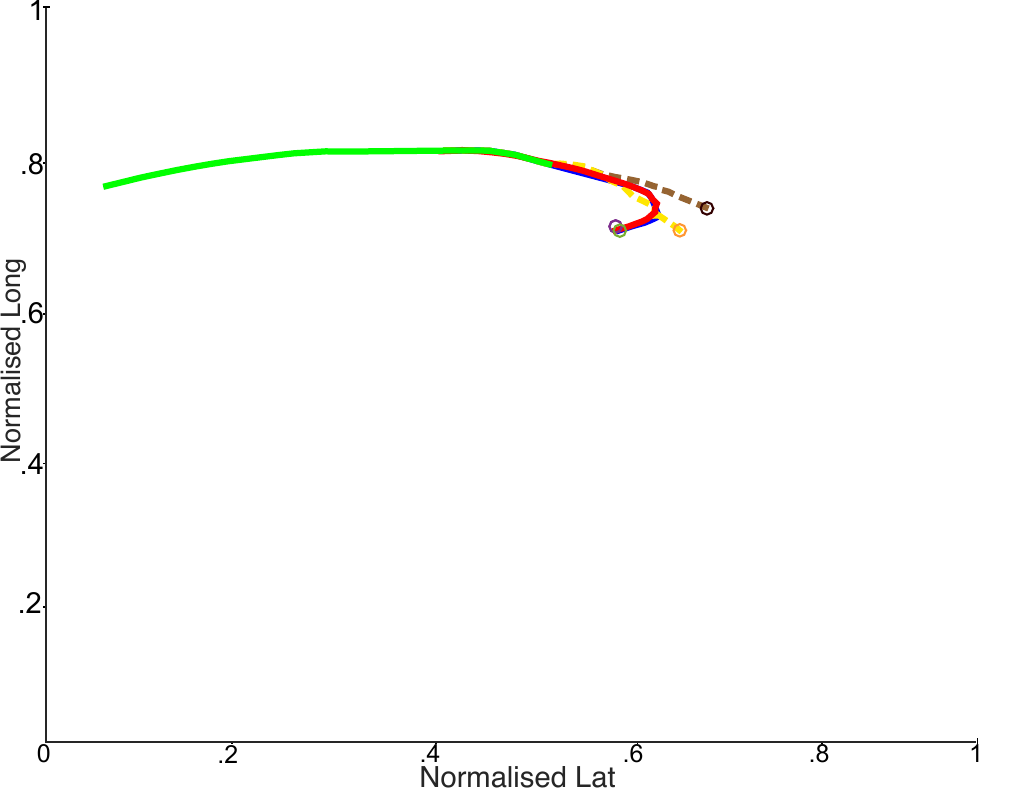}}
\subfigure[Latitude vs Altitude]{\includegraphics[width = .38\textwidth]{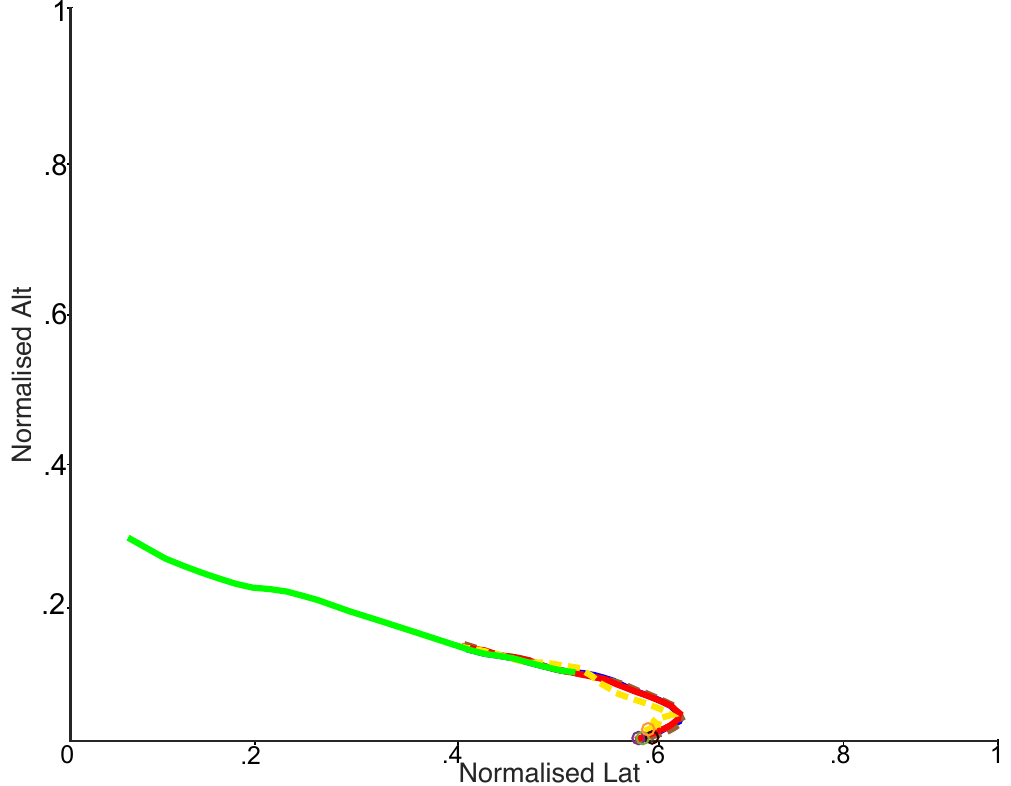}}

\subfigure[Latitude vs Longitude]{\includegraphics[width = .38 \textwidth]{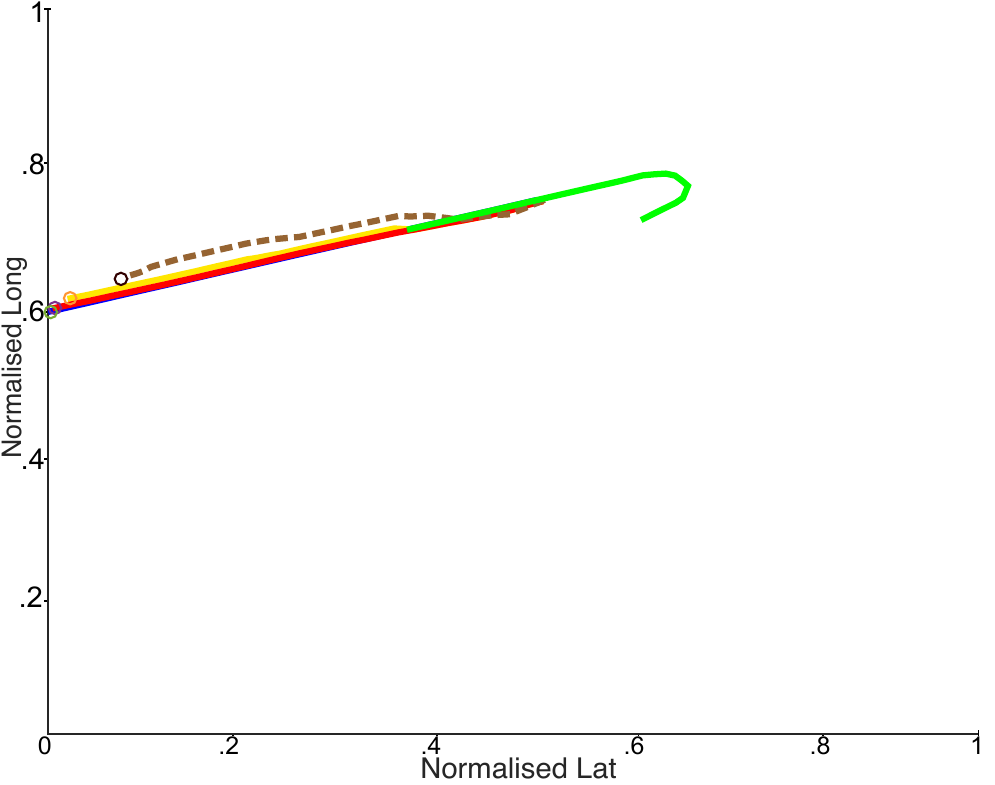}}
\subfigure[Latitude vs Altitude]{\includegraphics[width = .38\textwidth]{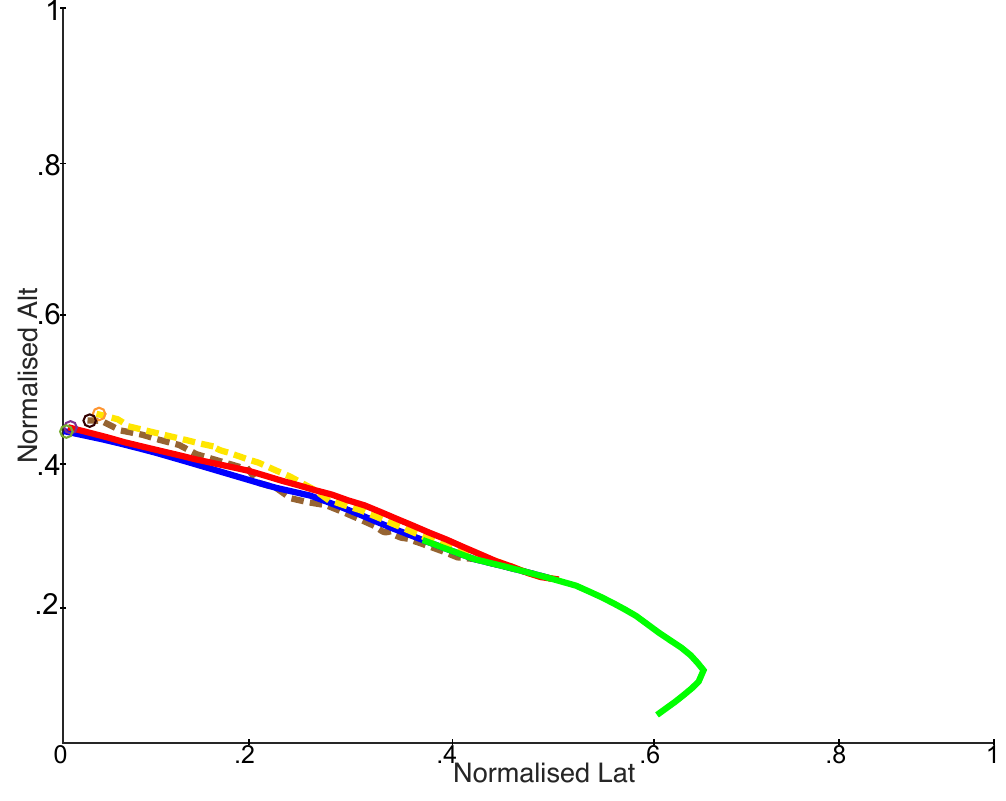}}

\subfigure[ Latitude vs Longitude]{\includegraphics[width = .38 \textwidth]{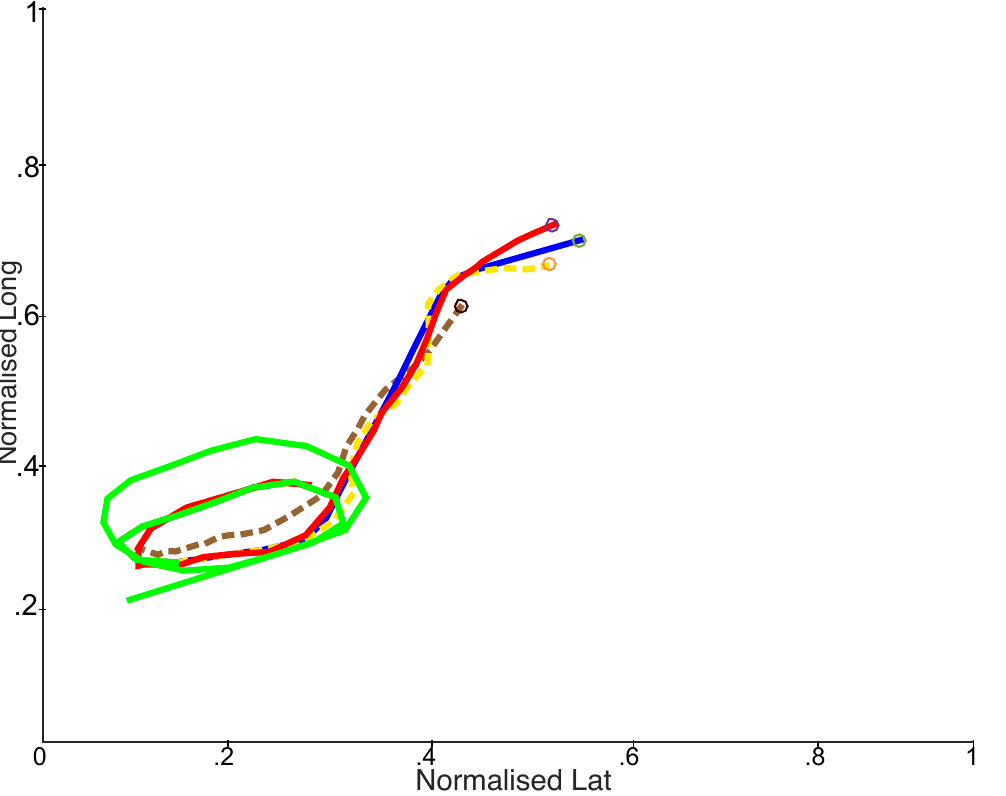}}
\subfigure[ Latitude vs Altitude]{\includegraphics[width = .38\textwidth]{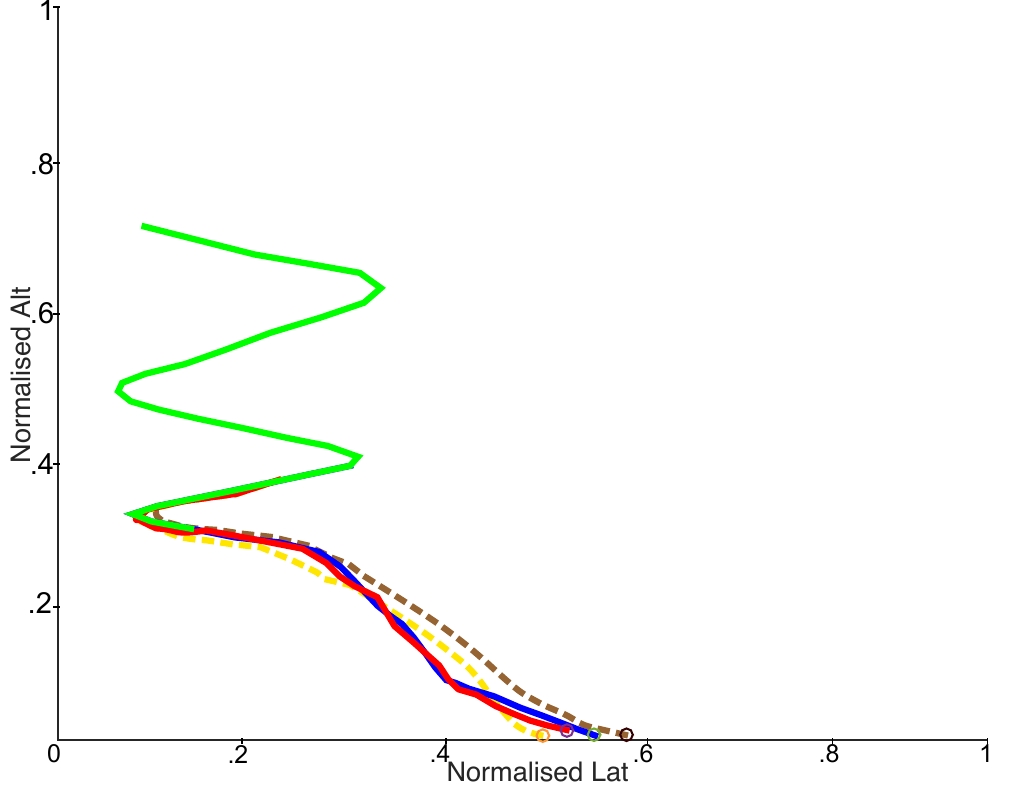}}
\end{center}

\caption{Qualitative results under nominal conditions: Each row represents a particular example. Given (in green), Ground Truth (in Blue) and Predicted trajectories from the $\textbf{TMN}$ model (in red), from the \textbf{DMN} model (in yellow), from the \textbf{HMM} model (in brown).}
\label{fig:fig5}
\end{figure}

In Fig. \ref{fig:fig5} we show prediction results of the \textbf{HMM}, \textbf{DMN} and our approach (\textbf{TMN}) on the aircraft trajectory dataset. We show the trajectories on normalised scales for visual clarity as it better demonstrates the dispersion of the predictions from the ground truth. It should be noted that our model generates more accurate predictions across the highly varied scenarios depicted in the database (i.e take off, landing, cruising, etc). For instance in the 1st and 2nd rows we show how the same model adapts to takeoff and landing scenarios.  In the last row of Fig. \ref{fig:fig5} we show some failure cases. The reason for such deviations from the ground truth were mostly due to sudden turns and movements. Even though these trajectories do not match the ground truth, the proposed method still outperforms the current state-of-the-art methods, and generates more realistic trajectories. 

Considering the results presented in Tab. \ref{tab:tab_1} and the visualisation in Fig. Fig. \ref{fig:fig5}, we observe that the \textbf{HMM} approach performs poorly in the ALE metric compared to the CE and AE metrics. Our database contains variety of aircraft manoeuvres including take off, landing and cruising; and the landing and takeoff manoeuvres involve sudden changes in the aircraft motion which is hard to capture with the limited capacity of the \textbf{HMM}. This is reflected in higher error values for ALE metric to CE and AE, which only consider the dispersions along the latitude and longitude directions.

The \textbf{DMN} module improves upon the \textbf{HMM}'s performance using a sequential memory, which acts as a short term memory of the aircraft motion patterns. This captures the dynamics within a particular trajectory but cannot match dependencies across long time spans, such as how different flight schedules affect the trajectory patterns, and attend to them systematically to extract important information. This results in the \textbf{TMN} module achieving the best performance in all considered metrics.

\subsubsection{Handling different weather conditions}
In order to verify the capability of the proposed model to understand and capture the effect of weather on aircraft trajectory prediction via historical trajectories alone, we conducted a separate experiment where the model is used to predict the air traffic on a stormy day. Severe storms affected South East Queensland on 28th of November 2015. We tested the model with the trajectories from 27th November to 29th November (1916 trajectories) as it allows a sufficient number of examples to initialise the memory module. These examples are not used for training the model. We compare the proposed model against the \textbf{HMM} \citep{aircraftTrajectoryPrediction}, which explicitly incorporates weather information.

\begin{table}[!h]
  \centering
  \begin{tabular}{|c|c|c|c|}
    \hline
   Metric  & HMM & TMN \\
    \hline
   AE & 1.689  & \textbf{1.146} \\
  
    \hline\hline
    CE & 1.967  & \textbf{1.513} \\

       \hline\hline
    ALE& 203.754 & \textbf{88.397} \\
   
    \hline
  \end{tabular}
  \caption{Quantitative results for aircraft trajectory prediction under stormy conditions. For both methods the forecast trajectories are of length 25 frames. The first row reports the along track error (AE), the second row shows cross track error (CE) and the final row shows the altitude error (ALE).}
  \label{tab:tab_3}
\end{table}
Comparing Table. \ref{tab:tab_3} with Table. \ref{tab:tab_1}, the accuracy of the \textbf{TMN} predictions are slightly reduced, but are still more accurate than \cite{aircraftTrajectoryPrediction}, in which the error has increased dramatically indicating that the baseline model has not adapted well to the changed weather conditions. 

\begin{figure}[!htpb]
\begin{center}

\subfigure[Latitude vs Longitude]{\includegraphics[width = .38 \textwidth]{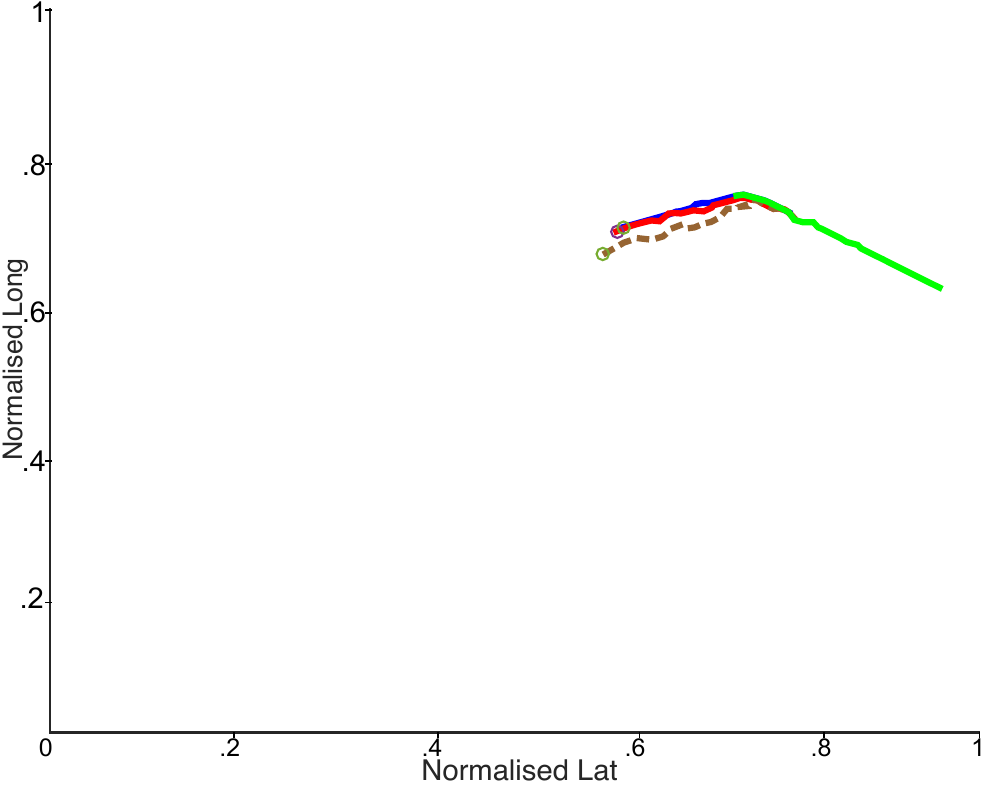}}
\subfigure[Latitude vs Altitude]{\includegraphics[width = .38\textwidth]{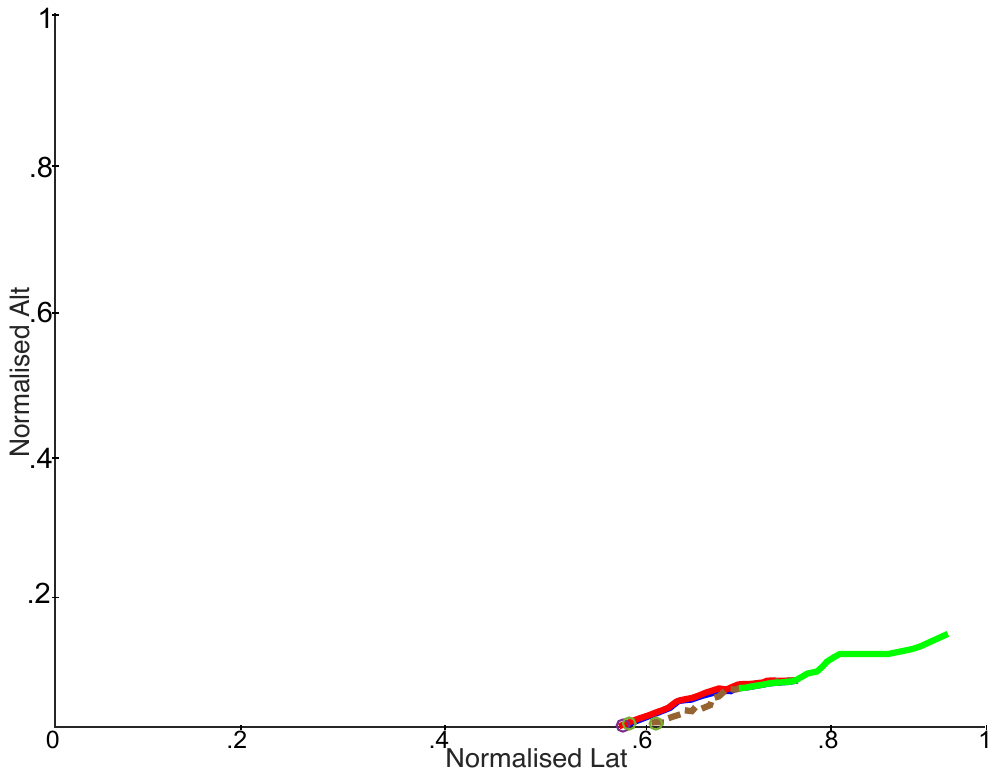}}

\subfigure[Latitude vs Longitude]{\includegraphics[width = .38 \textwidth]{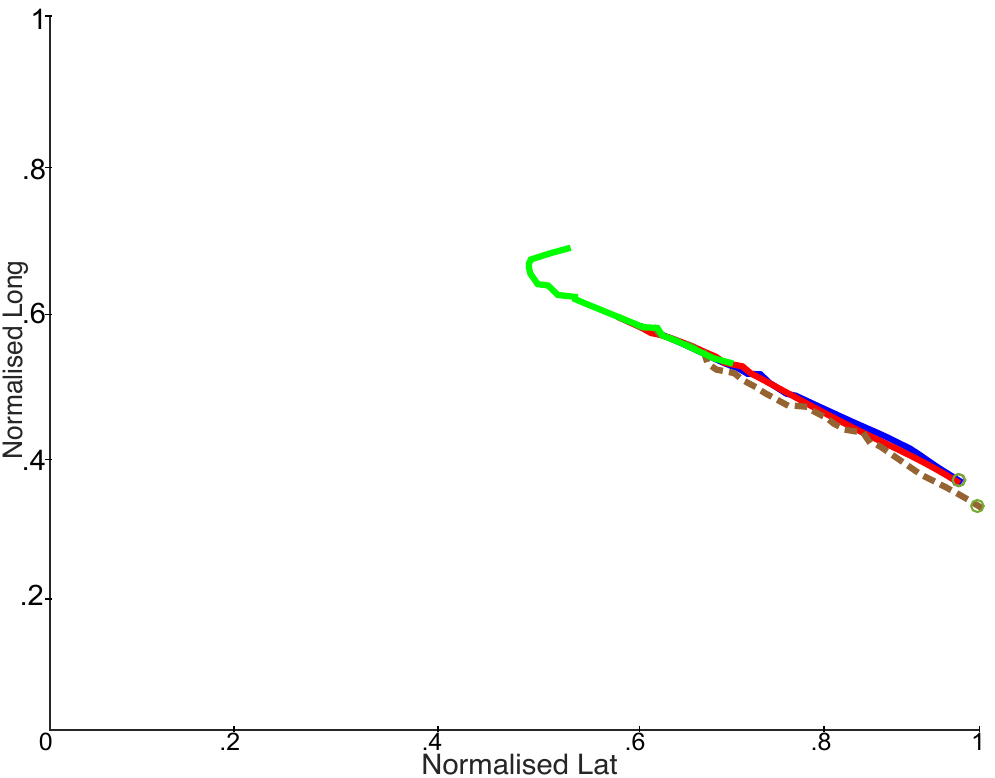}}
\subfigure[Latitude vs Altitude]{\includegraphics[width = .38\textwidth]{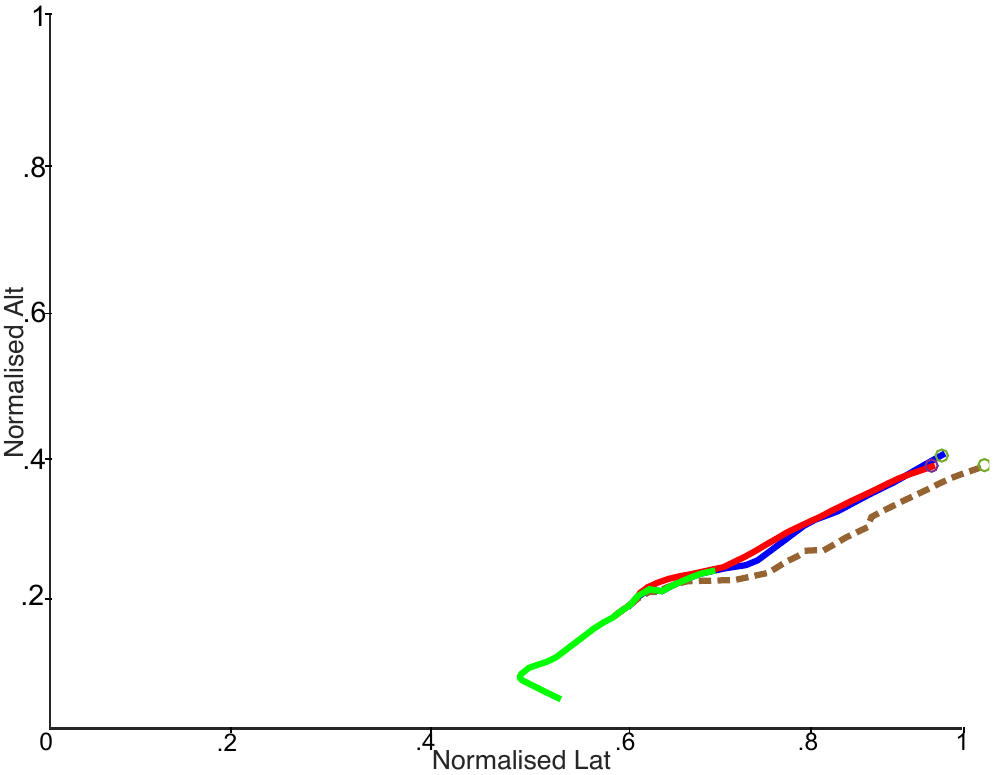}}

\subfigure[Latitude vs Longitude]{\includegraphics[width = .38 \textwidth]{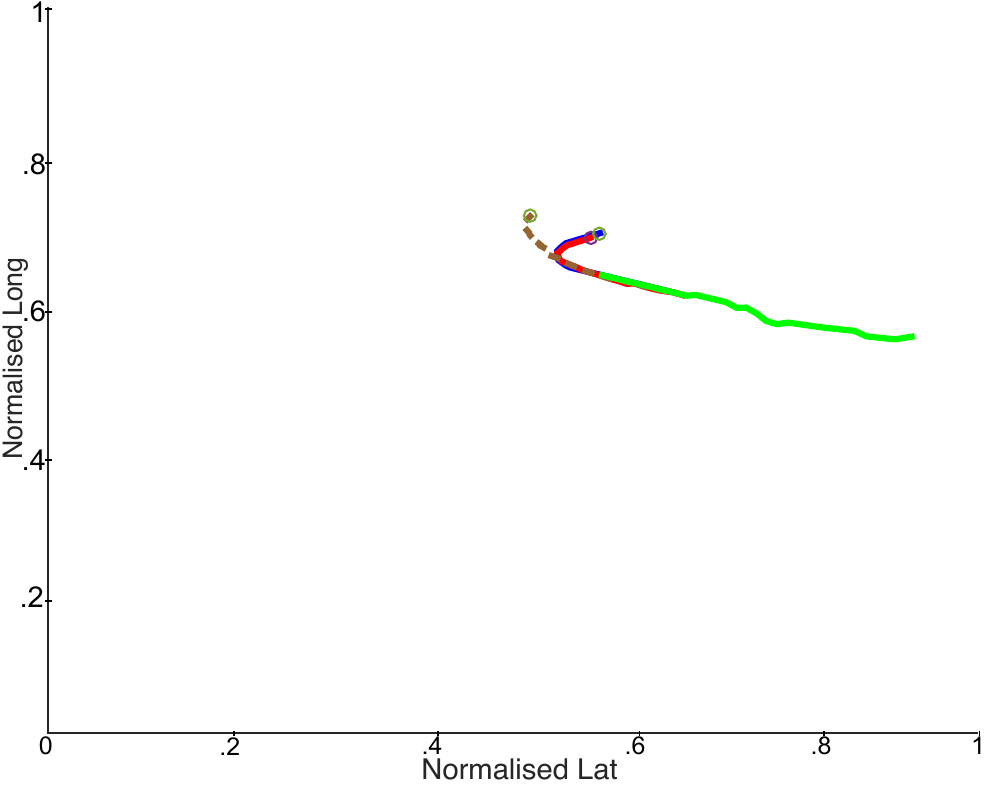}}
\subfigure[Latitude vs Altitude]{\includegraphics[width = .38\textwidth]{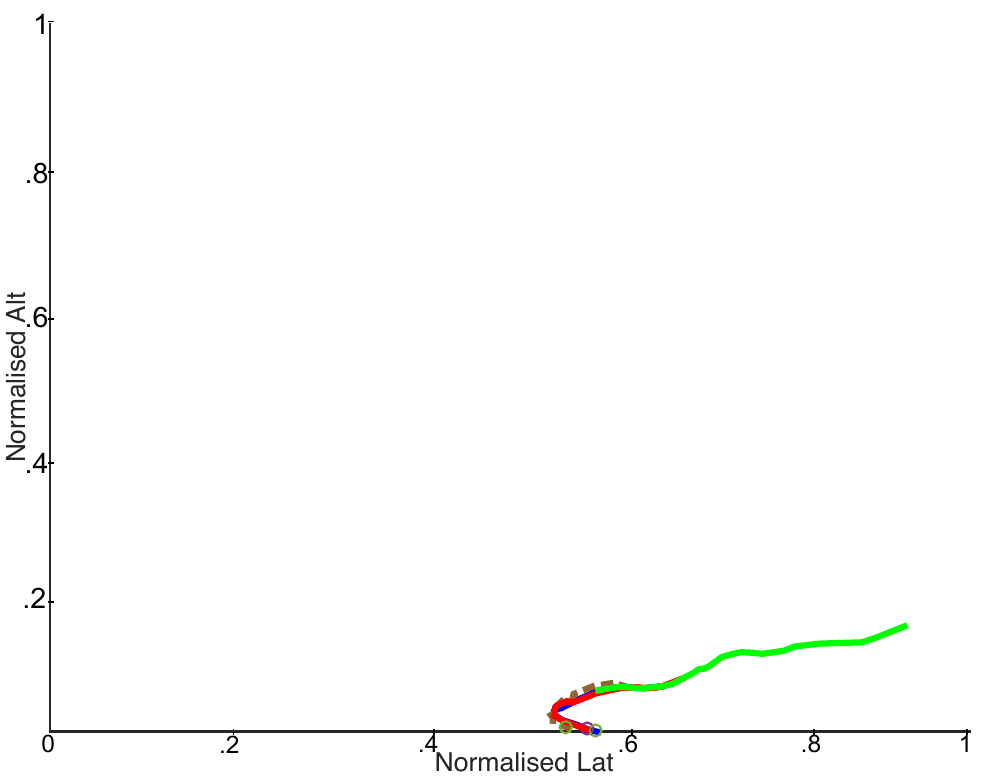}}

\subfigure[Latitude vs Longitude]{\includegraphics[width = .38 \textwidth]{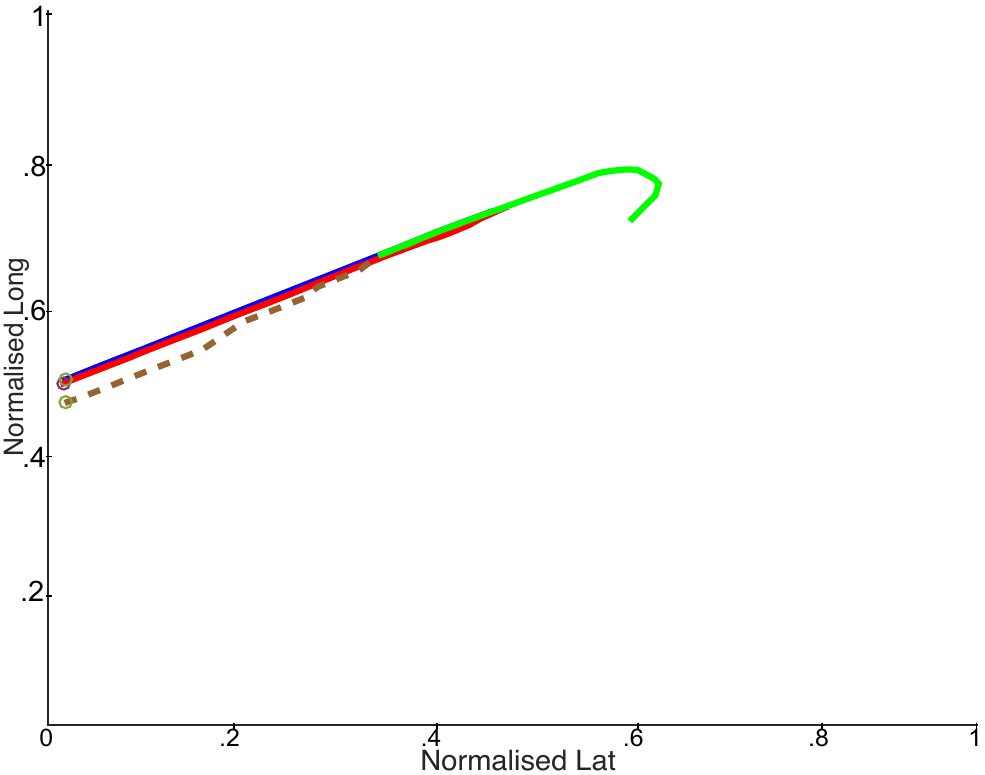}}
\subfigure[Latitude vs Altitude]{\includegraphics[width = .38\textwidth]{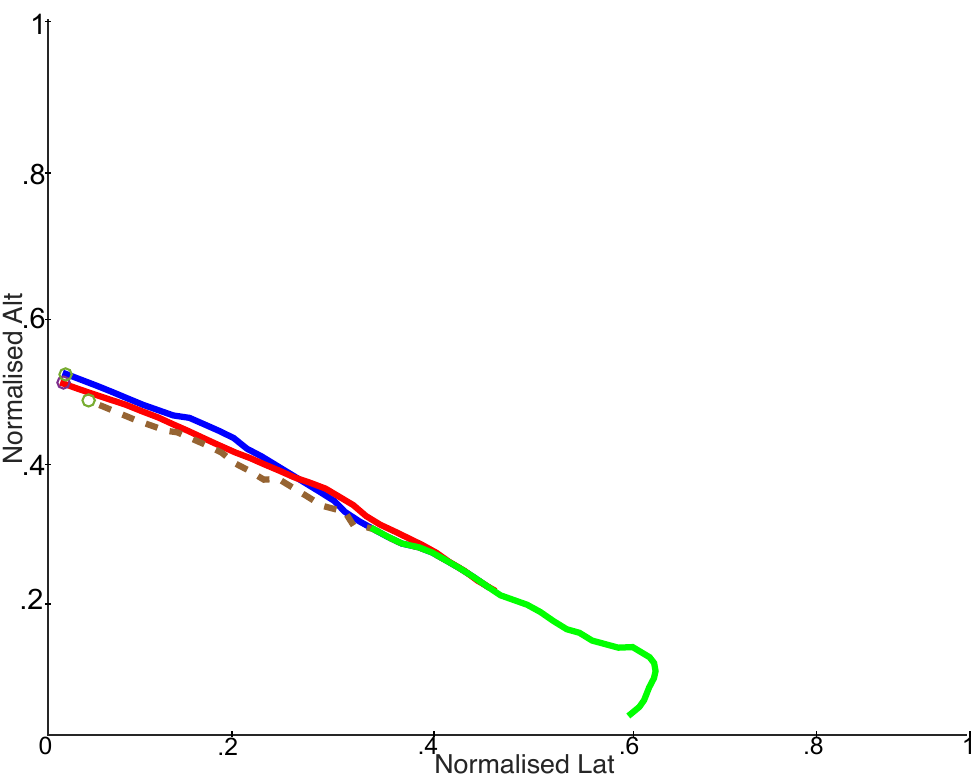}}

\end{center}
\caption{Qualitative results under storm conditions: Each row represents a particular example. Given (in green), Ground Truth (in Blue) and Predicted trajectories from $\textbf{TMN}$ model (in red), from \textbf{HMM} model (in brown). }
\label{fig:AirTrj_Storm}
\end{figure}
Referring to the results presented in Fig. \ref{fig:AirTrj_Storm}, it is evident that the non uniform nature of the air traffic in the stormy weather conditions is effectively modelled by the proposed approach. Even with the weather information, the \textbf{HMM} fails to effectively exploit this data and generates erroneous trajectories. The \textbf{TMN} model anticipates the uneven nature by looking at the recent history, while also mapping how correlated trajectories have behaved over the long term history. \par 
Finally, we would like to emphasise that for the \textbf{TMN} model, even in storm conditions, the altitude error is within the +-100ft altimeter tolerances provided to private pilots \cite{aip}. 

\subsection{Experiment 2: Pedestrian trajectory prediction}
\label{sec:ex_2}
In this experiment we considered 3 months worth of trajectories from the Edinburgh Informatics Forum database \citep{EIF}. We train our model on 60,000 trajectories and test on 12,000 trajectories. For all the models we observed the trajectory for 30 frames and predicted the trajectory for the next 30 frames. In this experiment the trajectories are represented as 2 dimensional data where each point $\bf{x_t^{i}}$ of the input sequence $X^{i}=[\mathbf{x_{1}^{i}}, \mathbf{x_{2}^{i}}, \ldots, \mathbf{x_{T}^{i}}]$ is represented as,
\begin{equation}
\mathbf{x_t^{i}}=\left (
                \begin{array}{ll}
                  x^{i}_{t}\\
                  y^{i}_{t}\\
                \end{array}
              \right ).
\end{equation}

\par For comparison we implemented Soft+hard wired attention (\textbf{SH-Atn}) model from \cite{our_WACV}, the Social LSTM (\textbf{So-LSTM}) model given in \cite{social_LSTM} and Dynamic Memory Networks (\textbf{DMN}) given in \cite{askMeAnything}. For the \textbf{So-LSTM} model, a local neighbourhood of size 32px was considered and the embedding dimension of all the LSTMs are set to 180 as recommended in \cite{social_LSTM}. For \cite{our_WACV} we considered a neighbourhood size of 10 in the left, right and front directions and an embedding size of 300 hidden units. For \textbf{DMN} and \textbf{TMN} models we use the same experimental settings given the in previous experiment.
Similar to \citep{our_WACV, social_LSTM} we report prediction accuracy with the following 3 error metrics. We are predicting a trajectory for the period from $t= T^{obs}+1$ to $T^{pred}$ while observing the same trajectory from $t= 1$ to $T^{obs}+1$. Let $n$ be the number of trajectories in the testing set, $\hat{X^{i}_{t}}$ be the predicted position for the trajectory $i$ at the $t^{th}$ time instance, and $X^{i}_{t}$ be the respective observed positions then:
\begin{enumerate}
\item \textit{Average displacement error (ADE): } 
\begin{equation}
ADE=\cfrac{\sum\limits_{i=1}^{n}\sum\limits_{t=T^{obs}+1}^{T^{pred}}(\hat{X^{i}_{t}}-X^{i}_{t})^2}{n(T^{pred}-(T^{obs}+1))} .
\end{equation}
\item \textit{Final displacement error (FDE) :} 
\begin{equation}
FDE=\cfrac{\sum\limits_{i=1}^{n}\sqrt{(\hat{X}^{i}_{T^{pred}}-X^{i}_{T^{pred}})^2}}{n} .
\end{equation}
\item \textit{Average non-linear displacement error (n-ADE): } The average displacement error for the non-linear regions of the trajectory.
\begin{equation}
n-ADE=\cfrac{\sum\limits_{i=1}^{n}\sum\limits_{t=T^{obs}+1}^{T^{pred}}I(\hat{X^{i}_{t}})(\hat{X^{i}_{t}}-X^{i}_{t})^2}{\sum\limits_{i=1}^{n}\sum\limits_{t=T^{obs}+1}^{T^{pred}}I(\hat{X^{i}_{t}})} ,
\end{equation}
where,
\begin{equation}
I(\hat{X^{i}_{t}}) = \begin{cases} 1 &\mbox{if }  \cfrac{d^2y^{i}_{t}}{d(x^{i}_{t})^2} \neq 0 .\\ 
0 & o. w \end{cases}
\end{equation}
\end{enumerate} 

\begin{table}[!h]
  \centering
  \begin{tabular}{|c|c|c|c|c|}
    \hline
   Metric  & SH-Atn & So-LSTM & DMN & TMN \\
    \hline
   ADE & 1.066 & 1.843 & 1.798 & \textbf{1.051} \\
  
    \hline\hline
    FDE & 1.551 & 2.421 & 2.276 & \textbf{1.398} \\

       \hline\hline
    n-ADE& 1.021 & 1.988 & 1.456 & \textbf{0.987} \\
   
    \hline
  \end{tabular}
  \caption{Quantitative results for pedestrian trajectory prediction. In all the methods the forecast trajectories are of length 30 frames. The first row represents the average displacement error (ADE), the second row shows the final displacement error (FDE) and the third row shows the average non-linear displacement error (n-ADE).}
  \label{tab:tab_2}
\end{table}

As shown in Table \ref{tab:tab_2}, the proposed model outperforms the \textbf{SH-Atn} model, \textbf{So-LSTM} model and \textbf{DMN} in all three error metrics. The dataset is considered quite challenging as there are multiple source and sink locations, different crowd motion patterns are present, and motion paths are heavily crowded. 

\begin{figure}[!h]
\begin{center}
\subfigure[]{\includegraphics[width = .3 \textwidth]{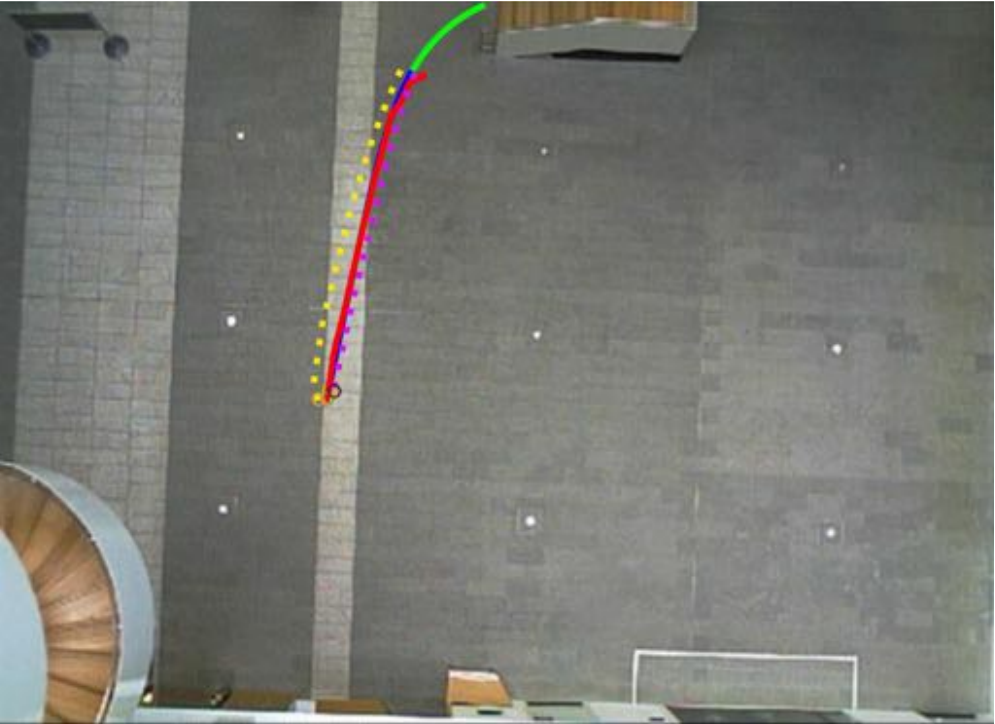}}
\subfigure[]{\includegraphics[width = .3\textwidth]{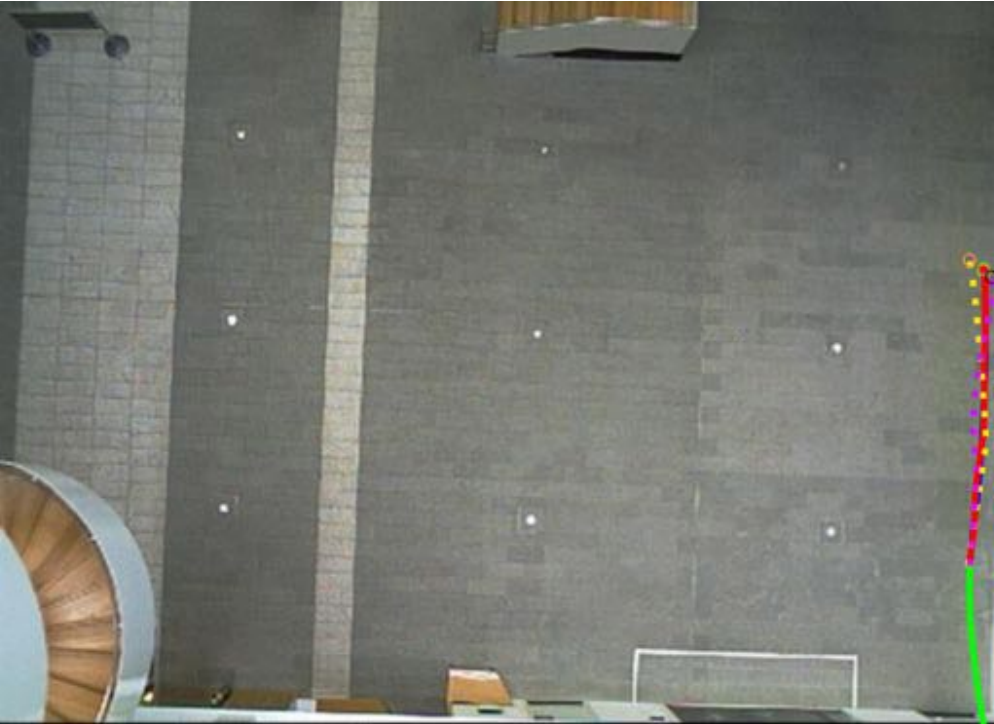}}
\subfigure[]{\includegraphics[width = .3\textwidth]{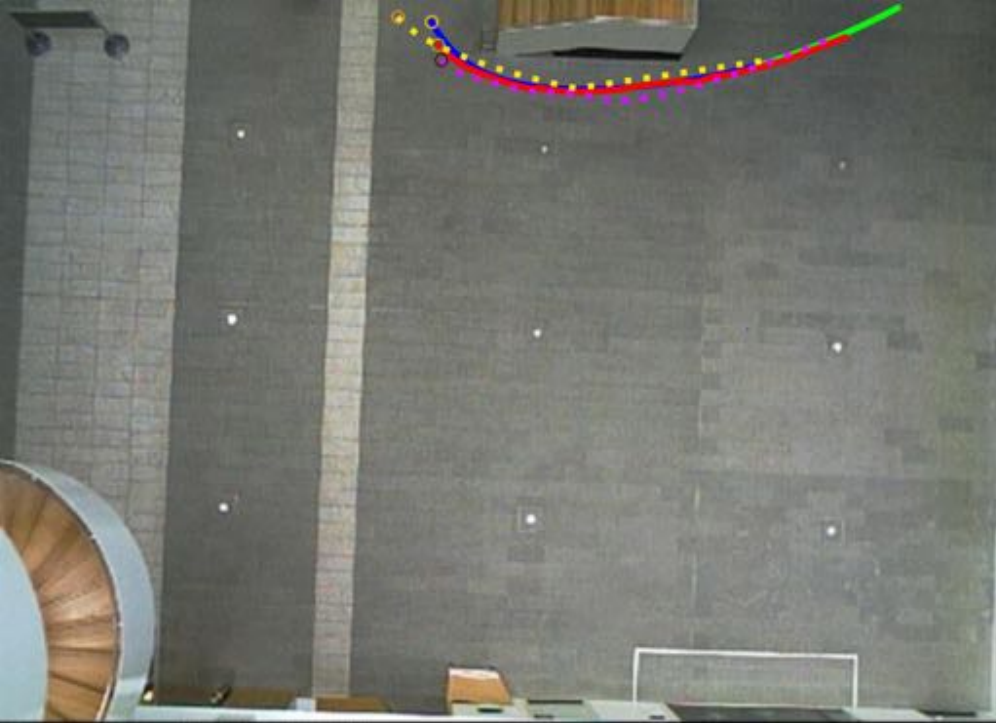}}

\subfigure[]{\includegraphics[width = .3 \textwidth]{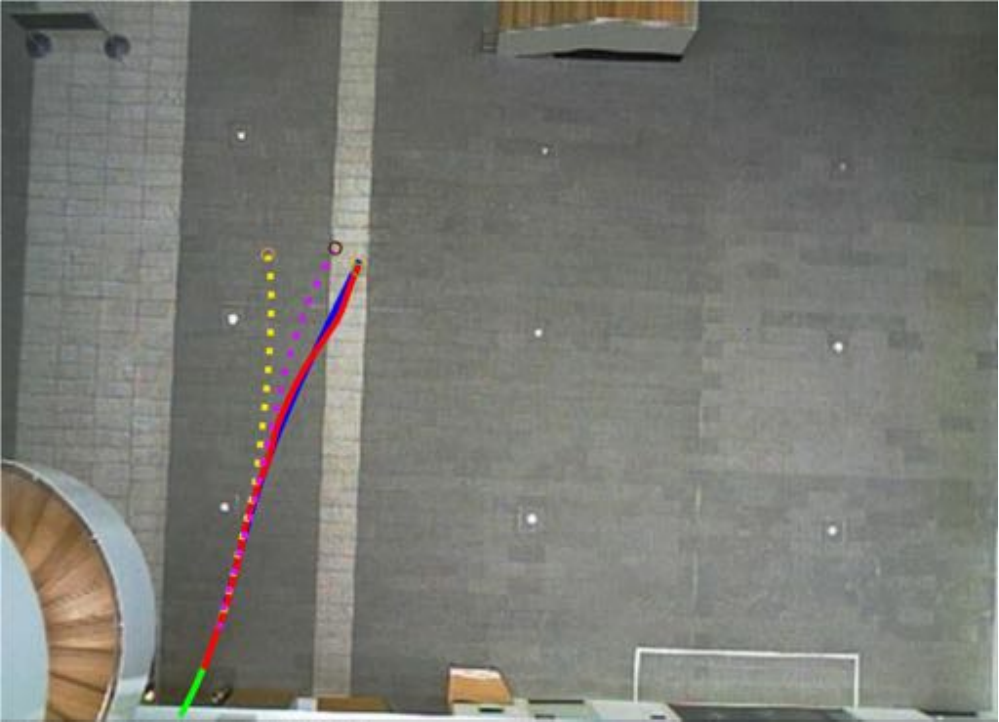}}
\subfigure[]{\includegraphics[width = .3\textwidth]{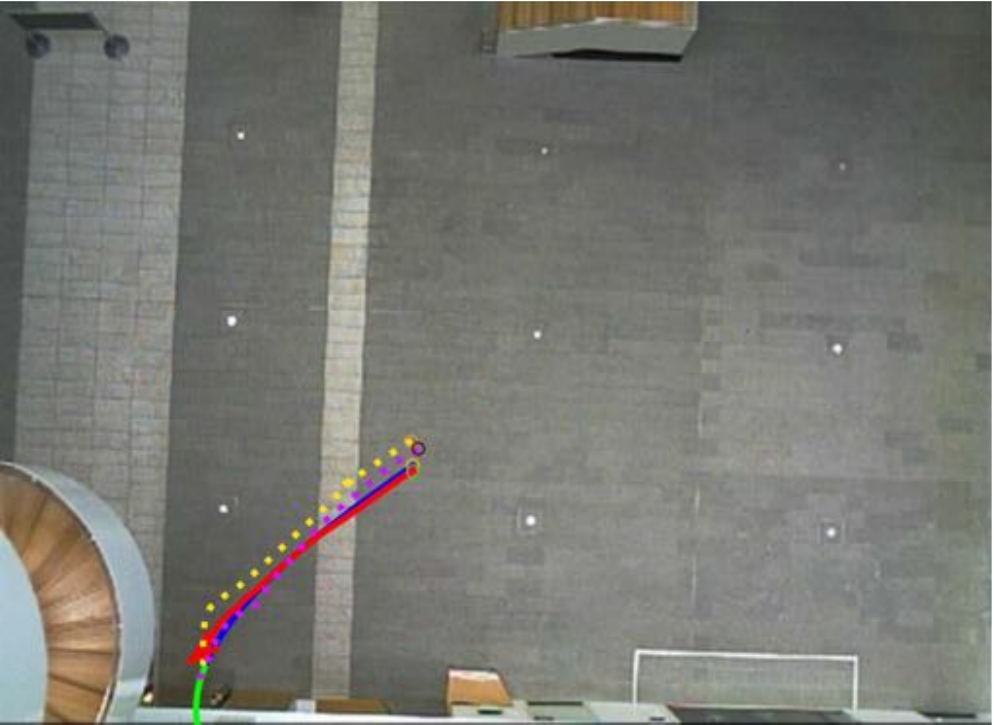}}
\subfigure[]{\includegraphics[width = .3\textwidth]{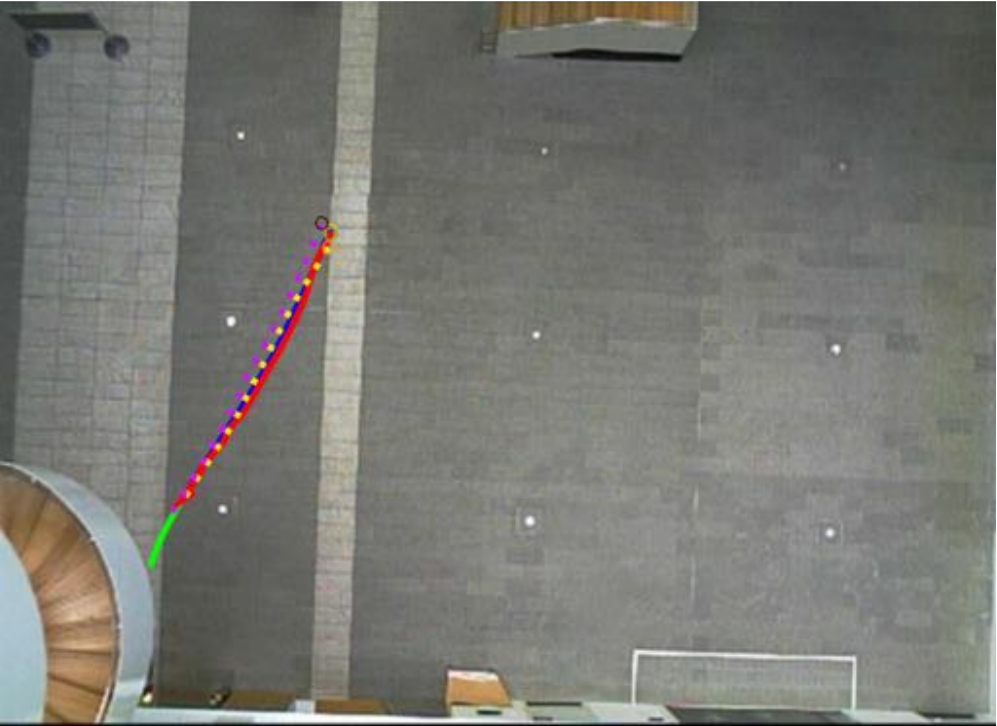}}

\subfigure[]{\includegraphics[width = .3 \textwidth]{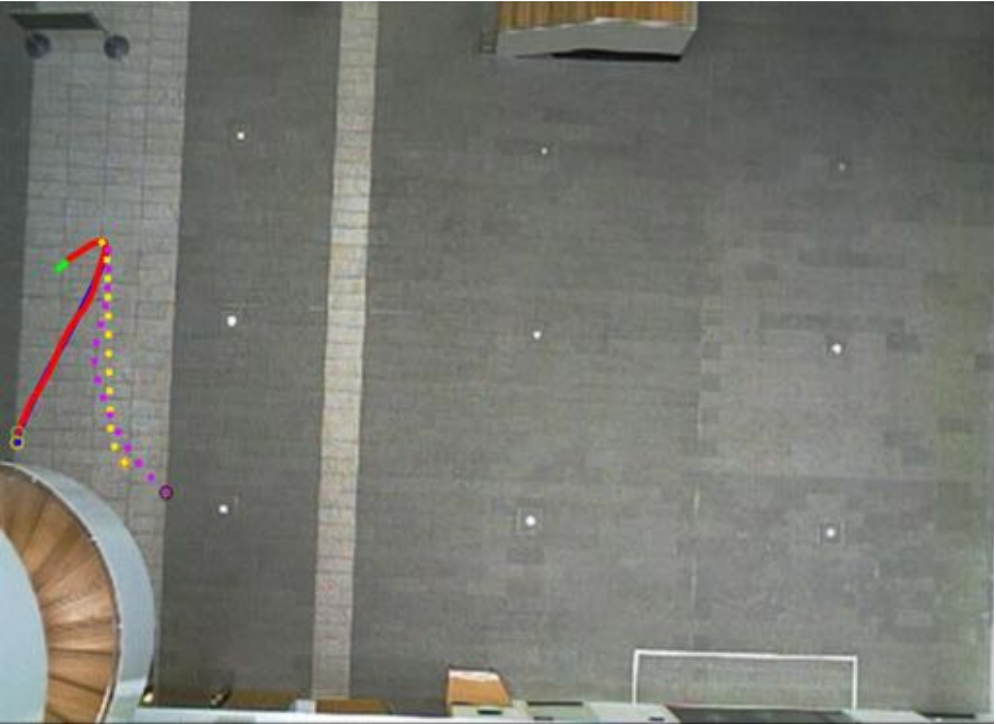}}
\subfigure[]{\includegraphics[width = .3\textwidth]{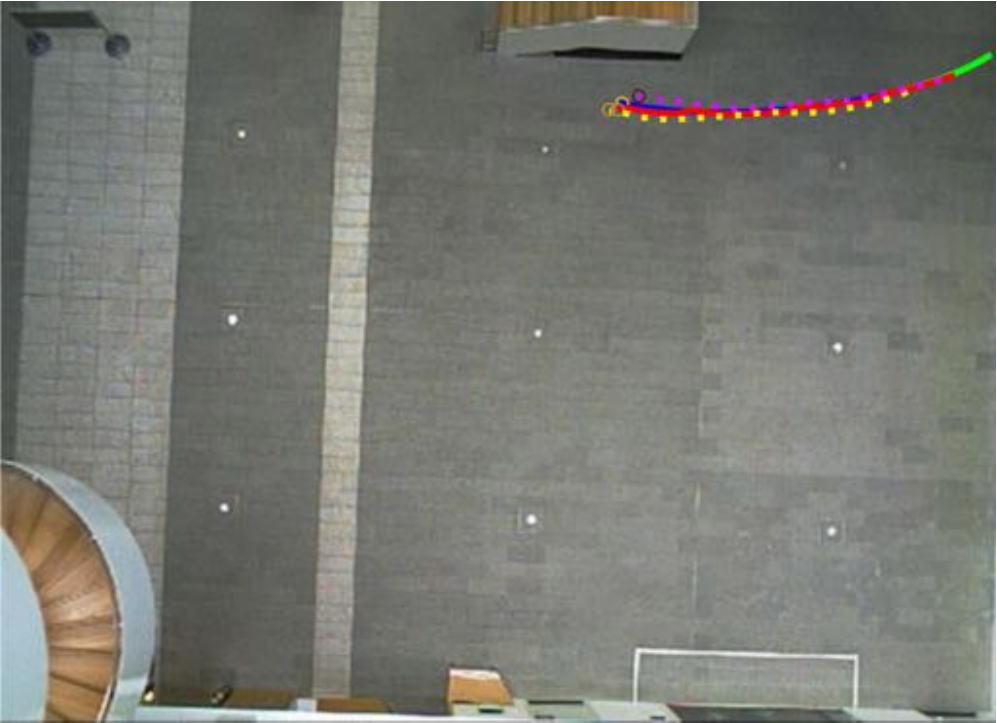}}
\subfigure[]{\includegraphics[width = .3\textwidth]{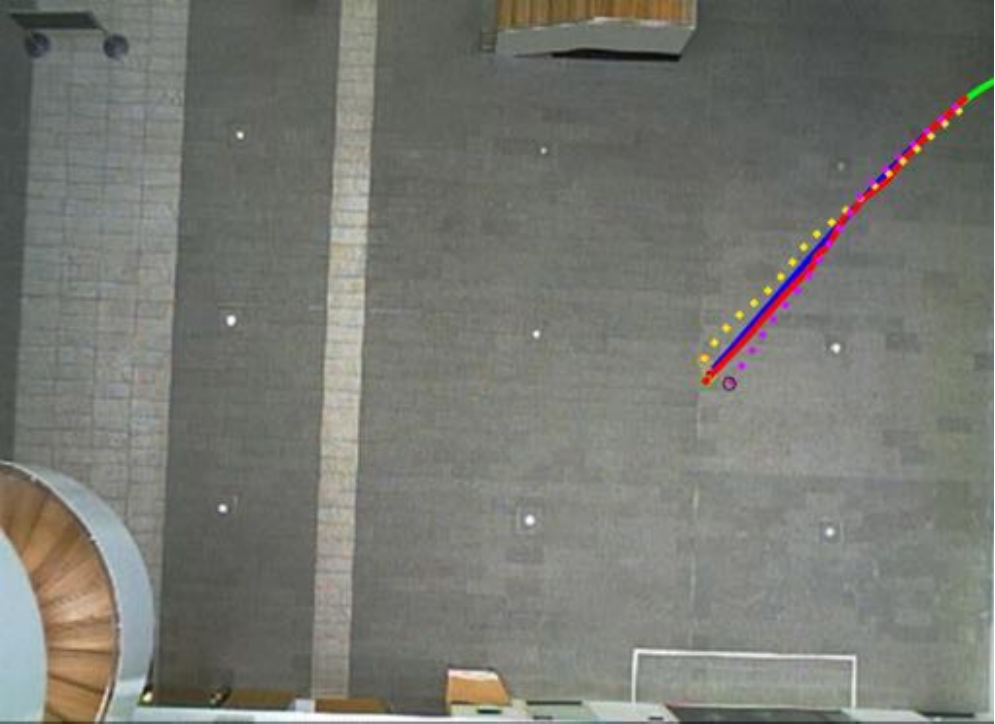}}

\subfigure[]{\includegraphics[width = .3 \textwidth]{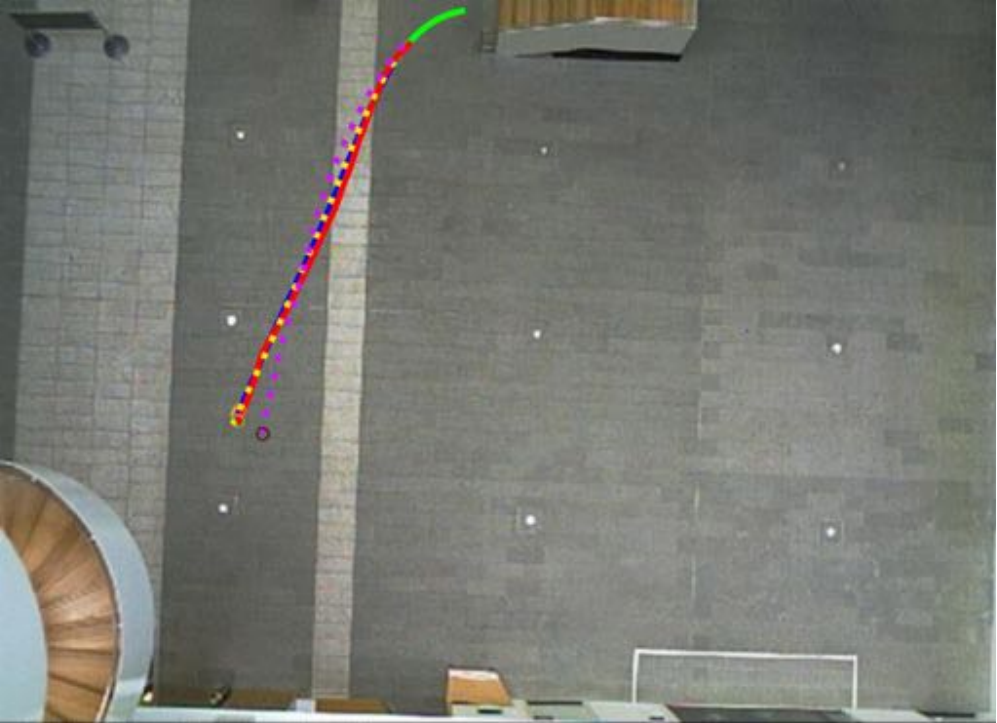}}
\subfigure[]{\includegraphics[width = .3\textwidth]{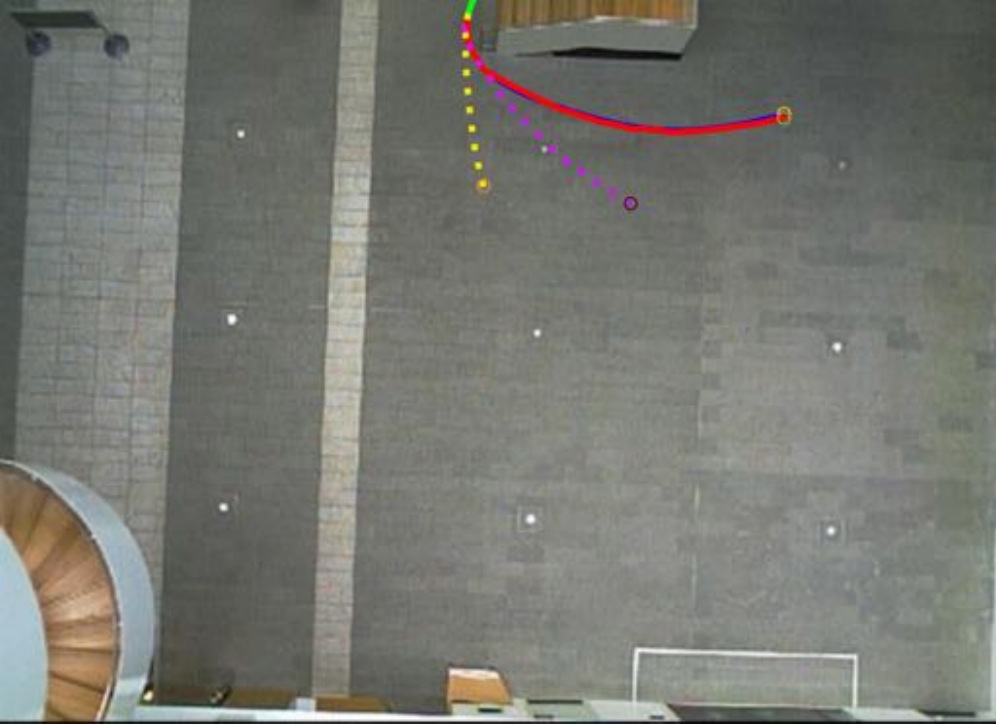}}
\subfigure[]{\includegraphics[width = .3\textwidth]{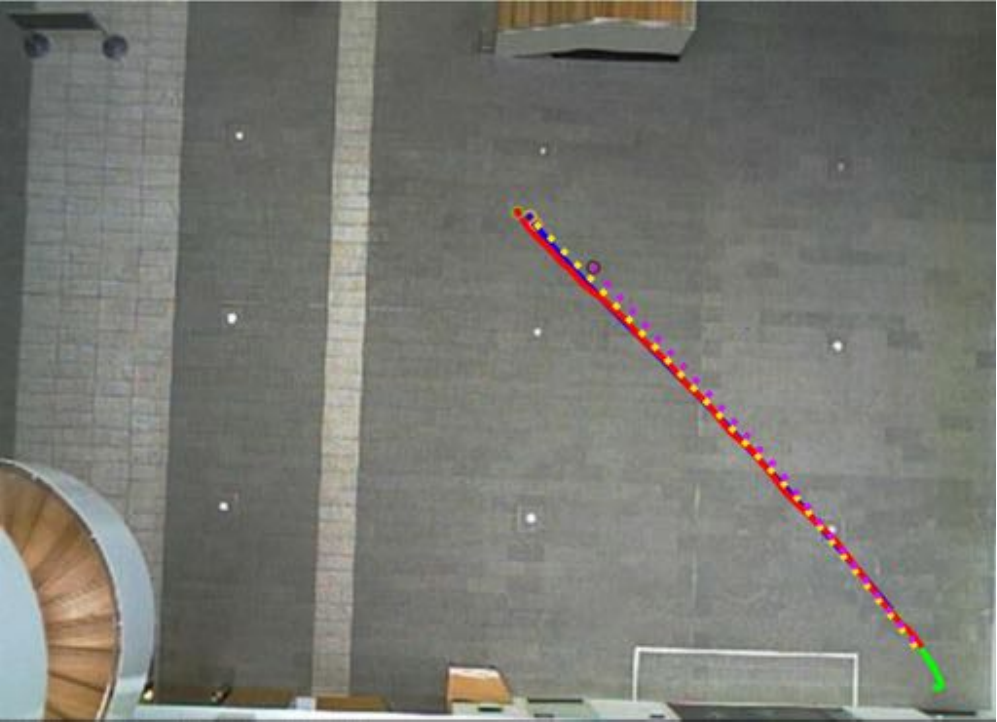}}

\end{center}
\caption{Qualitative results: Given (in green), Ground Truth (in Blue) and Predicted trajectories from $\textbf{TMN}$ model (in red), from \textbf{DMN} model (in yellow), from \textbf{SH-Atn} model (in purple). }
\label{fig:fig4}
\end{figure}

The \textbf{So-LSTM} model has the lowest accuracy, as its attention mechanism is limited to the immediately preceding state of the neighbourhood. The \textbf{DMN} model considers the short term history of the entire trajectory, leading to improved performance compared to \textbf{So-LSTM}. By incorporating local neighbourhood history, \textbf{SH-Atn} is able to further improve on performance. However, despite not explicitly modelling the neighbourhood as done by the \textbf{SH-Atn} and \textbf{So-LSTM} models, the proposed approach is able to outperform these state-of-the-art techniques. The \textbf{TMN} approach extends the notion of neighbourhood history to consider longer temporal dependencies. It not only considers the short term environment context, where temporally adjacent trajectories are, but also considers how similar trajectories have behaved over the long term history. This is further demonstrated by the results presented in Fig. \ref{fig:fig4}.

Fig. \ref{fig:fig4} shows prediction results for the \textbf{SH-Atn} model,  \textbf{DMN} model and our model $\textbf{TMN}$ on the EIF trajectory dataset. From the examples shown it is evident how different modes of human motion are captured and represented through the proposed memory module. For example in Fig. \ref{fig:fig4} (g) and Fig. \ref{fig:fig4} (k) the pedestrians exhibit a sudden change in motion which all baseline models fail to capture. But the proposed model has successfully anticipated that motion through recalling similar historic behaviour. 

\begin{figure*}[t]
\begin{center}														    
\includegraphics[width = \textwidth]{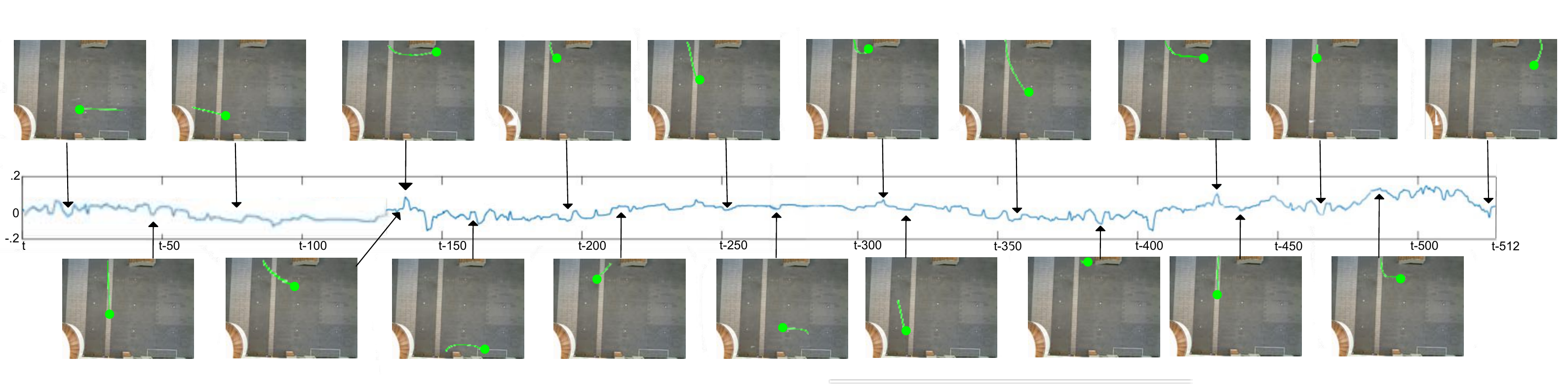}
\caption{Distribution of memory activations from the first layer of the \textbf{TMN} module for the first data point of Fig. \ref{fig:fig4} (k). This layer contains 512 memory which are denoted $t$ to $t-512$ indicating the history that has been observed. For different peaks and valleys of the memory activations we also show what the memory has seen at those particular time steps. The model generates higher activations for trajectories that change the heading direction and perform a turn as in Fig. \ref{fig:fig4} (k), and activations closer to zero for cases where the pedestrian in demonstrating different behaviour. We effectively propagate this information from the first layer of the memory to the top most layer via combining the salient information in a hierarchical manner. }
\label{fig:mem_content}
\end{center}
\end{figure*}
 
Fig. \ref{fig:mem_content} shows the distribution of activations from the first layer of the proposed TMN module, for the first data point of the pedestrian trajectory given in Fig. \ref{fig:fig4} (k). As $p=512$, there exist 512 memory slots in this layer. We denote the current time as t, hence the memory slots range from $t$ to $t-512$ in the history. For different peaks and valleys in the memory activations, we show what the model has seen at that particular time step.

 The \textbf{TMN} provides higher responses for recent events as well as for similar trajectory patterns in the long term history. Considering both spatial locations as well as the velocity encoded by the spatial dispersion between the consecutive points, the model is able to anticipate that the pedestrian is more likely to change their heading, indicated by higher activations to similar trajectories that reside within the entire history captured by the memory module (see the activation peeks between $t-100$ to $t-200$ and $t-300$ to $t-500$).
 
The flat memory structure of the \textbf{DMN} doesn't have the ability to capture such long term dependencies due to the short term history dominance issues inherent with sequential LSTM architectures, which we further discuss and demonstrate in Sec. 5.3. With $p=180$, the memory of the \textbf{DMN} has seen similar pedestrian behaviour to that in Figure \ref{fig:mem_content} (k), but cannot identify the importance of those examples as the short term history dominates the output. This clearly verifies the importance of efficiently modelling dependencies with a hierarchical structure. It is not sufficient to just have a large (i.e. long history) memory module, the module also needs to effectively propagate relevant historic examples to the output module to generate better predictions.

\section{Discussion}

\subsection{Flexibility of the TMN framework}
We selected the two problem domains, aircraft trajectories (see Section \ref{sec:ex_1}) and pedestrian trajectories (see Section \ref{sec:ex_2}), to highlight the flexibility of the \textbf{TMN} model. Even though both domains consist of trajectories, the structure and dynamics of the domains show vast differences.

Aircraft trajectories capture 3 dimensional motion patterns, with rigid structure, following a particular flight schedule throughout the year. We do not offer any information to the model on to the specifics of the flight dynamics or manoeuvres. The model has to learn those characteristics directly from data by comparing and contrasting the temporal evolution of the large number of trajectories. Furthermore, there exist different trajectory patterns in take off, landing and cruising of flights; and between different aircraft types including commercial airliners, helicopters, surveillance flights, etc. We do not filter any of those categories from the data, and feed them all together to the model. The \textbf{TMN} model successfully understands these specifics through querying from the long term history. 

In contrast to the structured pattern of the aircraft trajectories, pedestrian trajectories are highly unstructured. The model has to identify that pedestrians vary their velocity and heading directions more frequently and rapidly, often in response to other very recent observations, compared to the aircraft trajectories.  

We do not provide any supervision to our model or change the structure of the network between the two experiments. Based on the experimental results in Tab. \ref{tab:tab_1} and Tab. \ref{tab:tab_2} the proposed model successfully identifies those differences and demonstrates flexibility in adaptation to the different conditions. 

To further demonstrate the flexibility of the proposed model we use different trajectory lengths for the two experiments.

In the case of aircraft trajectories, the TAAATS samples aircraft position at a rate of 1 Hz. As we are using data only from the SEQ region of Australia, this gives us relatively short trajectory lengths, but with higher variability between the data points. In contrast pedestrian trajectories are sampled at 25 fps giving us more lengthy trajectories and less variability between the data points.  Due to these differences in capture rate and the typical length of trajectories in each dataset, we use different length trajectories when predicting future behaviour: using 25 frames of data to predict the next 25 frames for aircraft trajectories, and using 30 frames to predict the next 30 for pedestrian motion prediction.

Results presented in Sections \ref{sec:ex_1} and \ref{sec:ex_2} show how the proposed \textbf{TMN} model is able to adapt to these changes without explicit supervision. 

\subsection{Hardware and Implementation Details}
The TMN module doesn't require any special hardware such as GPUs to run and has 8.1M trainable parameters. We ran the test set of experiment 1 on a single core of an Intel Xeon E5-2680 2.50GHz CPU and the TMN algorithm was able to generate 1000 predicted trajectories with 50, 3 dimensional data points in each trajectory (i.e. using 25 observations to predict the next 25 data points) in 12.20 seconds.

\begin{figure}[!htpb]
\begin{center}
\subfigure[Memory Length vs Runtime]{\includegraphics[width = .48\textwidth]{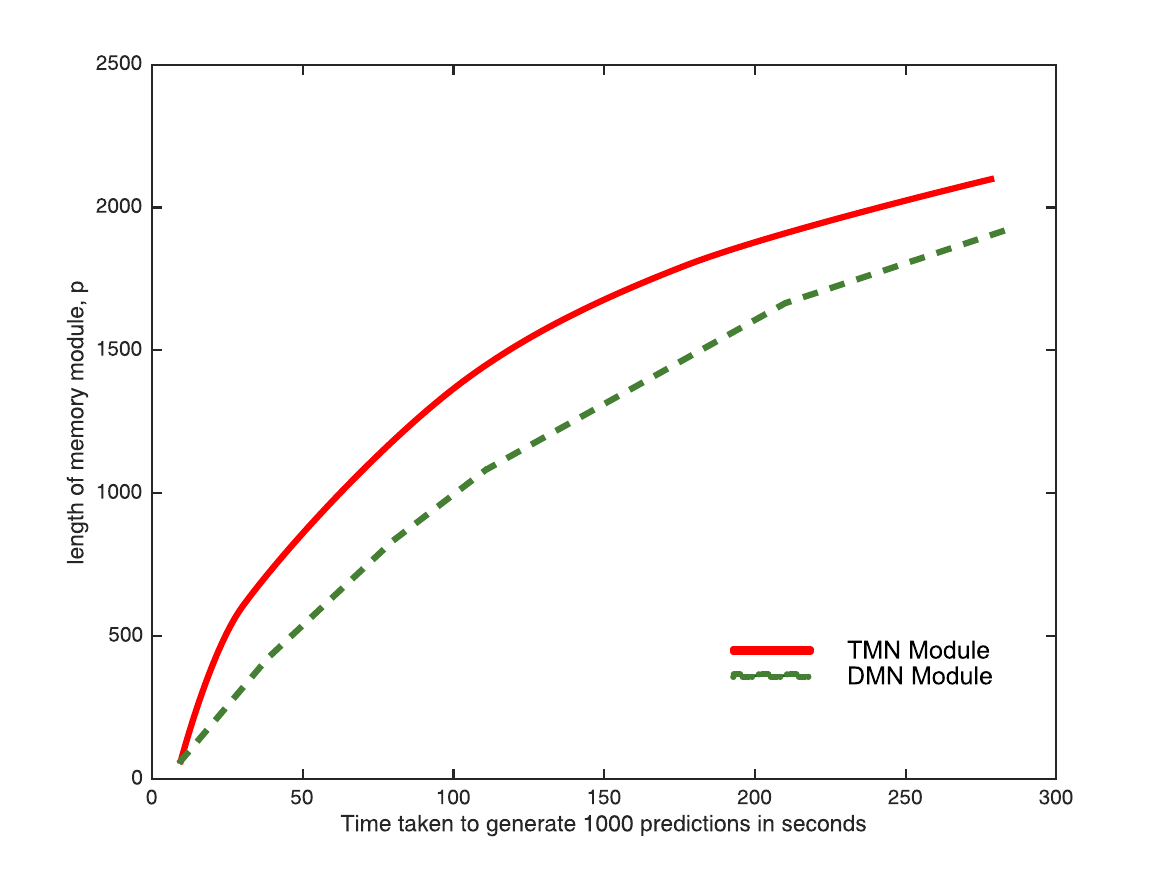}}
\subfigure[Sequence Length vs Runtime]{\includegraphics[width = .48\textwidth]{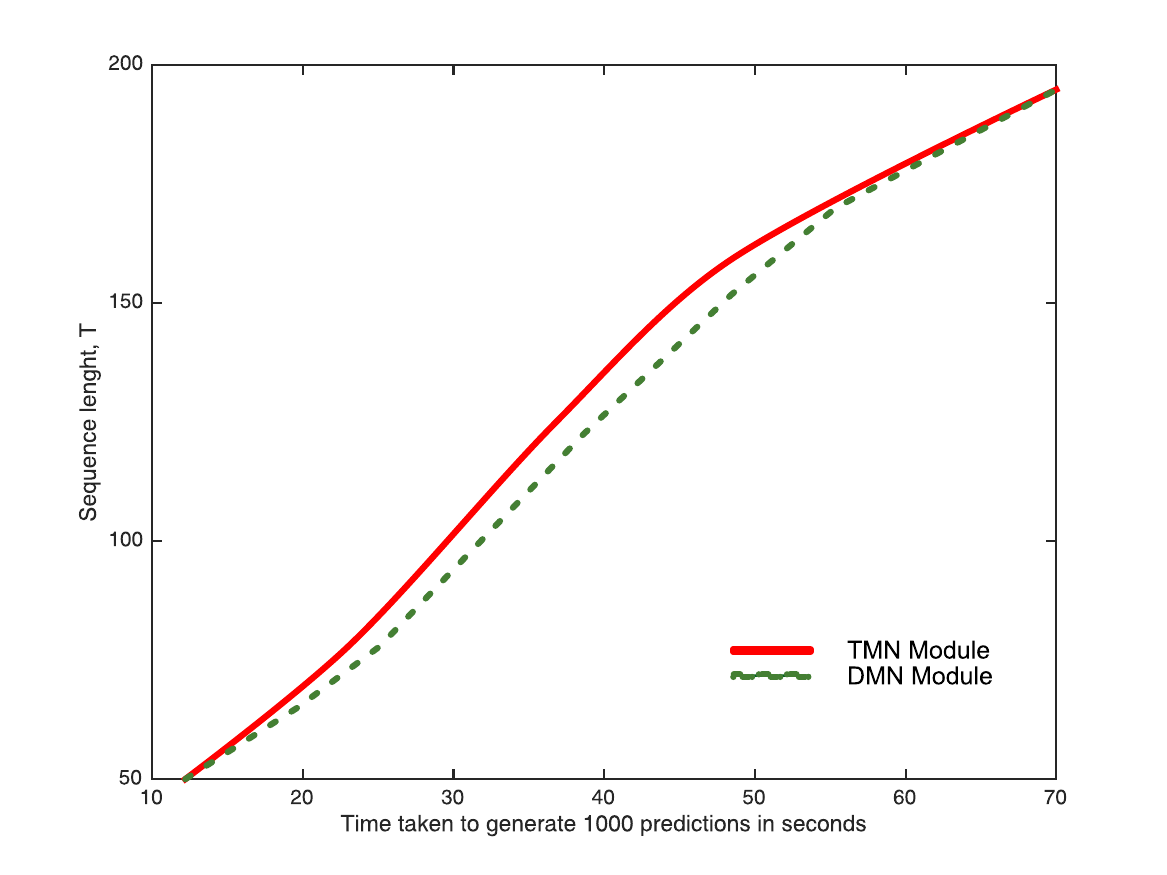}}
\caption{Evaluation of run times for different memory sizes and sequence lengths}
\label{fig:time_eval_P_T}
\end{center}
\end{figure}

In addition, we measured the time required to generate 1000 predicted trajectories for different lengths of the memory module, p, and different sequence lengths T. Results are presented in Fig. \ref{fig:time_eval_P_T} along with the respective evaluations for \textbf{DMN} module. The runtime grows approximately logarithmically against the memory size, as multiple S-LSTM layers are added to accommodate the increasing size of the memory component. There is an additional cost for using the proposed \textbf{TMN} over the \textbf{DMN} for a given memory size, however this cost is roughly consistent and does not vary greatly with the size of the memory. The time efficiency against sequence length exhibits a linear relationship, as sequence length only affects the encoding and decoding of the trajectories. We held the memory length p=512 constant for both modules in this experiment. The DMN module exhibits a similar distribution of runtimes and is slightly more time efficient. The efficiency gains are largely due to the update mechanism, as the sequential update is much simpler than the hierarchical update.

The implementation of the TMN module presented in this paper is completed using Keras \cite{chollet2017keras} with the Theano \cite{bergstra2010theano} backend, and only accepts equal length trajectory proportions for observation and prediction. Yet this is only a limitation in our implementation of the TMN, and the proposed algorithm is flexible and able to handle variable proportions in observed and predicted trajectories.

\subsection{Analysis of Memory Activations}
\label{Analysis_of_memory_activations}
In this section we analyse the memory activations of the \textbf{TMN} and \textbf{DMN} \cite{askMeAnything} models for various examples of the test set from Experiment 2 (see Section \ref{sec:ex_2}, Table \ref{tab:tab_2} and Figure \ref{fig:fig4}) in order to demonstrate that the sequential LSTM structure is biased towards recent history, and illustrate how the proposed model overcomes this via the hierarchical structure of the memory module.

In particular, we aim to show that the DMN model and the first layer of the proposed TMN approach have activations based heavily on the recent inputs to the model; while the last layer of the proposed approach has activations that are driven by the input itself rather than the recent history, due to it's ability to better capture the long term dependencies. We conduct this investigation with the pedestrian dataset as it is simpler to visualise. 

\subsubsection{Correlated activations from the first layer}

\begin{figure*}[t]
\begin{center}														    
\subfigure[]{\includegraphics[width = .3 \textwidth]{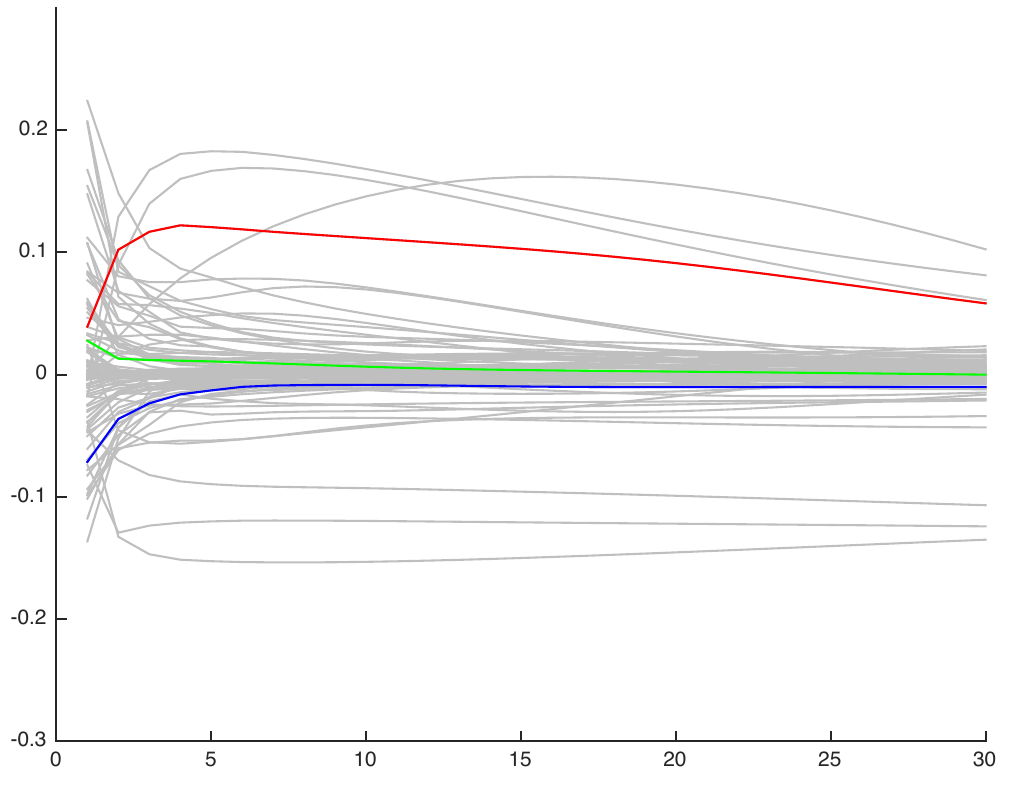}}
\subfigure[]{\includegraphics[width = .3\textwidth]{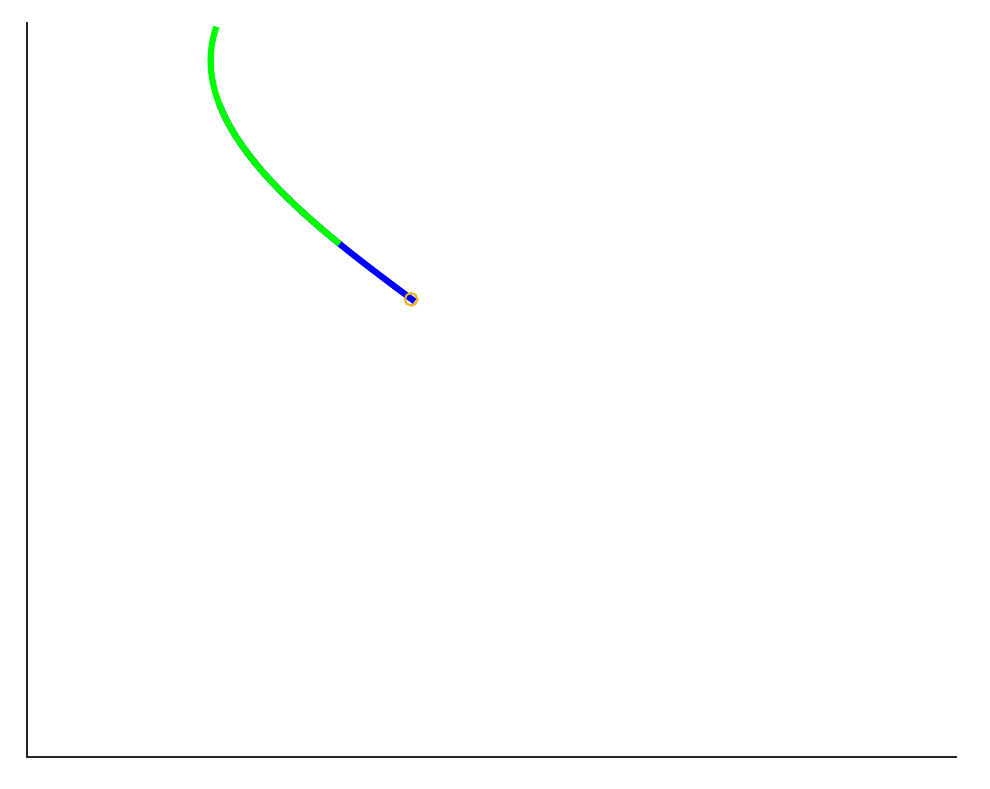}}
\subfigure[]{\includegraphics[width = .3\textwidth]{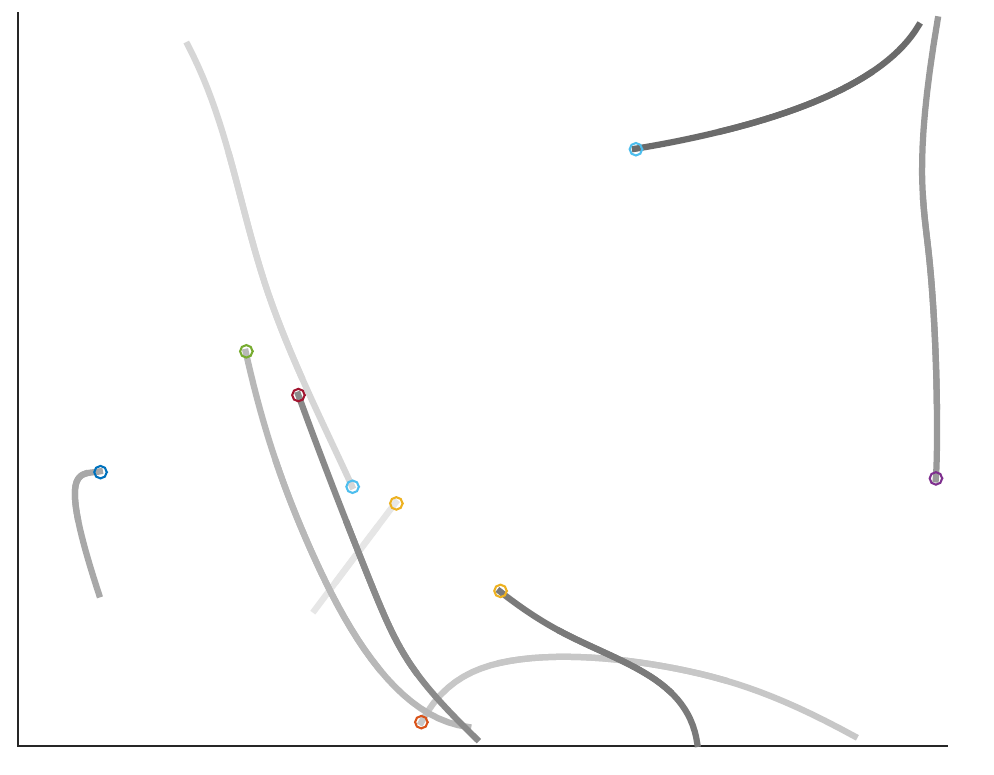}}

\subfigure[ ]{\includegraphics[width = .3 \textwidth]{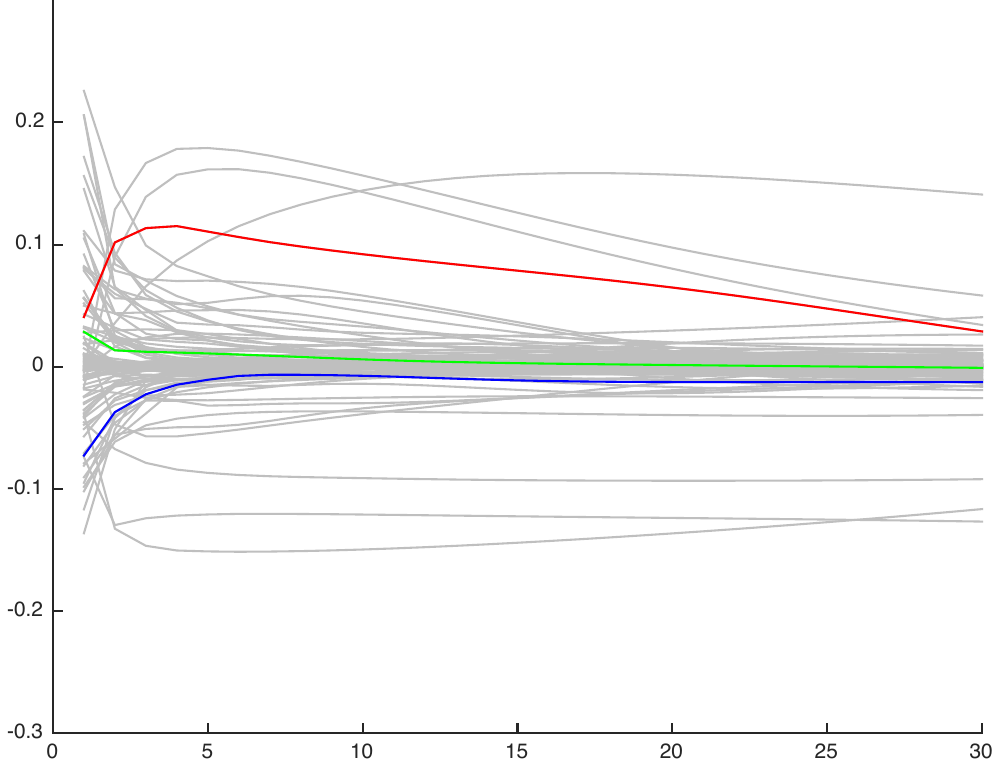}}
\subfigure[]{\includegraphics[width = .3\textwidth]{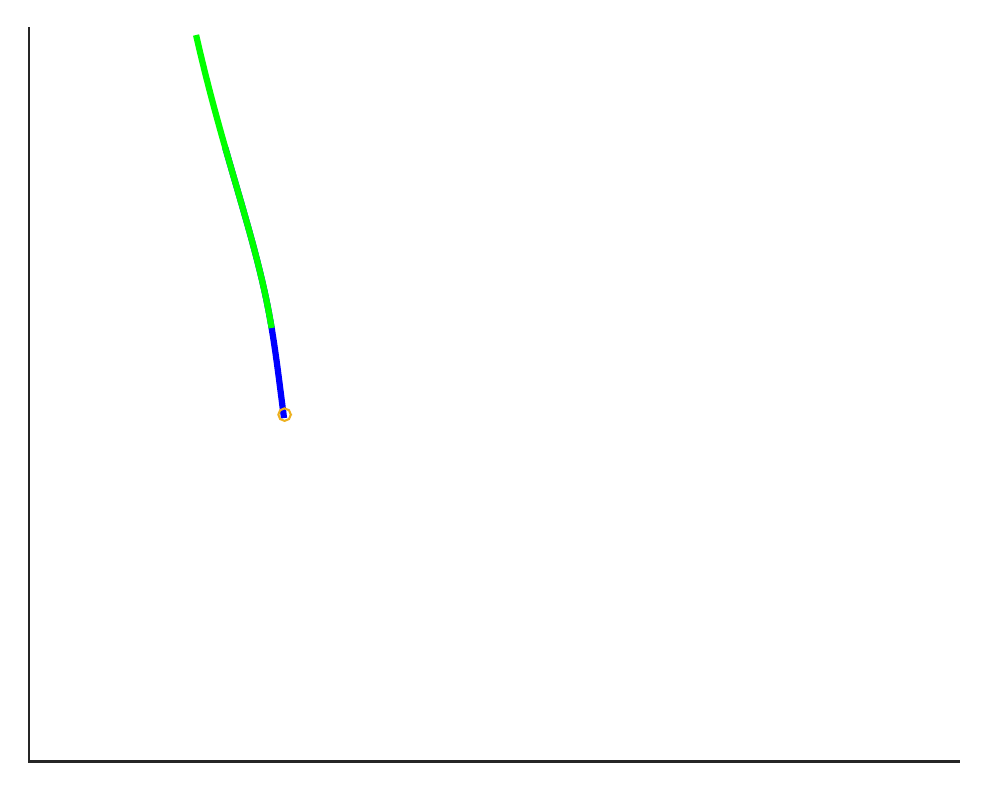}}
\subfigure[]{\includegraphics[width = .3\textwidth]{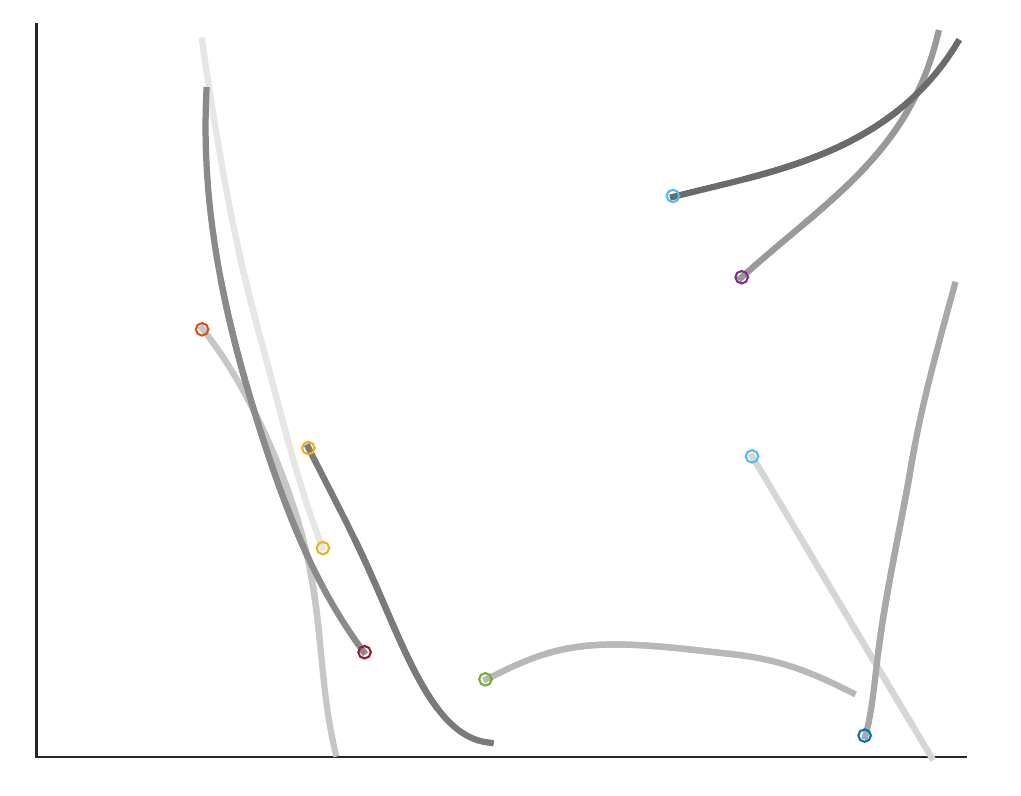}}

\subfigure[]{\includegraphics[width = .3 \textwidth]{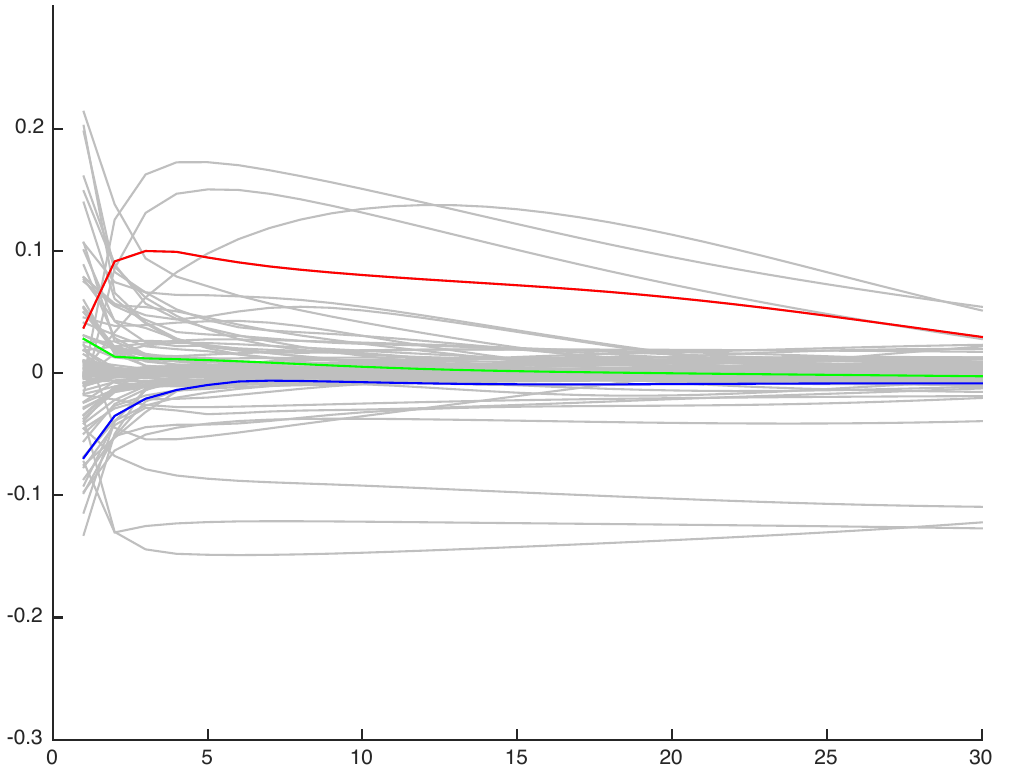}}
\subfigure[]{\includegraphics[width = .3\textwidth]{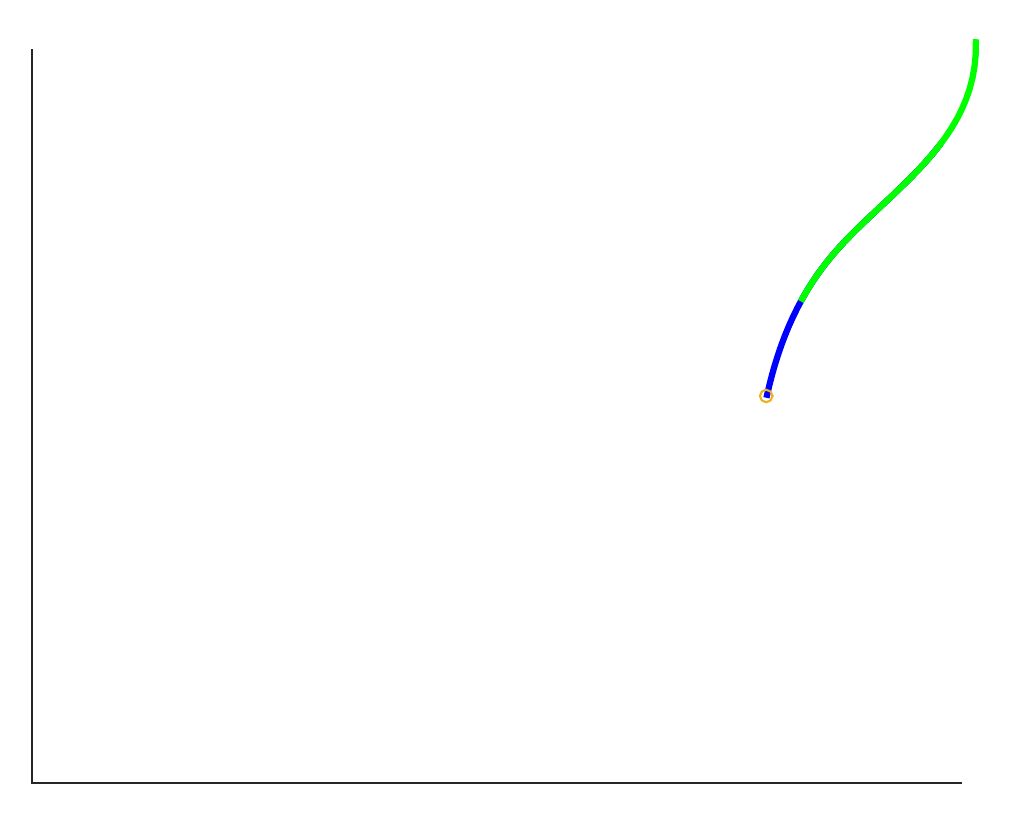}}
\subfigure[]{\includegraphics[width = .3\textwidth]{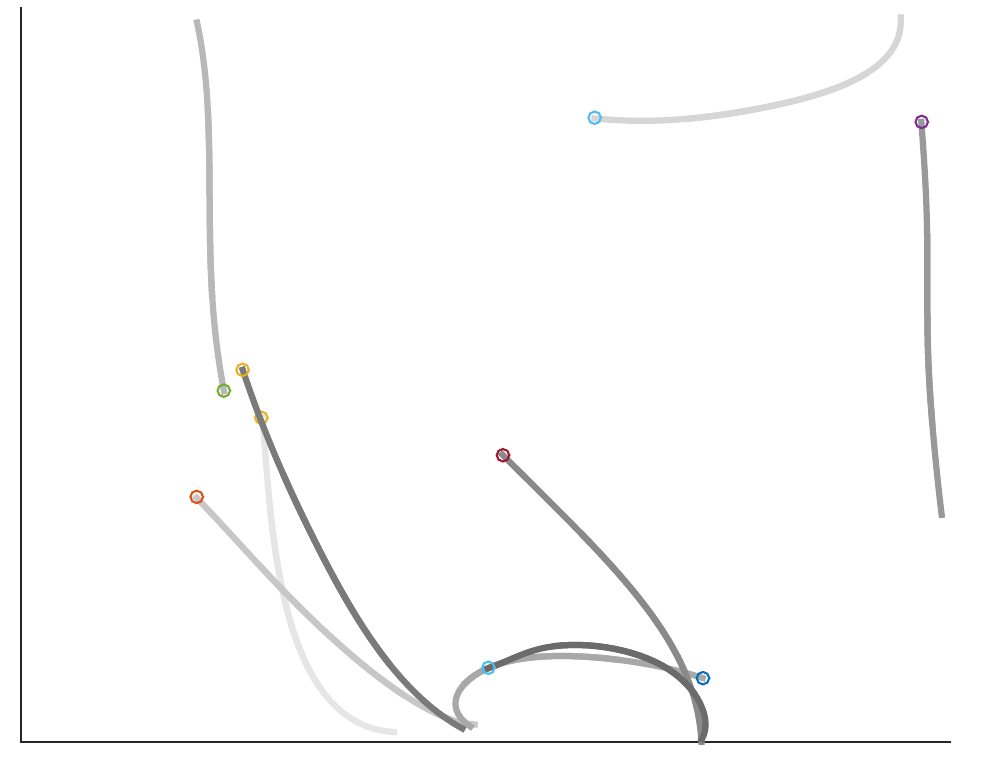}}

\subfigure[ ]{\includegraphics[width = .3 \textwidth]{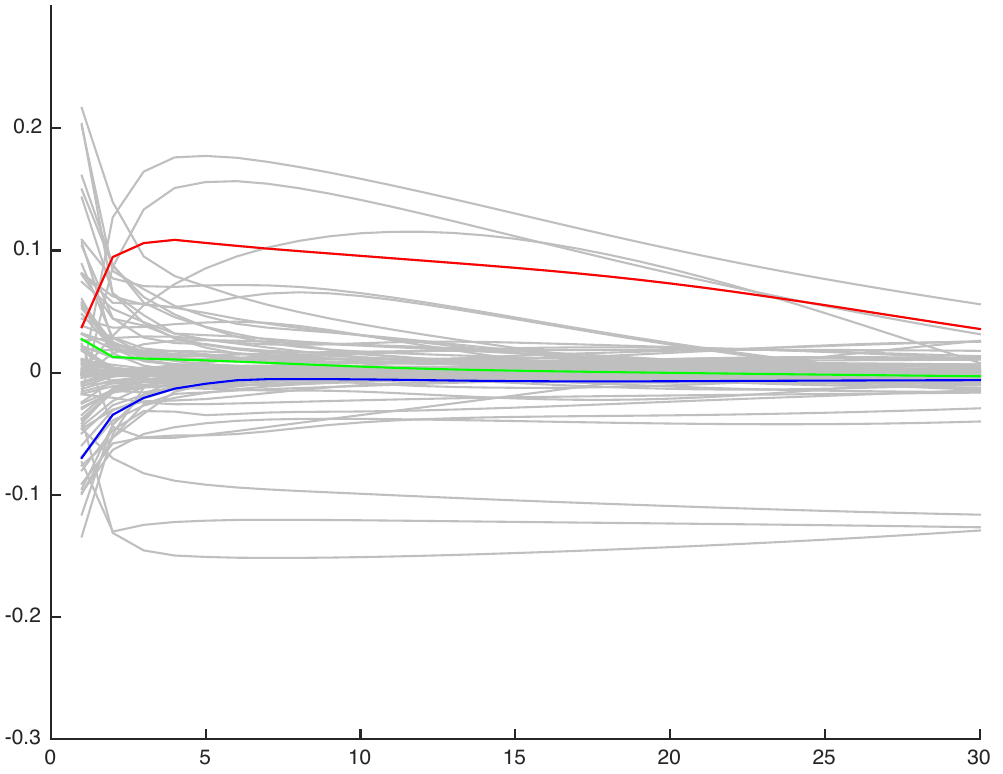}}
\subfigure[ ]{\includegraphics[width = .3\textwidth]{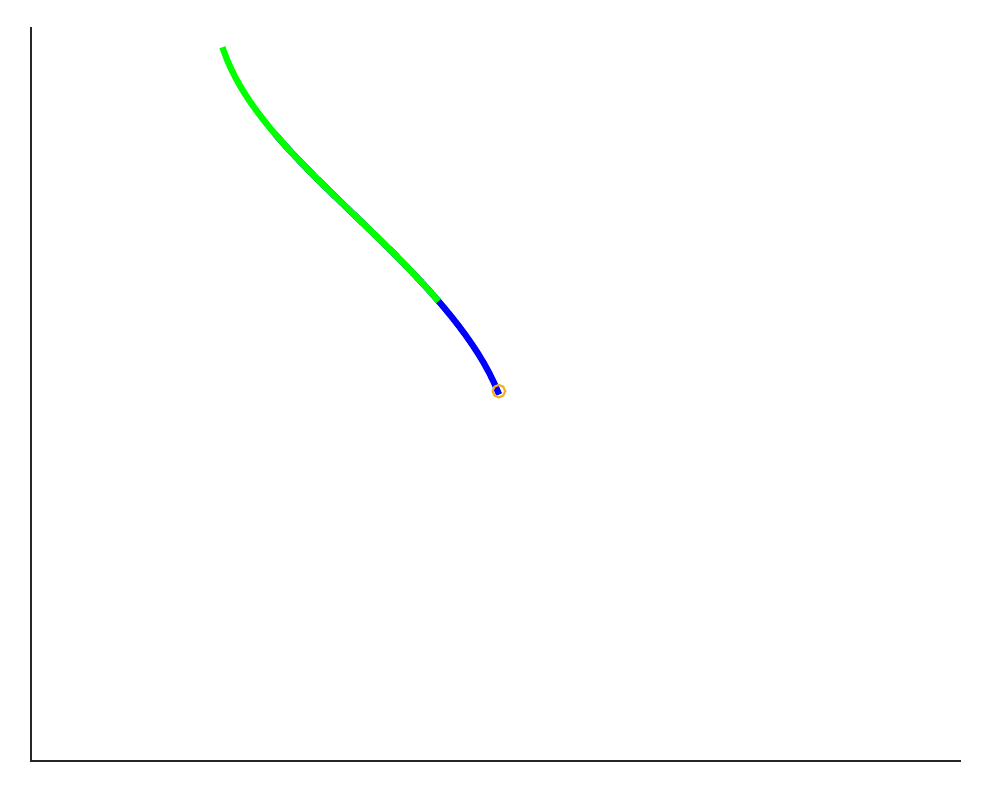}}
\subfigure[ ]{\includegraphics[width = .3\textwidth]{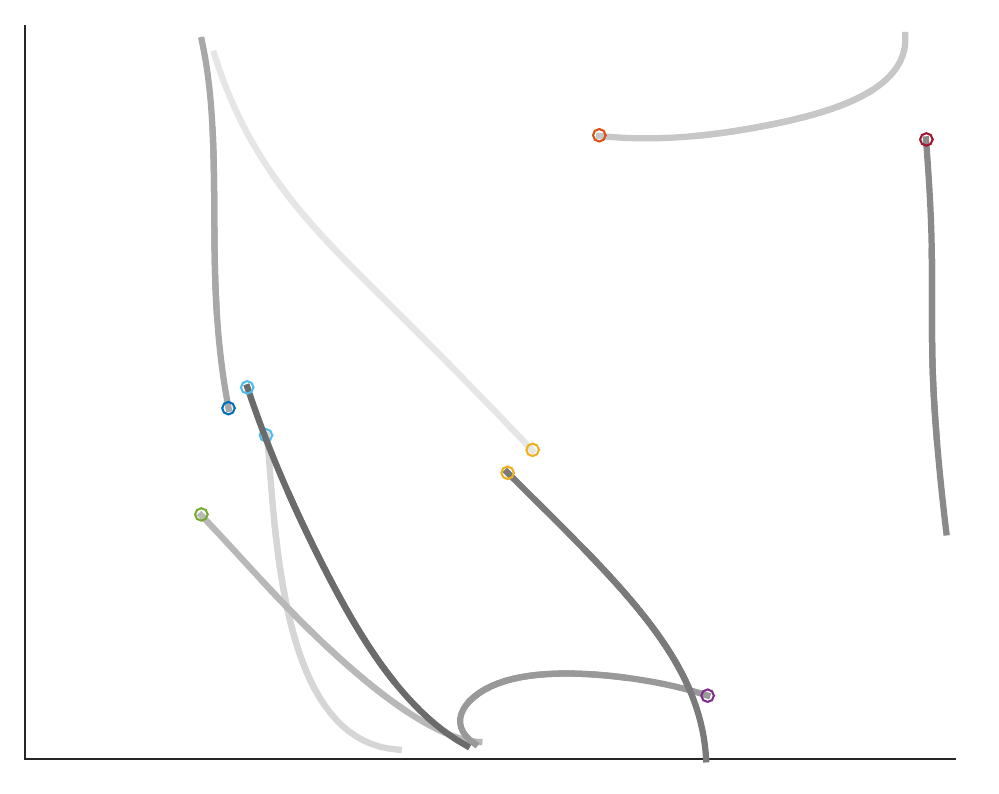}}

\end{center}
\caption{Pedestrian trajectories: Correlated activations from the first layer of the memory.  First column: Highest correlated Memory activations for the pattern selected (in colours) and the rest of the activations (in grey) over time. Second column: The input (observed (in green) and predicted (in blue)) to the model at that time step. Third column: Previous 10 trajectories that reside in the memory. Black to white represents most recent to oldest. }
\label{fig:fig9}
\end{figure*}
We randomly select a memory cell from the first layer of the memory, and analysed the activations of the hidden states of that particular cell. 
Fig. \ref{fig:fig9} shows the results of our analysis. We searched our test set for common activation patterns. The first column of Fig. \ref{fig:fig9} shows the memory activations, with the most highly correlated memory patterns shown in red, green and blue; and the remainder of the activations shown in grey. The second column shows the input to the model (in green) and the predicted sequence (in blue). The previous 10 trajectories that are fed to the memory are shown in the third column. From black to white we have colour coded the most recent trajectory to the oldest trajectory. We expect the first layer of the memory module to generate similar activation patterns when there exists a similar set of historical trajectories, and if there exists a similarity between those historical trajectories and the input.

\subsubsection{Correlated activations from the last layer}

\begin{figure*}[t]
\begin{center}
\subfigure[ ]{\includegraphics[width = .3 \textwidth]{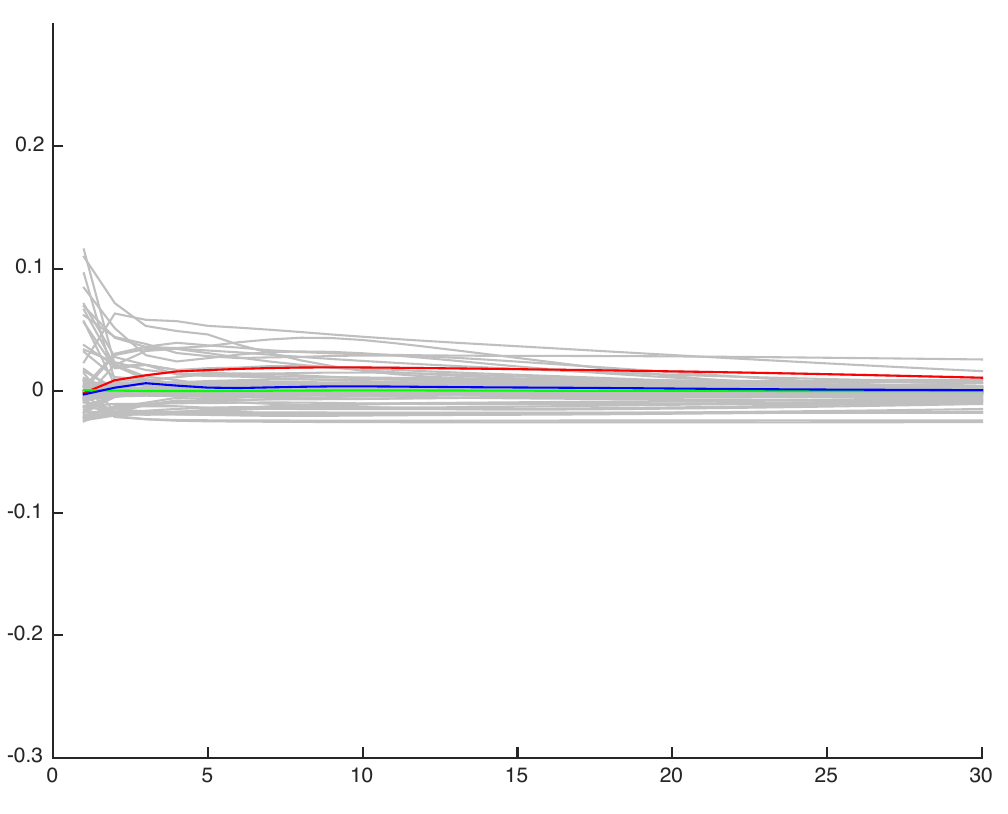}}
\subfigure[]{\includegraphics[width = .3\textwidth]{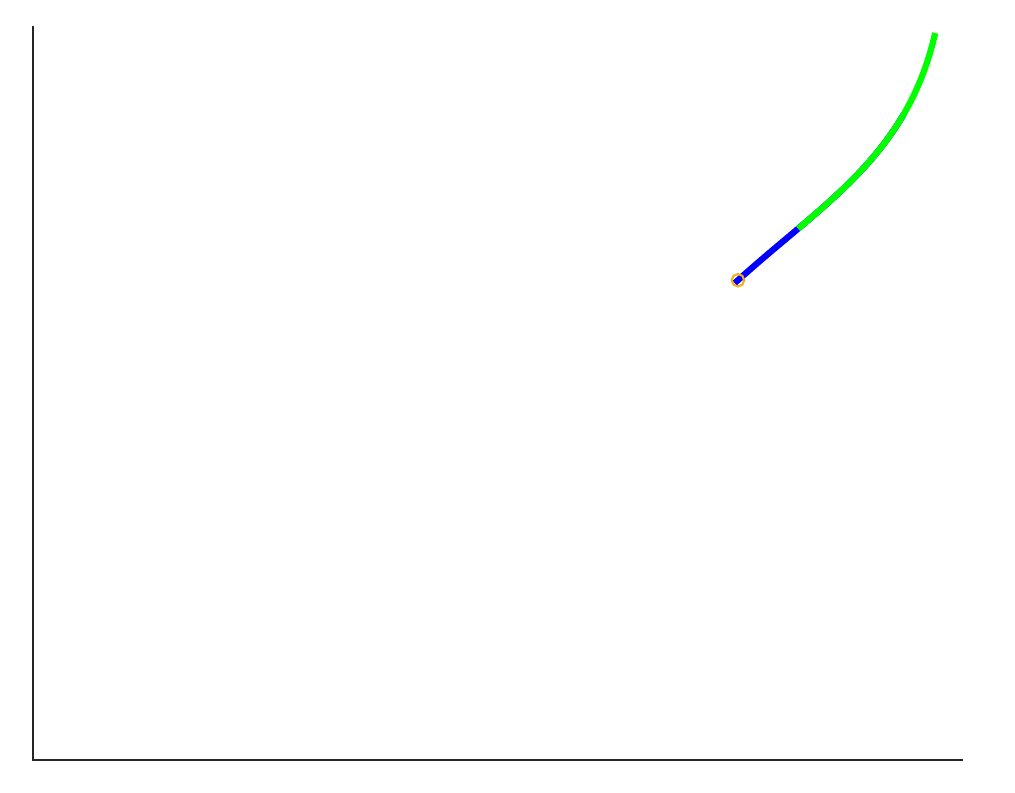}}
\subfigure[]{\includegraphics[width = .3\textwidth]{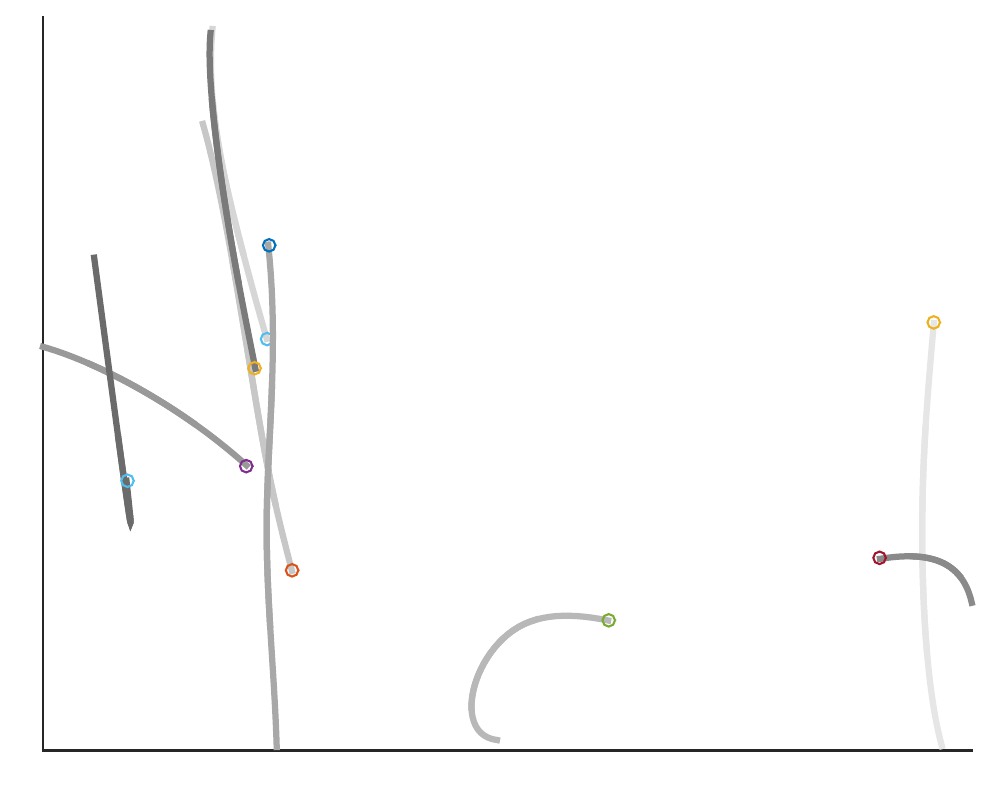}}

\subfigure[ ]{\includegraphics[width = .3 \textwidth]{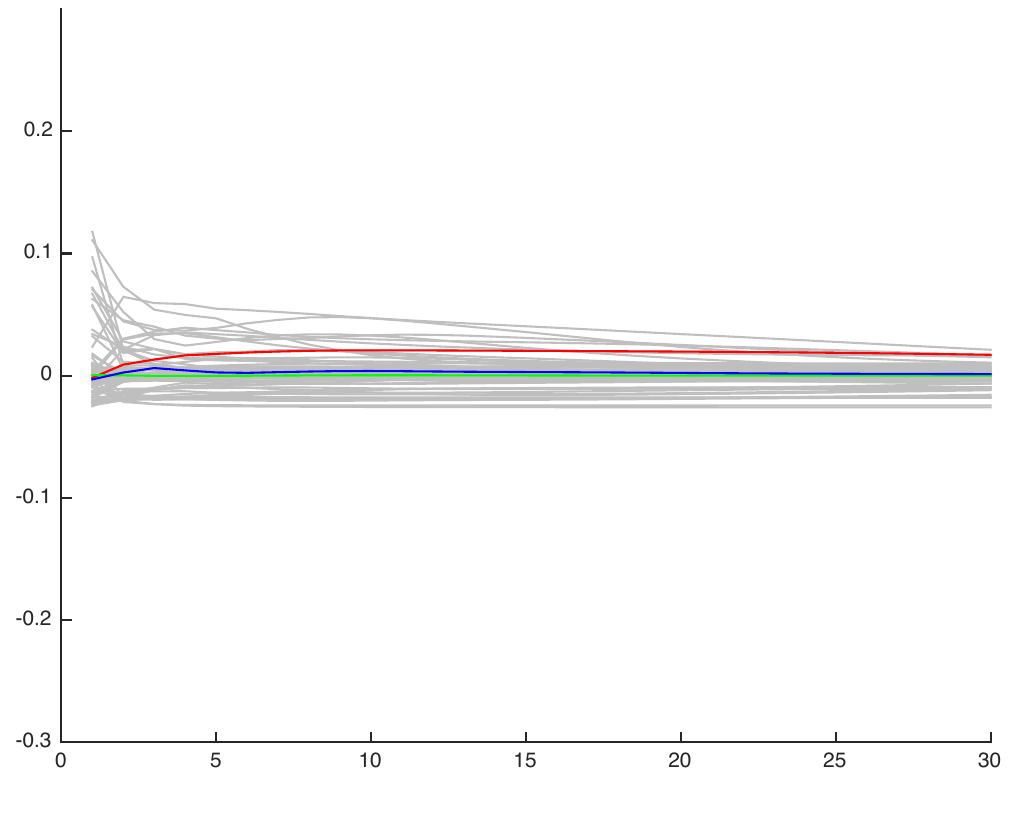}}
\subfigure[ ]{\includegraphics[width = .3\textwidth]{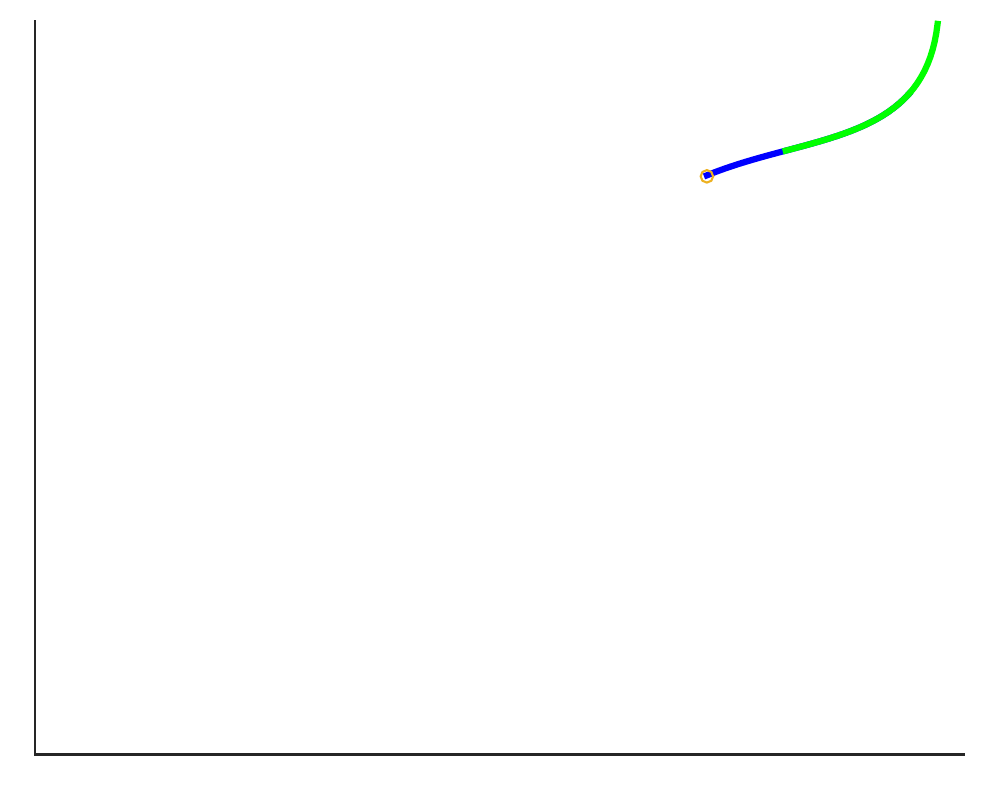}}
\subfigure[ ]{\includegraphics[width = .3\textwidth]{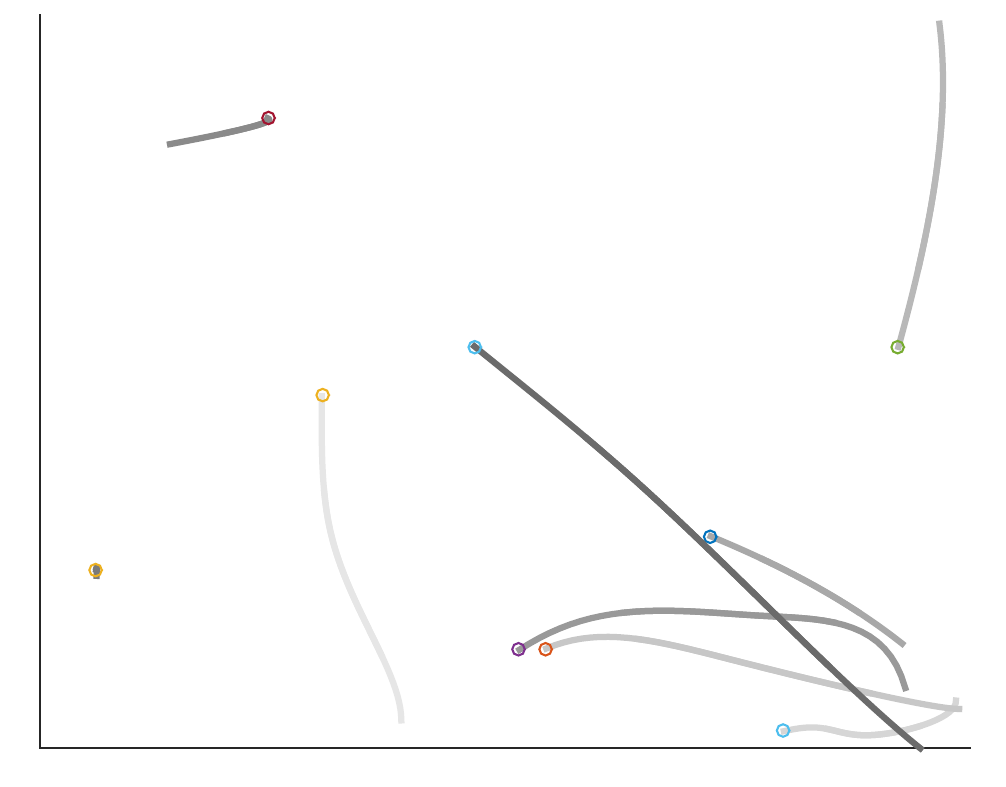}}

\subfigure[ ]{\includegraphics[width = .3 \textwidth]{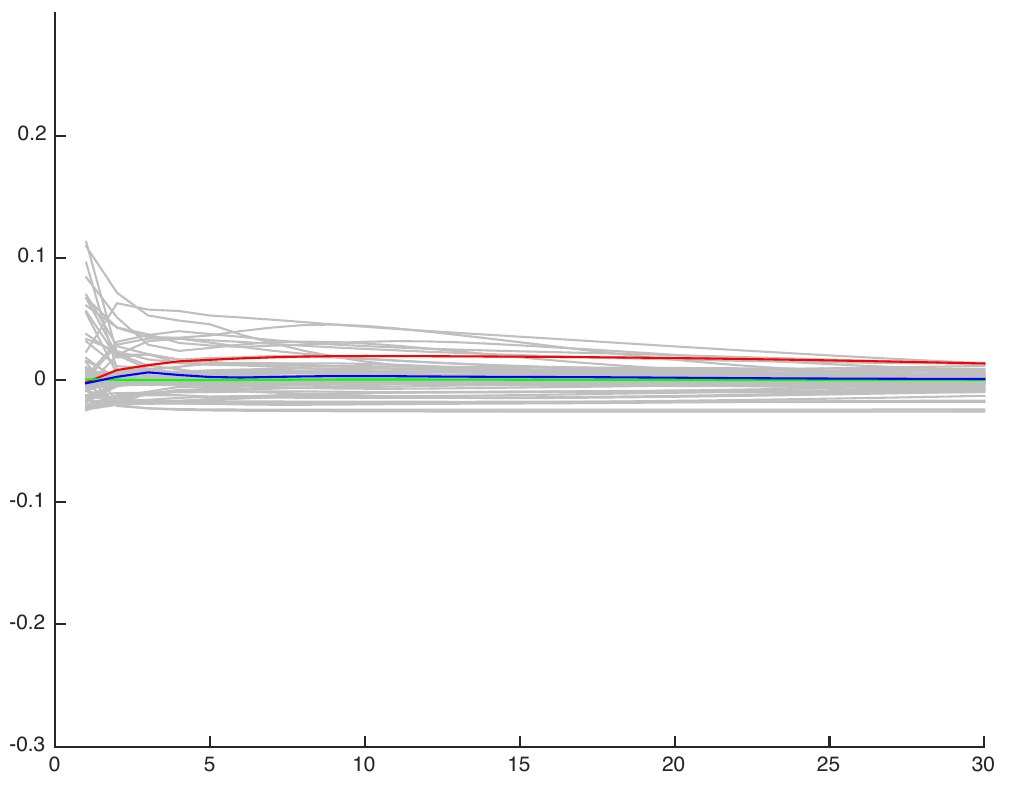}}
\subfigure[]{\includegraphics[width = .3\textwidth]{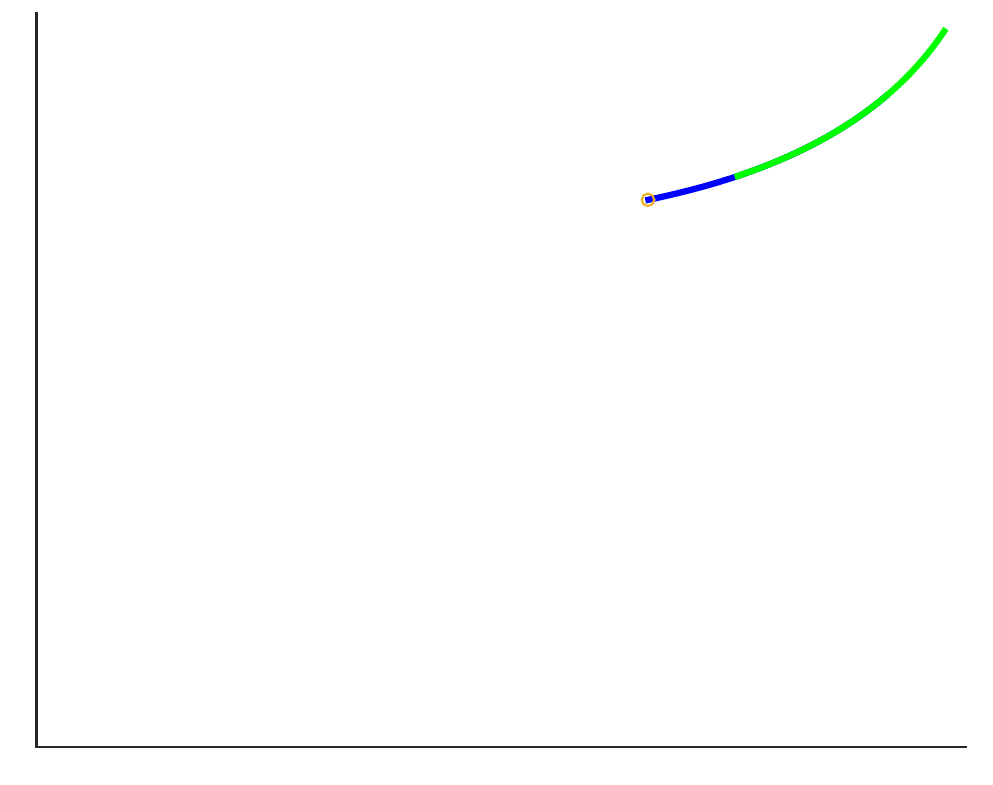}}
\subfigure[ ]{\includegraphics[width = .3\textwidth]{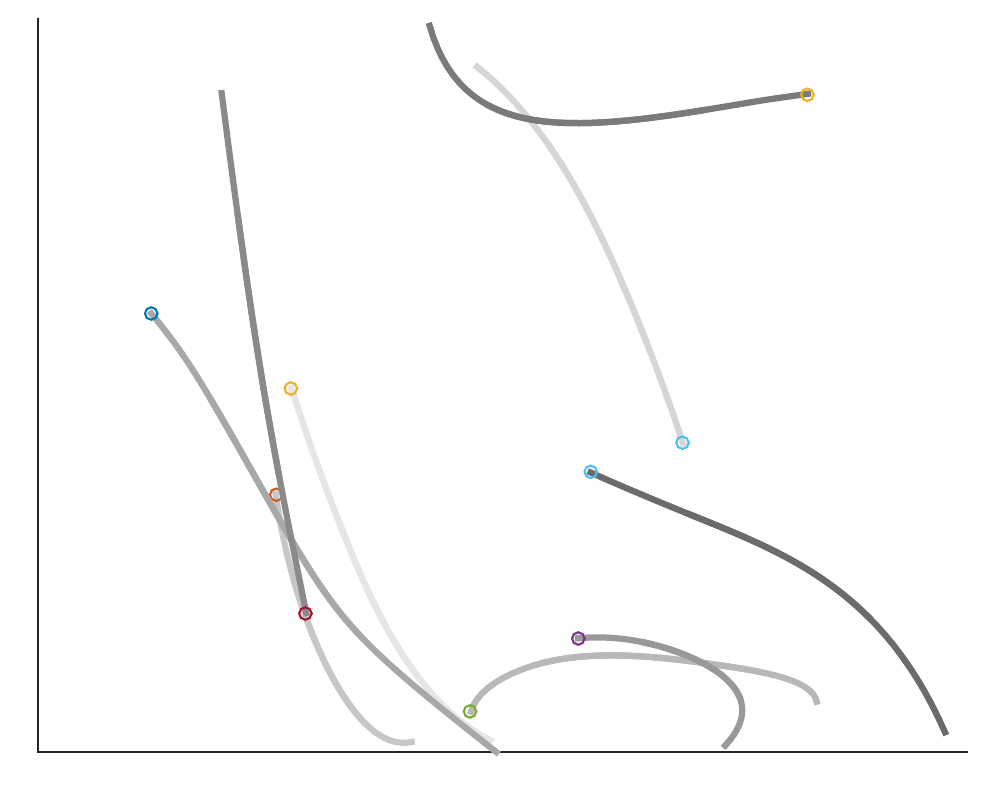}}

\subfigure[]{\includegraphics[width = .3 \textwidth]{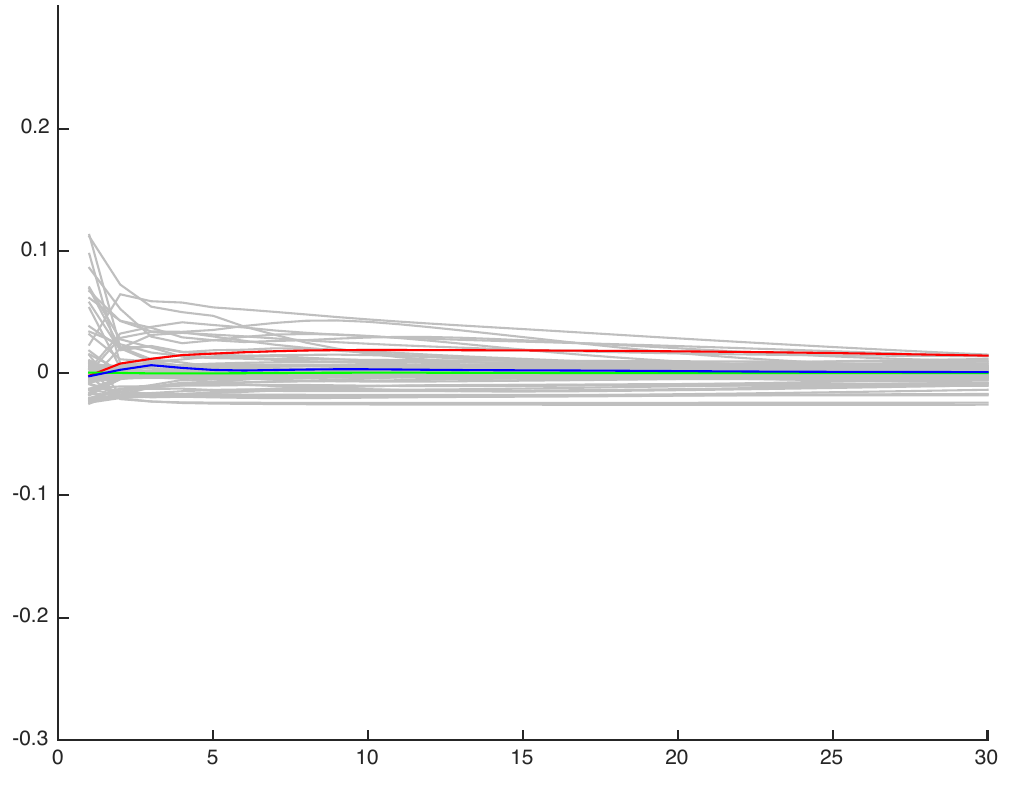}}
\subfigure[]{\includegraphics[width = .3\textwidth]{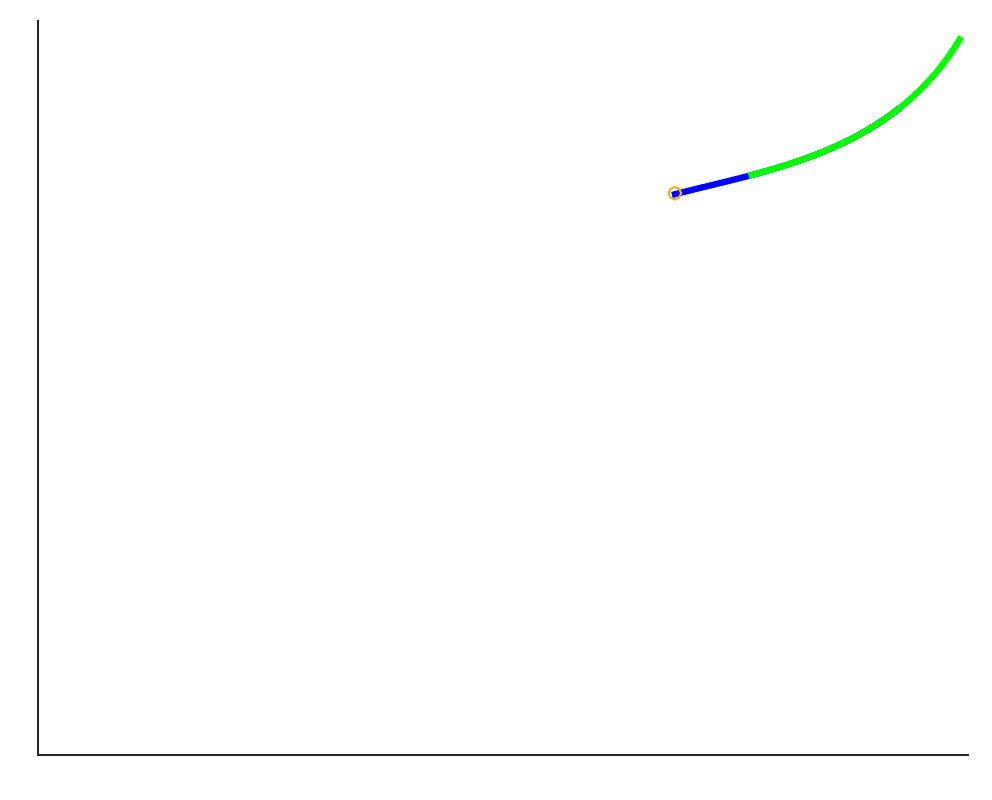}}
\subfigure[]{\includegraphics[width = .3\textwidth]{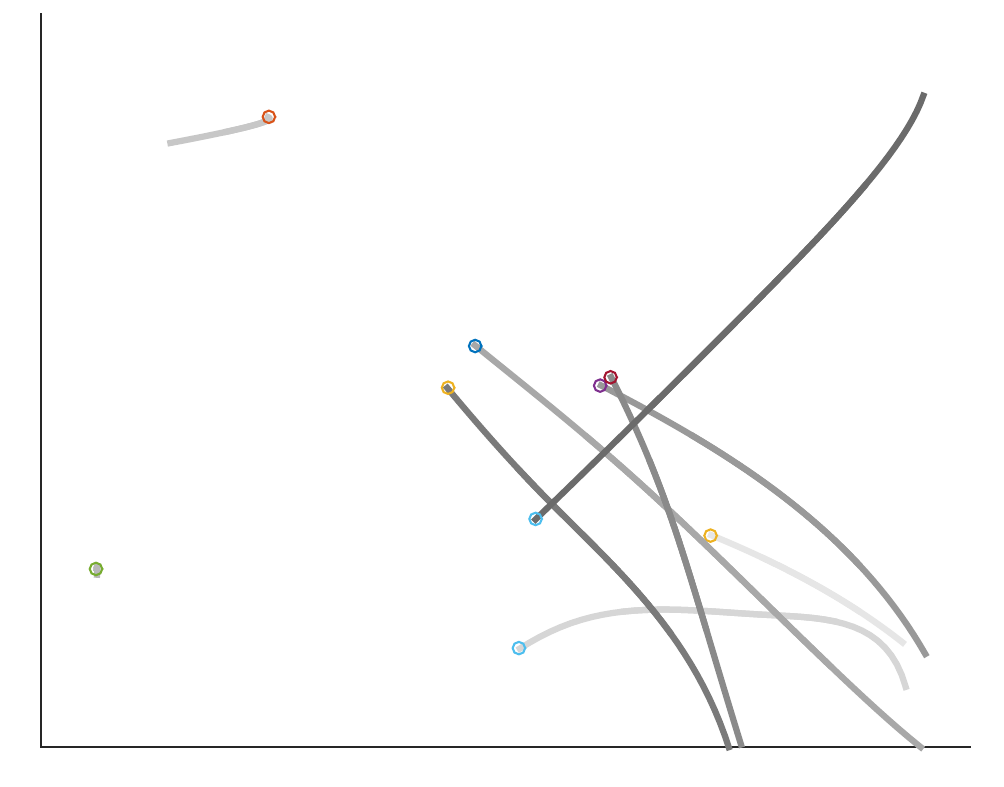}}
\end{center}
\caption{Pedestrian trajectories: Correlated activations from the last layer of the memory.  First column: Highest correlated Memory activations for the pattern selected (in colours) and the rest of the activations (in grey) over time. Second column: The input (observed (in green) and predicted (in blue)) to the model at that time step. Third column: Previous 10 trajectories that reside in the memory. Black to white represents most recent to oldest. }
\label{fig:fig10}
\end{figure*}

Fig. \ref{fig:fig10} illustrates the activations of the final memory cell (i.e last layer). Column descriptions are identical to the that of Fig. \ref{fig:fig9}. The second and third columns of Fig. \ref{fig:fig10} provide visual evidence that the proposed memory module has successfully learnt relationships among input trajectories. If our memory module has enough capacity and if it is capturing long term dependencies, the final layer should generate similar activations for similar input trajectory patterns, rather than being completely reliant on the current short term context. That is evident with the similarity shown in Fig. \ref{fig:fig10} where we observe similar activations (i.e column 1) for similar inputs patterns (i.e column 2). Importantly, we note that despite the similar activations, the recent history (i.e. column 3) differs significantly between cases; unlike in Fig. \ref{fig:fig9}.

\subsubsection{ Activations from the DMN \cite{askMeAnything} memory module}

\begin{figure}[!h]
\begin{center}
\subfigure[ ]{\includegraphics[width = .3 \textwidth]{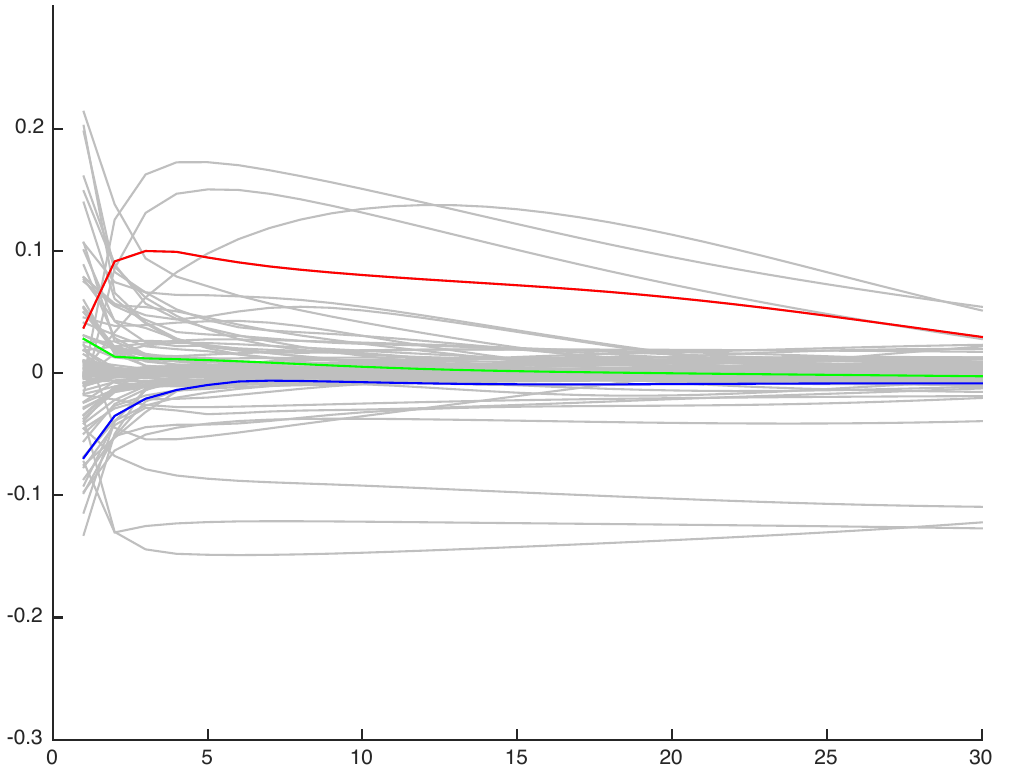}}
\subfigure[]{\includegraphics[width = .3\textwidth]{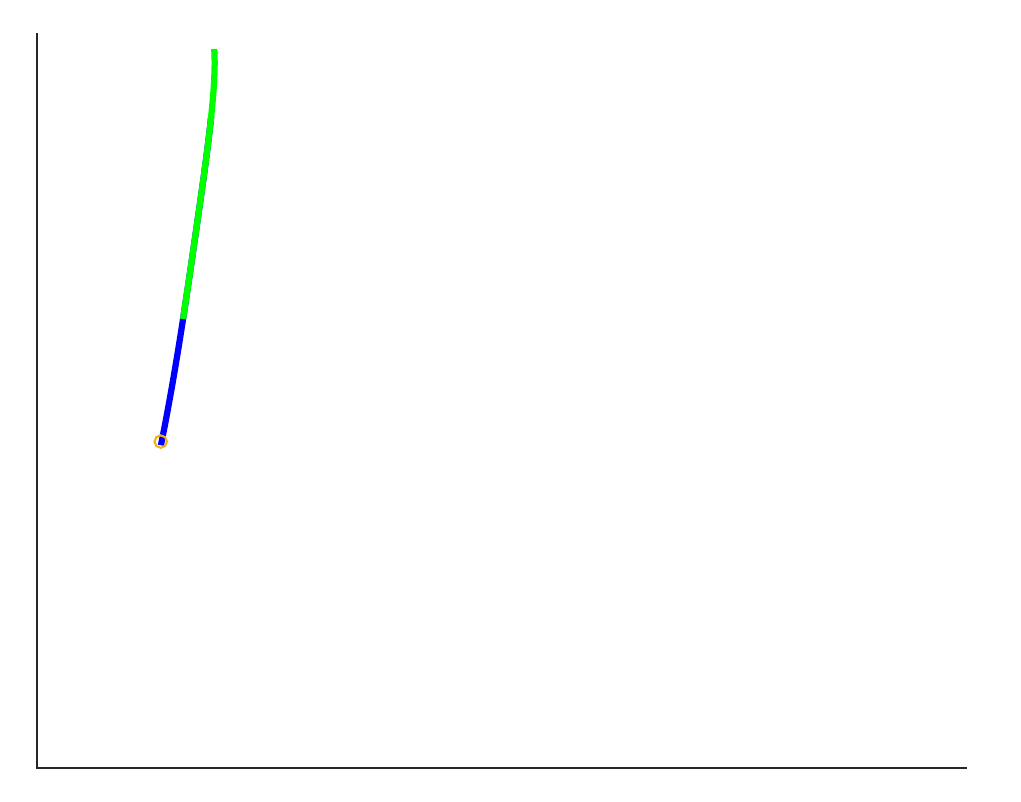}}
\subfigure[]{\includegraphics[width = .3\textwidth]{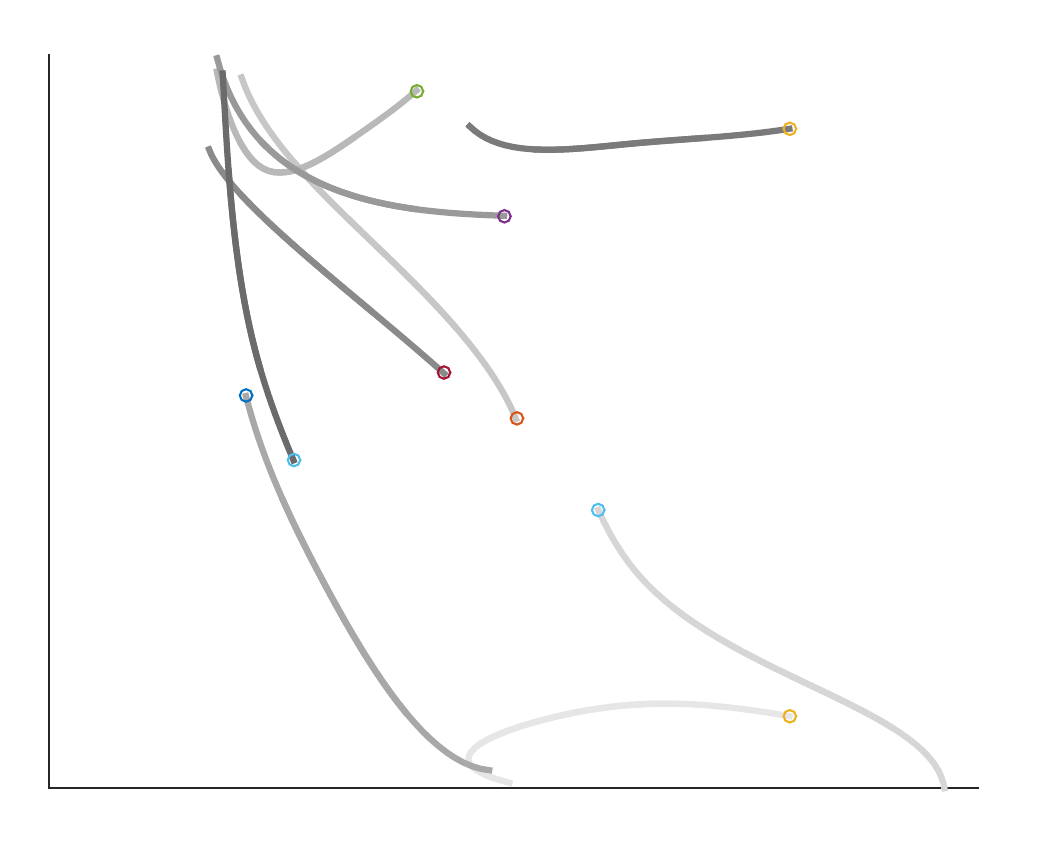}}

\subfigure[ ]{\includegraphics[width = .3 \textwidth]{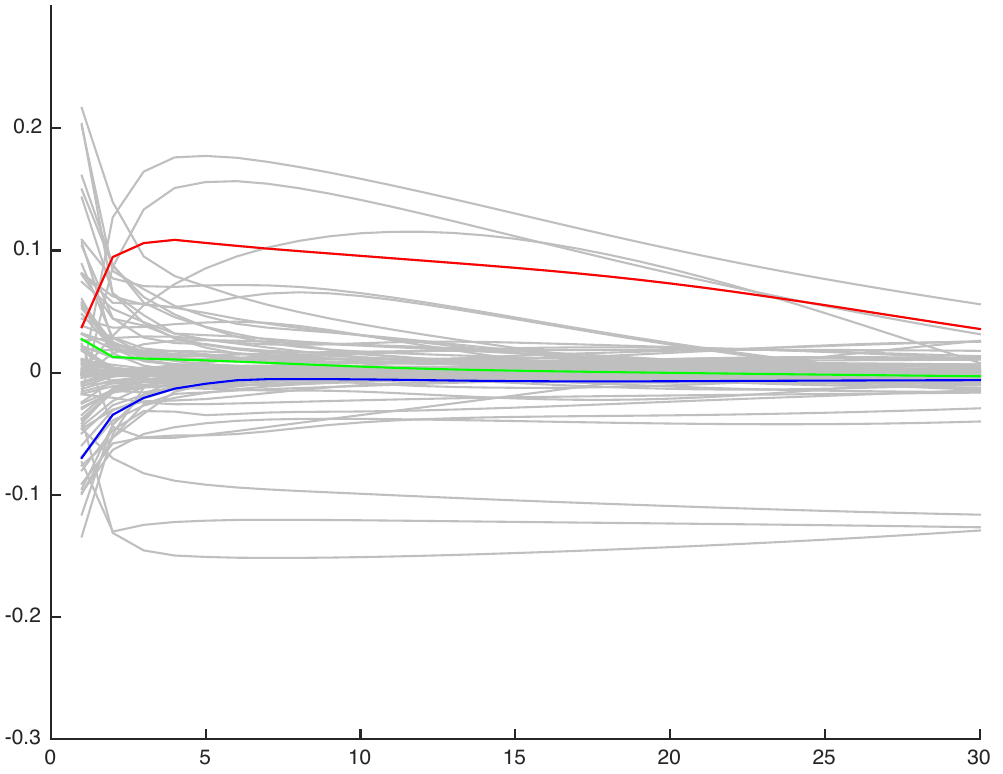}}
\subfigure[ ]{\includegraphics[width = .3\textwidth]{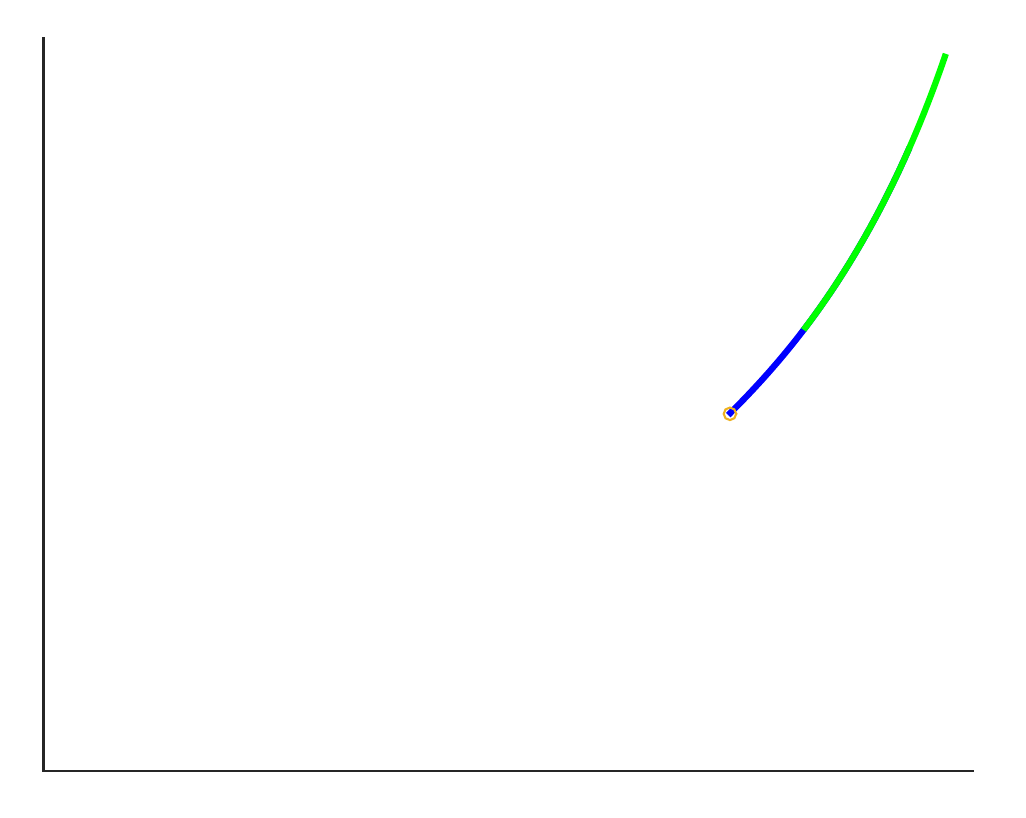}}
\subfigure[ ]{\includegraphics[width = .3\textwidth]{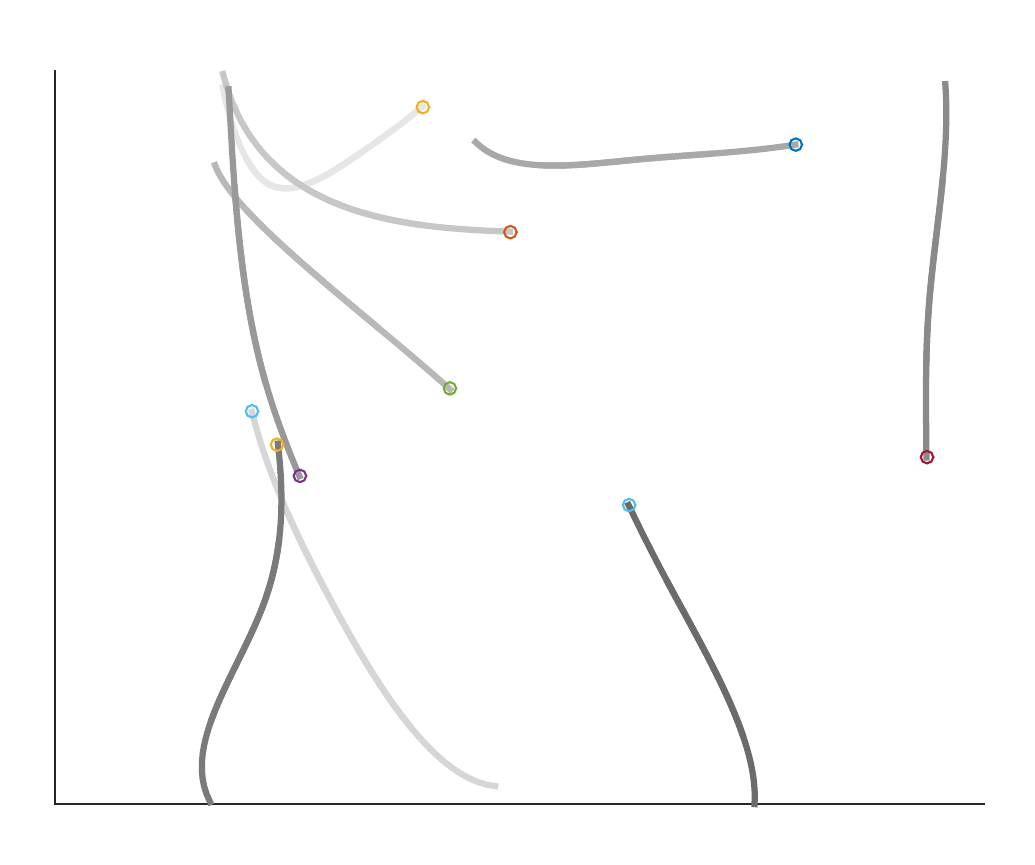}}

\subfigure[ ]{\includegraphics[width = .3 \textwidth]{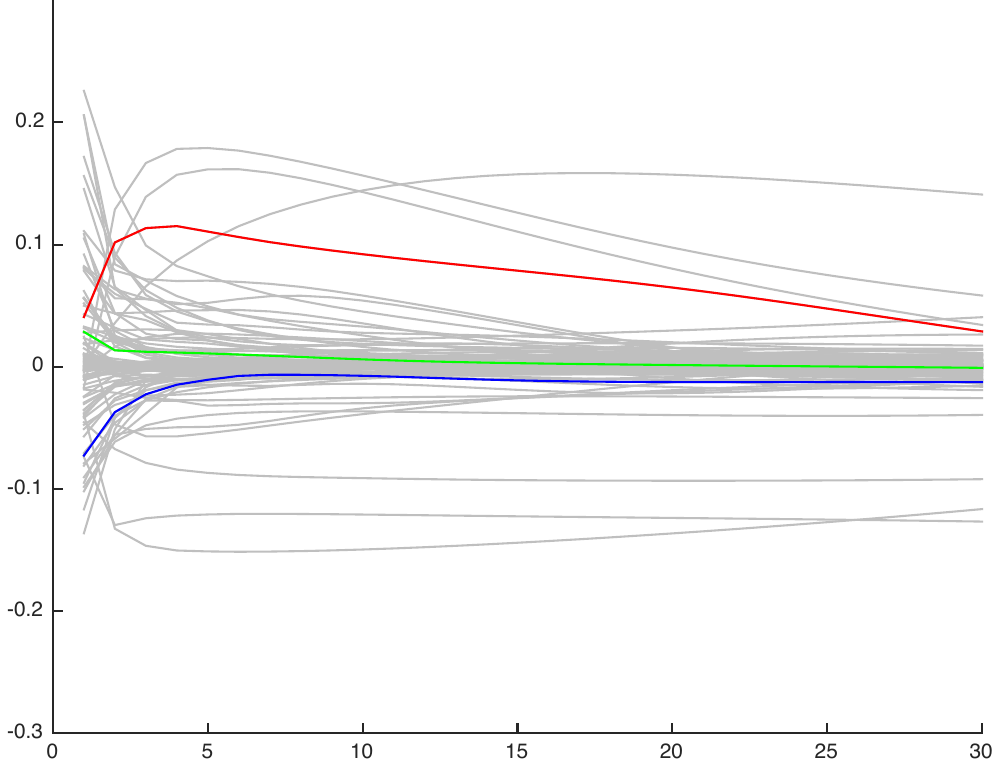}}
\subfigure[]{\includegraphics[width = .3\textwidth]{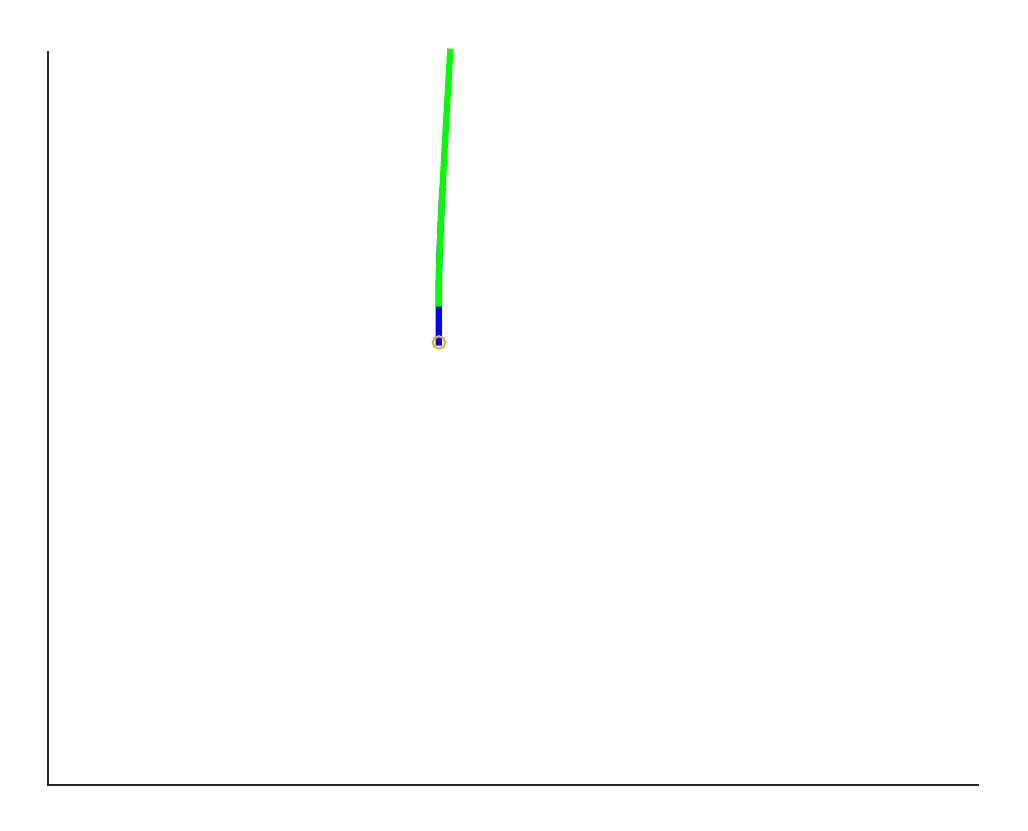}}
\subfigure[ ]{\includegraphics[width = .3\textwidth]{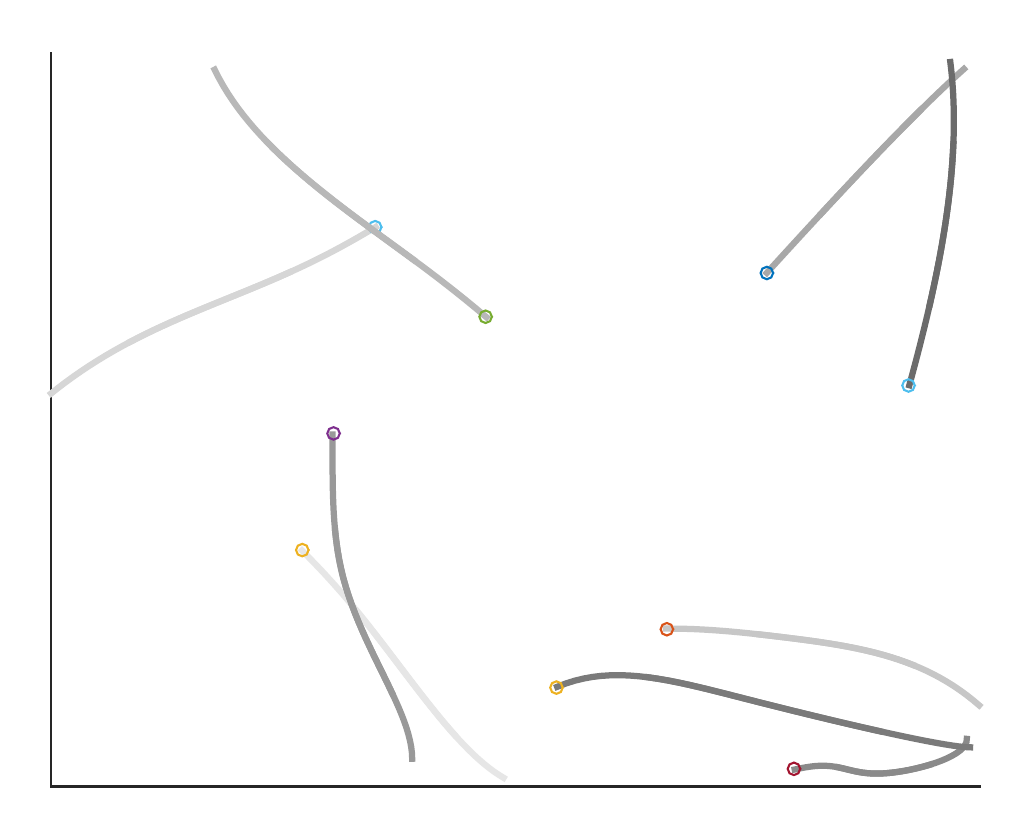}}

\subfigure[]{\includegraphics[width = .3 \textwidth]{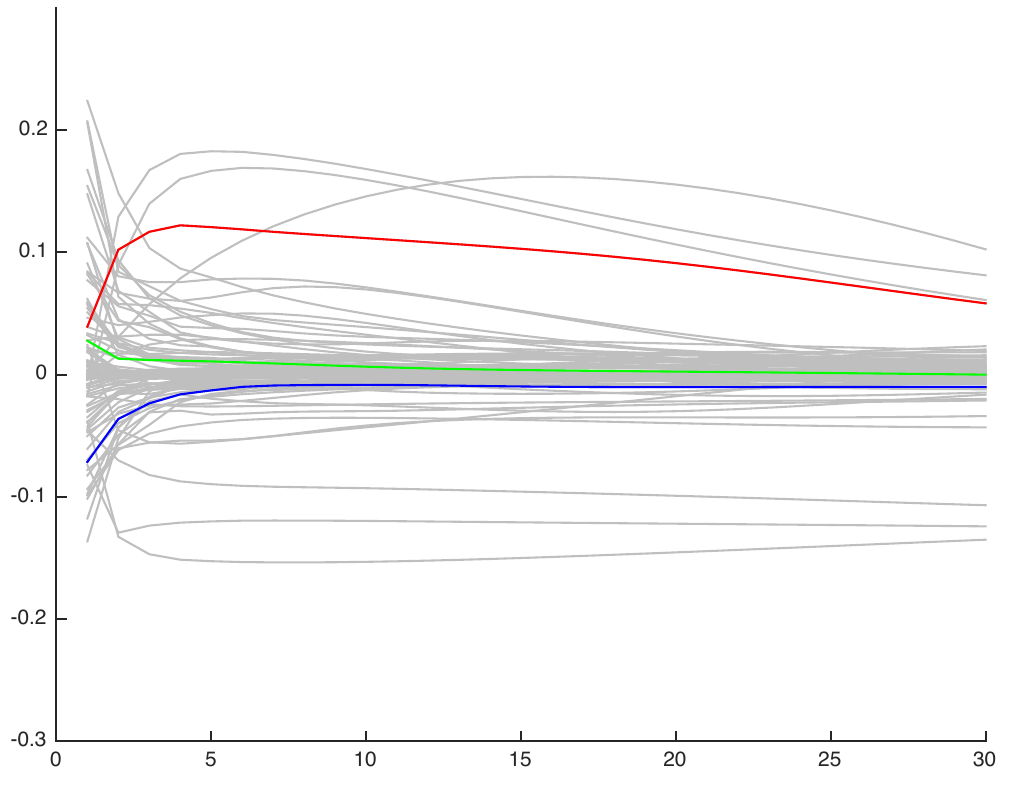}}
\subfigure[]{\includegraphics[width = .3\textwidth]{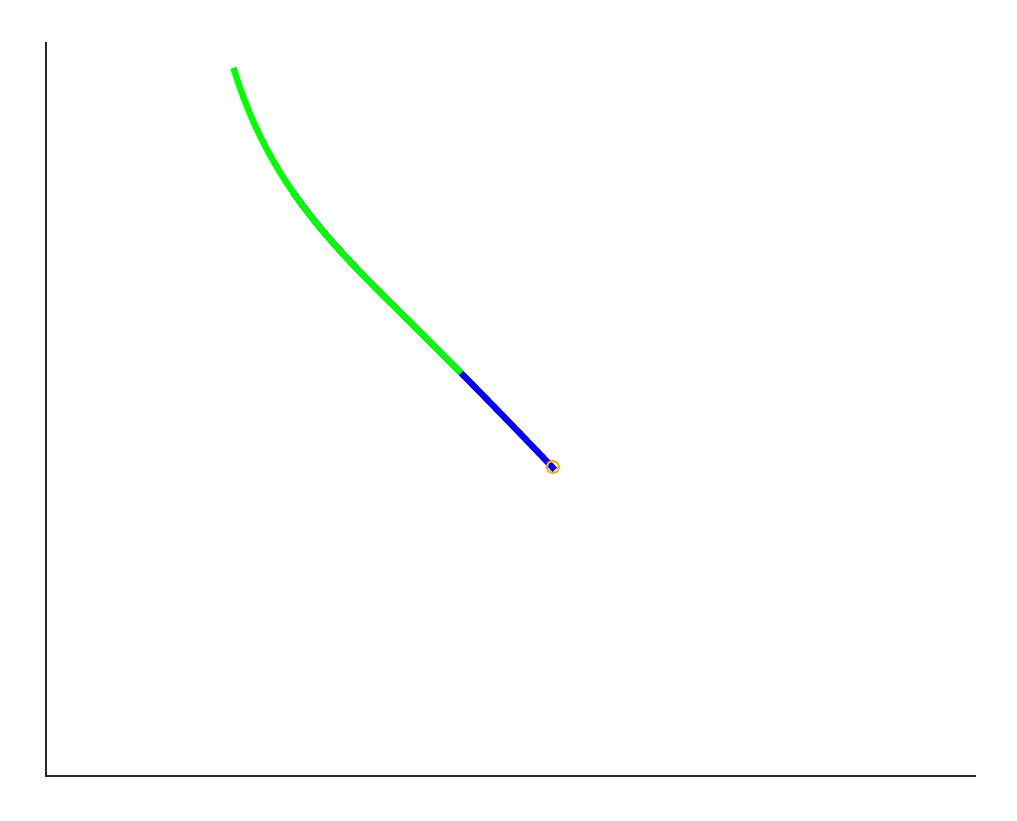}}
\subfigure[]{\includegraphics[width = .3\textwidth]{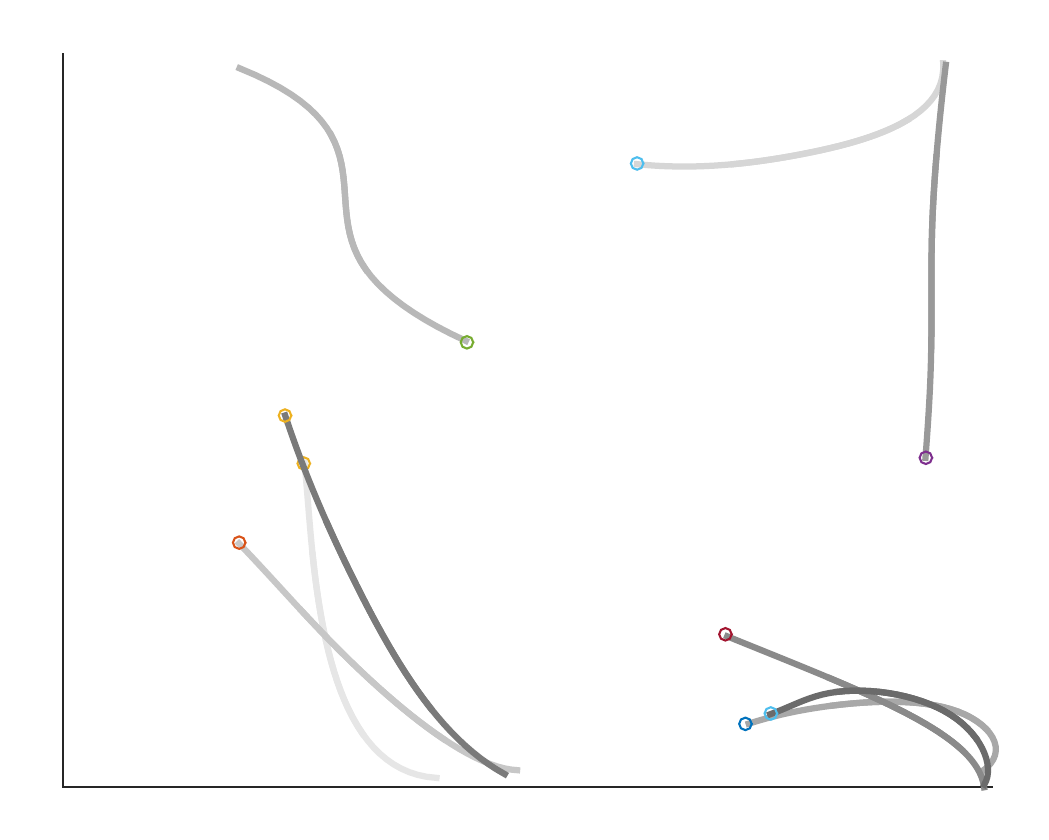}}
\end{center}
\caption{ Pedestrian trajectories: Correlated activations from baseline memory model.  First column: Highest correlated Memory activations for the pattern selected (in colours) and the rest of the activations (in grey) over time. Second column: The input (observed (in green) and predicted (in blue)) to the model at that time step. Third column: Previous 10 trajectories that reside in the memory. Black to white represents most recent to oldest. }
\label{fig:fig8}
\end{figure}

In Fig. \ref{fig:fig8} we visualise the correlated activations from the DMN memory model shown in Fig. \ref{fig:fig1} (a). The column labels are identical to that of Fig. \ref{fig:fig9}. We compare the activations in Fig. \ref{fig:fig10} to those in Fig. \ref{fig:fig8}. When observing Fig \ref{fig:fig8} column 3 it is evident that hidden state activations are dominated by the most recent inputs to the memory, and the long term dependencies are of little importance. This is noted to be an inherent problem with sequential LSTM architectures \cite{enhancingCombining}. Therefore regardless of the input to the model (shown in Fig \ref{fig:fig8} column 2) the memory module is generating similar activations and is only considering the short term context. Hence the prediction error is high. Furthermore we would like to highlight the similarity between Fig \ref{fig:fig8} and the first layer of the proposed memory module (shown in Fig. \ref{fig:fig9} ).

\par

\begin{figure}[!h]
\begin{center}
\subfigure[ ]{\includegraphics[width = .3 \textwidth]{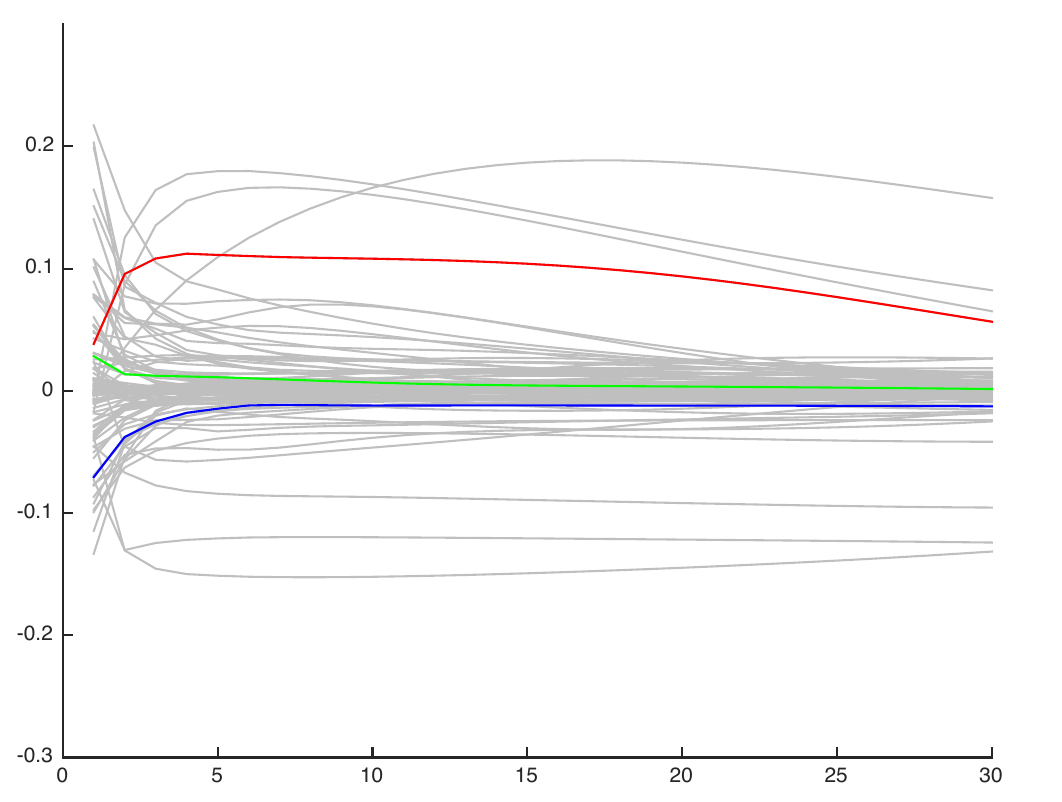}}
\subfigure[]{\includegraphics[width = .3\textwidth]{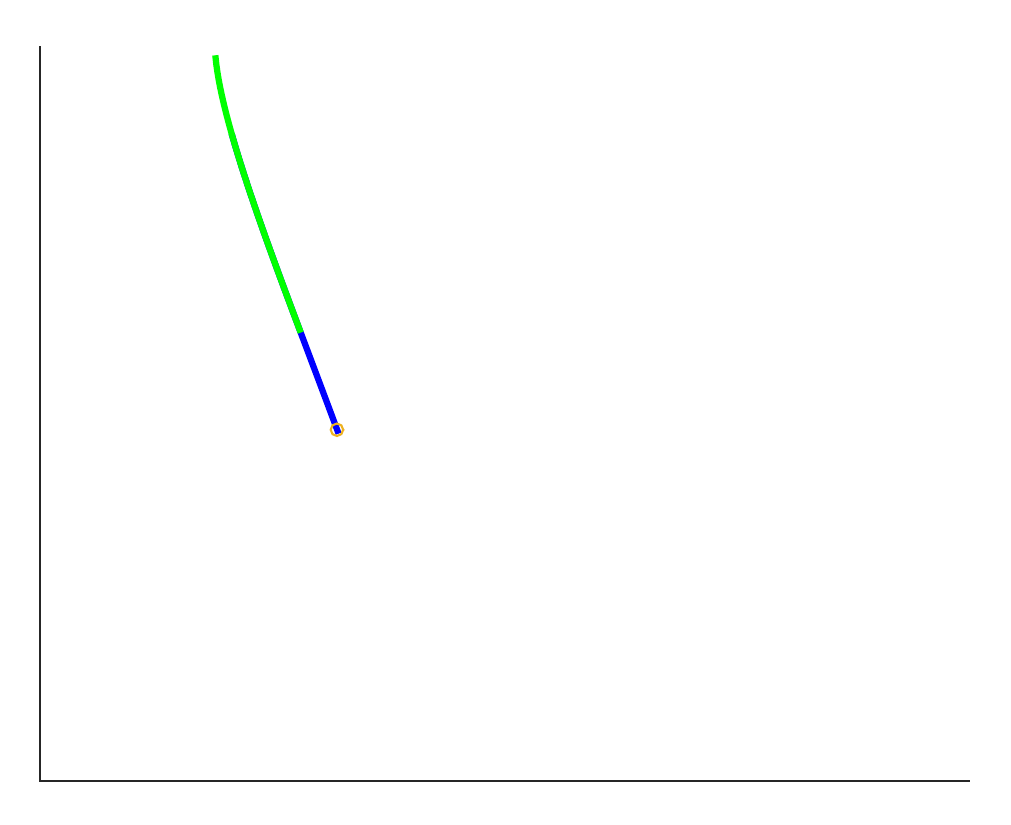}}
\subfigure[]{\includegraphics[width = .3\textwidth]{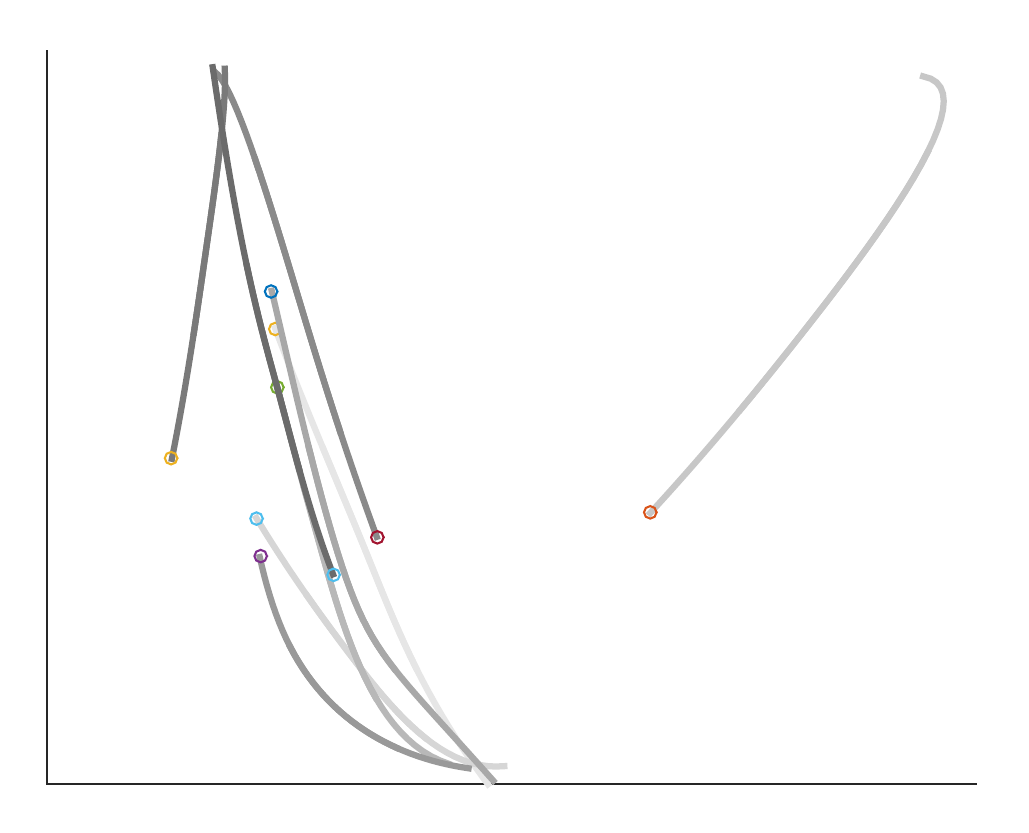}}

\subfigure[ ]{\includegraphics[width = .3 \textwidth]{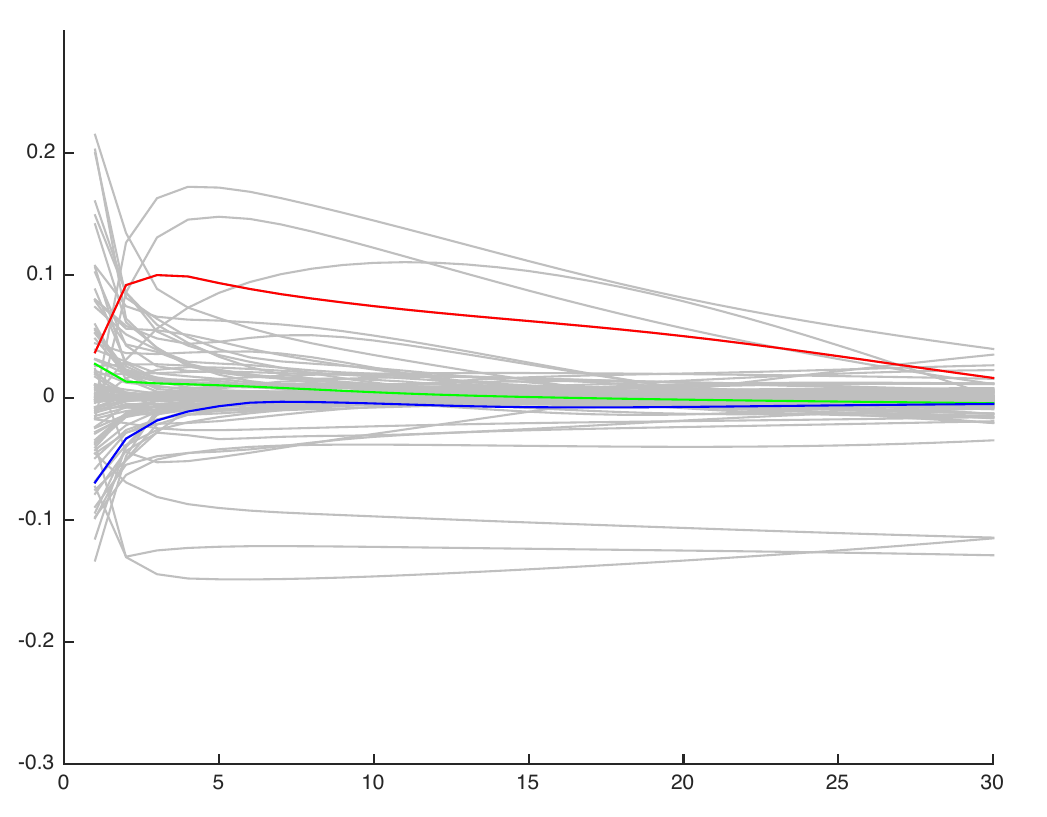}}
\subfigure[ ]{\includegraphics[width = .3\textwidth]{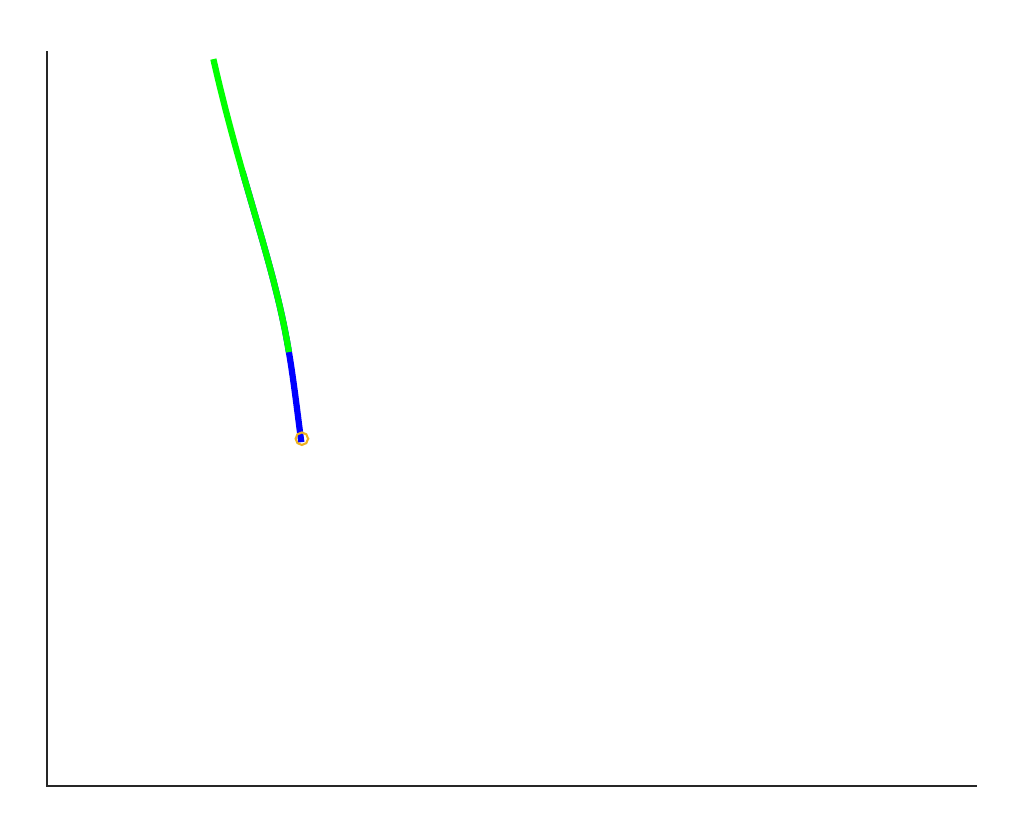}}
\subfigure[ ]{\includegraphics[width = .3\textwidth]{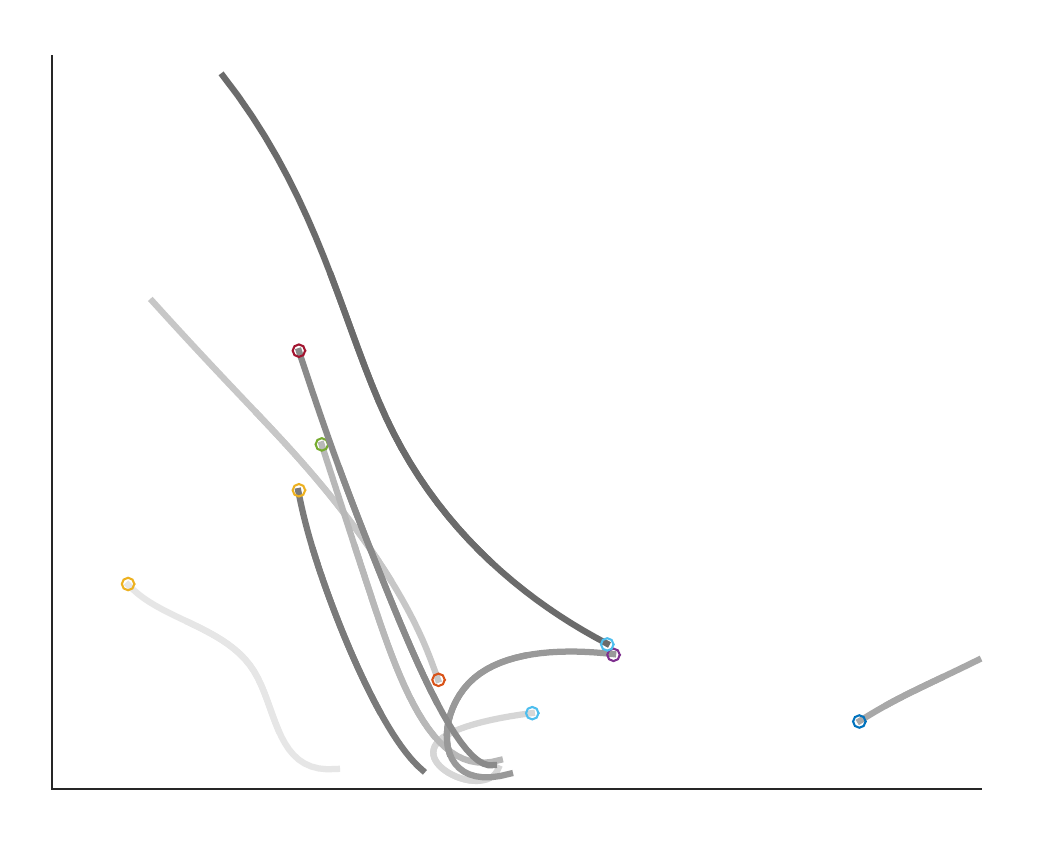}}

\subfigure[ ]{\includegraphics[width = .3 \textwidth]{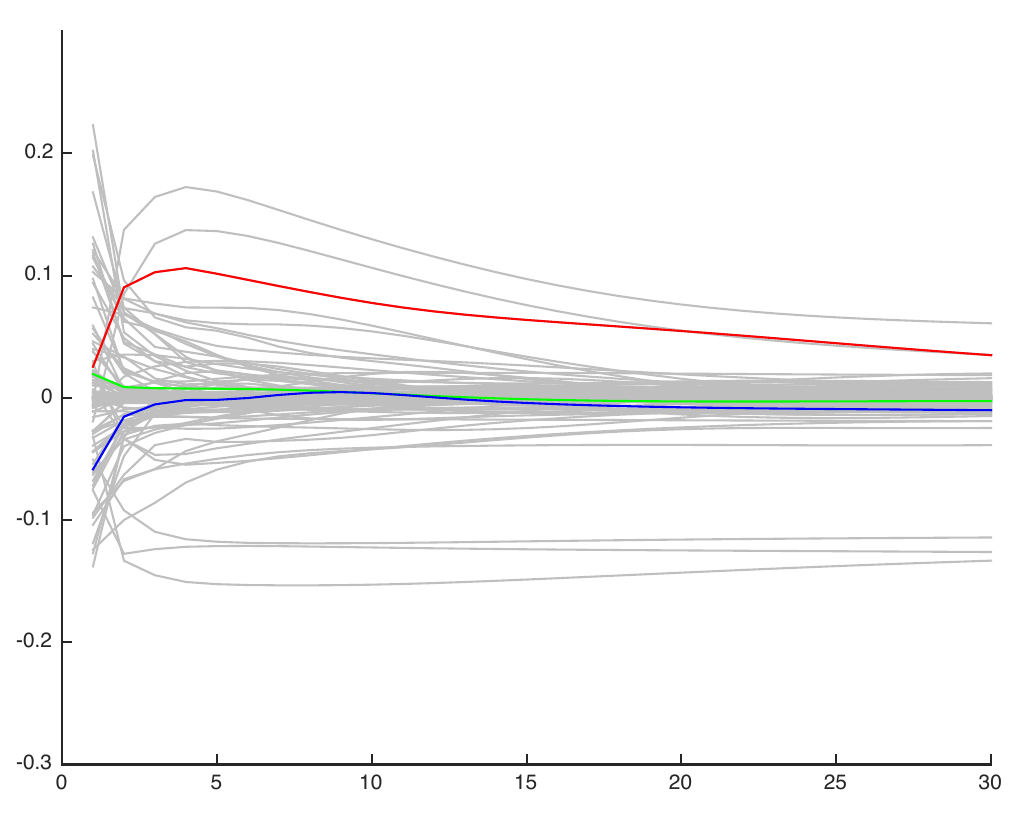}}
\subfigure[]{\includegraphics[width = .3\textwidth]{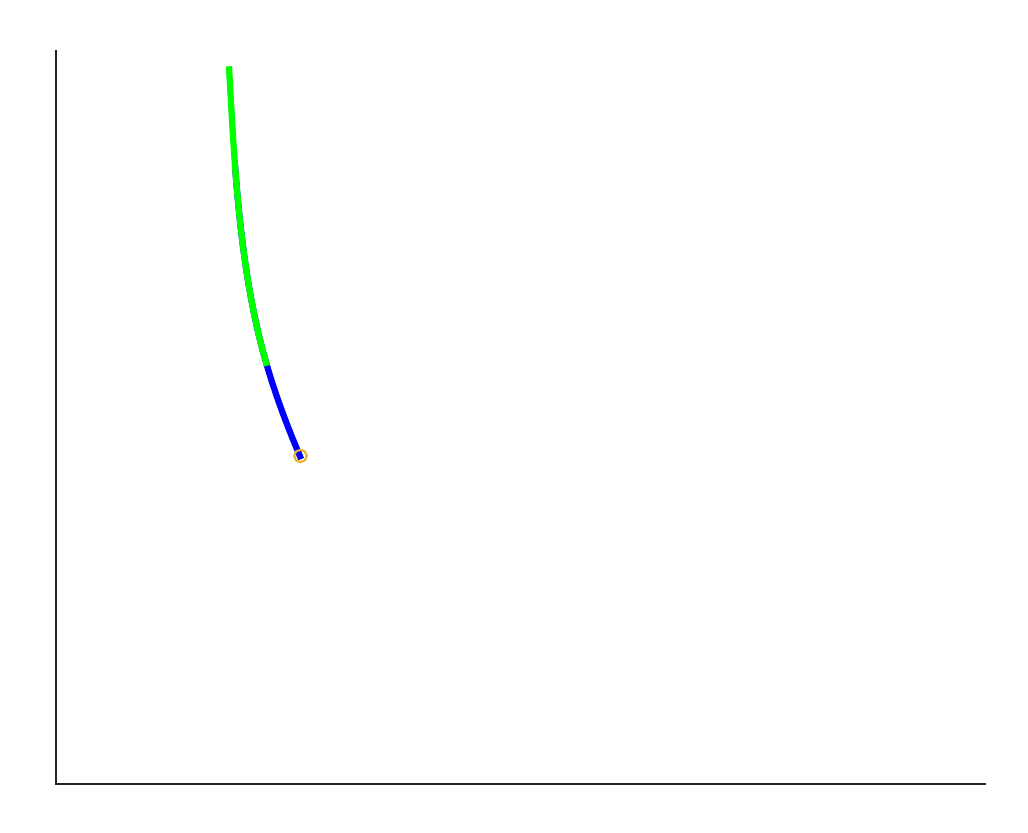}}
\subfigure[ ]{\includegraphics[width = .3\textwidth]{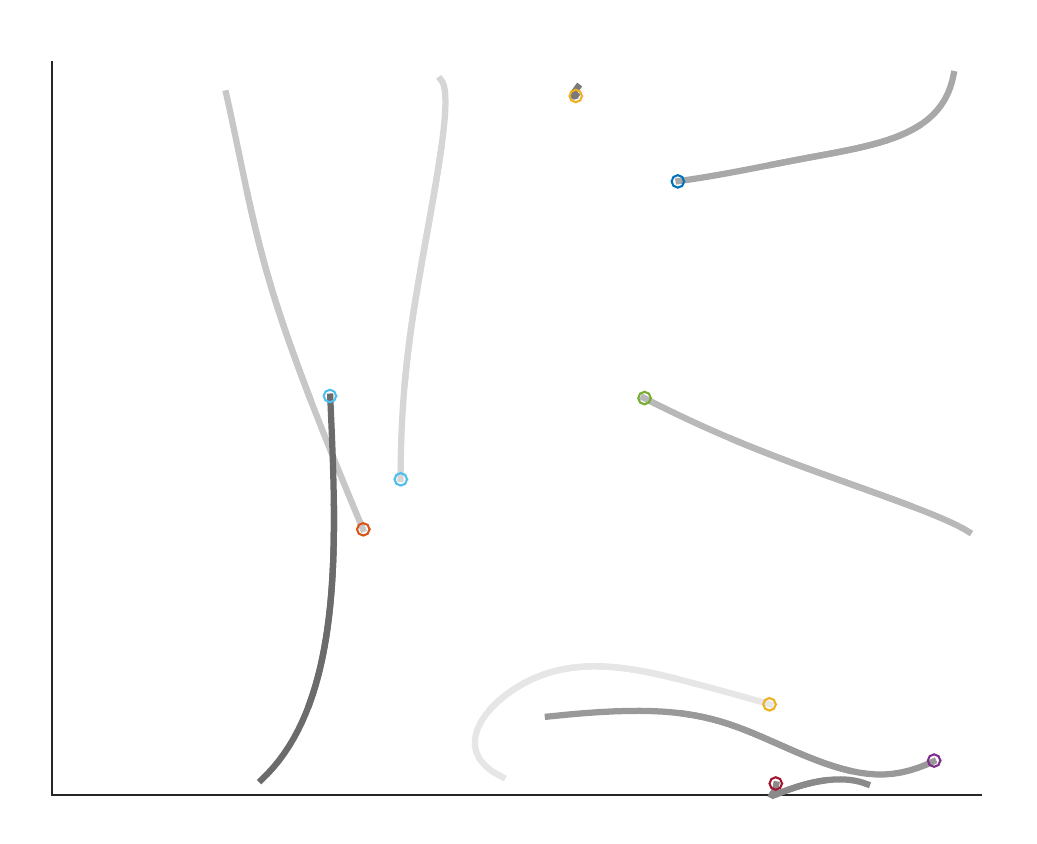}}

\subfigure[]{\includegraphics[width = .3 \textwidth]{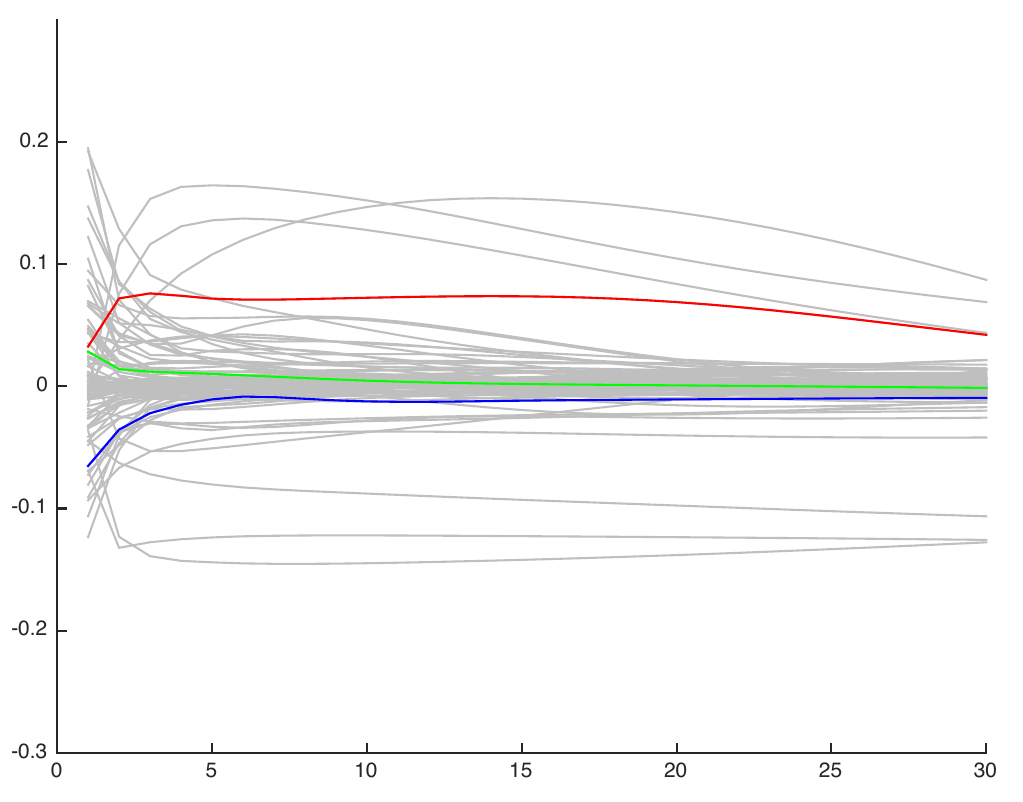}}
\subfigure[]{\includegraphics[width = .3\textwidth]{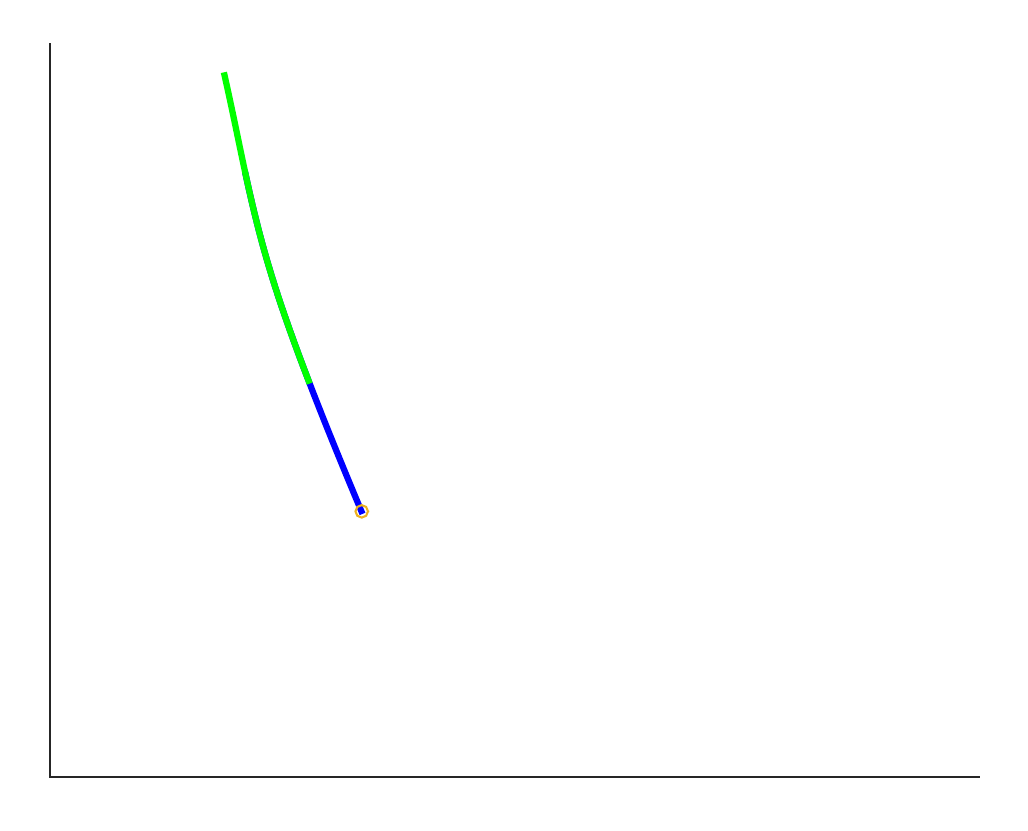}}
\subfigure[]{\includegraphics[width = .3\textwidth]{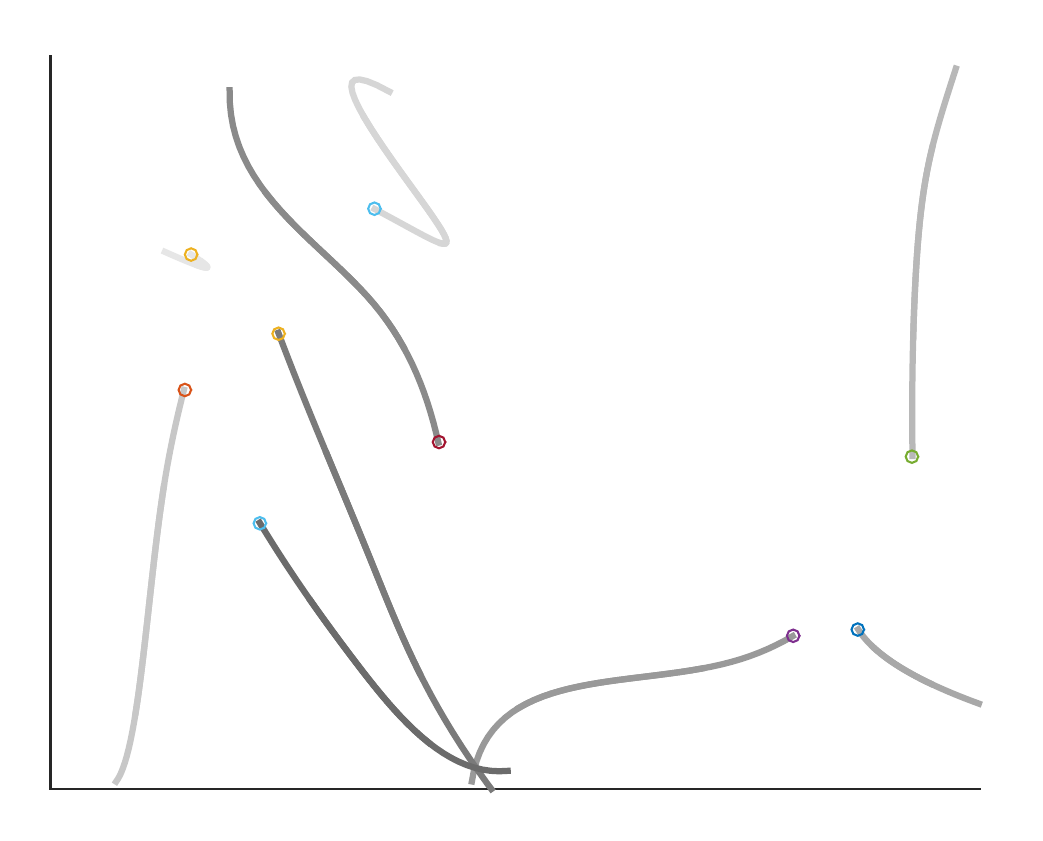}}
\end{center}
\caption{ Pedestrian trajectories: Memory activations from DMN memory model for correlated inputs.  First column: Memory activations over time, the pattern we are considering (in colours) and the rest of the activations (in grey). Second column: The correlated inputs (observed (in green) and predicted (in blue)). Third column: For the input selected in the second column, previous 10 trajectories that reside in the memory. Black to white represents most recent to oldest. }
\label{fig:fig11}
\end{figure}

To further illustrate the limitations of the DMN model, we cluster the input trajectories and from one particular cluster we extract out the input trajectories with the highest correlation (shown in Fig. \ref{fig:fig11}) within that cluster. Given that we have very similar inputs, we expect the DMN module to generate similar activations, however we observe that the DMN generates vastly different activations. In order to highlight the differences among memory activations we randomly selected 3 hidden units within the memory and illustrate their temporal evolutions (see the coloured lines in Fig. \ref{fig:fig11} column 1). From Fig. \ref{fig:fig11}, we can see that the short term history across the examples is varied (see Fig. \ref{fig:fig11} column 3), and the memory module generates vastly different activations for each, ignoring the given input. 

Considering Fig. \ref{fig:fig8} and Fig. \ref{fig:fig11} together, we can see that the activations are driven by the short term history. Similar short term histories with different inputs lead to similar activations (shown in Fig. \ref{fig:fig8} column 1); and different short term histories with similar inputs result in vastly different activations (shown in Fig. \ref{fig:fig11} column 1). This is in contrast to the proposed approach, where as shown in Fig. \ref{fig:fig10}, at higher levels of the hierarchy memory activations are driven by the input.

\section{Conclusion}
The proposed tree memory network (TMN) model is a generalised architecture for modelling long term and short term relationships, which can be applied directly for any sequence-to-sequence mapping task. Through the evaluation results we demonstrated that our proposed memory architecture is able to outperform all considered baselines, and we provide visual evidence on the power of TMN which is able to capture both long term and short term relationships via an efficient tree structure. We have demonstrated our approach on two different trajectory prediction applications. The varied nature of these problems demonstrates how the proposed TMN model can be directly applied to any sequence-to-sequence prediction problem where modelling long term relationships is necessary. In future work we will be exploring the applications of TMN as an encoding mechanism for large scale multi-modal inputs such as videos (i.e. a sequence of images), where the encoded vector representation of the memory can be utilised to generate a sparse representation of the entire video sequence with its temporal relationships. 

\bibliographystyle{elsarticle-num} 
\bibliography{my_database}

\end{document}